\documentclass[journal]{IEEEtran}

\usepackage{ulem}
\usepackage{times}
\usepackage{epsfig}
\usepackage{amsmath}
\usepackage{amssymb}
\usepackage{amsmath}

\usepackage[pagebackref=true,breaklinks=true,colorlinks,bookmarks=false]{hyperref}

\usepackage{enumerate}
\usepackage{multirow} 
\usepackage{stfloats}
\usepackage{booktabs}
\usepackage{diagbox}
\usepackage{makecell}

\usepackage{graphics}
\usepackage{subfigure}
\graphicspath{{figs/}}

\usepackage{algorithm}
\usepackage{algorithmic}
\usepackage{color}
\usepackage{colortbl}
\usepackage{comment}
\usepackage[table]{xcolor}
\usepackage[utf8]{inputenc}
\DeclareUnicodeCharacter{0303}{\~{}}

\usepackage[T1]{fontenc}

\definecolor{LightRed}{rgb}{1,0.92,0.92}
\definecolor{LightOrange}{rgb}{1,0.95,0.88}
\definecolor{LightYellow}{rgb}{1.0,1.0,0.84}
\definecolor{LightGreen}{rgb}{0.9,1.0,0.88}
\definecolor{LightCyan}{rgb}{0.9,1,1}
\definecolor{LightBlue}{rgb}{0.9,0.94,1}
\definecolor{LightIndigo}{rgb}{0.92,0.9,1}
\definecolor{LightMagenta}{rgb}{0.96,0.86,1}
\definecolor{DirtyWhite}{rgb}{0.96,0.96,0.96}

\begin{document}

\newcommand{\revise}[1]{\textcolor{black}{#1}}

\title{EarthNets: Empowering AI in Earth Observation}

\author{Zhitong~Xiong, \IEEEmembership{Member,~IEEE}, Fahong Zhang, Yi Wang, Yilei Shi, \IEEEmembership{Member,~IEEE}, \\and Xiao Xiang Zhu, \IEEEmembership{Fellow,~IEEE}.

\thanks{Zhitong Xiong, Fahong Zhang, and Yi Wang are with the Chair of Data Science in Earth Observation, Technical University of Munich, Munich 80333, Germany (email: zhitong.xiong@tum.de; fahong.zhang@tum.de; yi4.wang@tum.de).  }
\thanks{Y. Shi is with the Chair of Remote Sensing Technology, Technical University of Munich (TUM), 80333 Munich, Germany (e-mail: yilei.shi@tum.de).}
\thanks{X. X. Zhu is with the Chair of Data Science in Earth Observation, Technical University of Munich, 80333 Munich, Germany, and also with the Munich Center for Machine Learning, 80333 Munich, Germany (e-mail: xiaoxiang.zhu@tum.de).}
}


\maketitle

\begin{abstract}
    Earth observation (EO), aiming at monitoring the state of planet Earth using remote sensing data, is critical for improving our daily lives and living environment. With a growing number of satellites in orbit, an increasing number of datasets with diverse sensors and research domains are being published to facilitate the research of the remote sensing community. This paper presents a comprehensive review of more than \emph{500} publicly published datasets, including research domains like agriculture, land use and land cover, disaster monitoring, scene understanding, vision-language models, foundation models, climate change, and weather forecasting. We systematically analyze these EO datasets from four aspects: volume, resolution distributions, research domains, and the correlation between datasets. Based on the dataset attributes, we propose to measure, rank, and select datasets to build a new benchmark for model evaluation. Furthermore, a new platform for EO, termed EarthNets, is released to achieve a fair and consistent evaluation of deep learning methods on remote sensing data. EarthNets supports standard dataset libraries and cutting-edge deep learning models to bridge the gap between the remote sensing and machine learning communities. Based on this platform, extensive deep-learning methods are evaluated on the new benchmark. The insightful results are beneficial to future research. The platform and dataset collections are publicly available at \url{https://earthnets.github.io/}.
\end{abstract}

\begin{IEEEkeywords}
Dataset review, deep learning, Earth observation, foundation models, remote sensing
\end{IEEEkeywords}

\section{Introduction}
Earth Observation (EO) aims to monitor and assess the status of the Earth's surface using various Remote Sensing (RS) technologies \cite{toth2016remote,zhu2017deep}. EO can significantly contribute to our ability to better understand and analyze the planet Earth using RS data. The research in EO has been successfully applied to urban planning\cite{shaker2019automatic}, natural resources management\cite{bauer2020remote}, agriculture\cite{wojtowicz2016application}, food security \cite{karthikeyan2020review} and disaster monitoring \cite{joyce2009remote,van2000remote}. All these applications are important for the sustainable development of human society.

\begin{figure*}
	\begin{center}
		\includegraphics[width=0.99\textwidth]{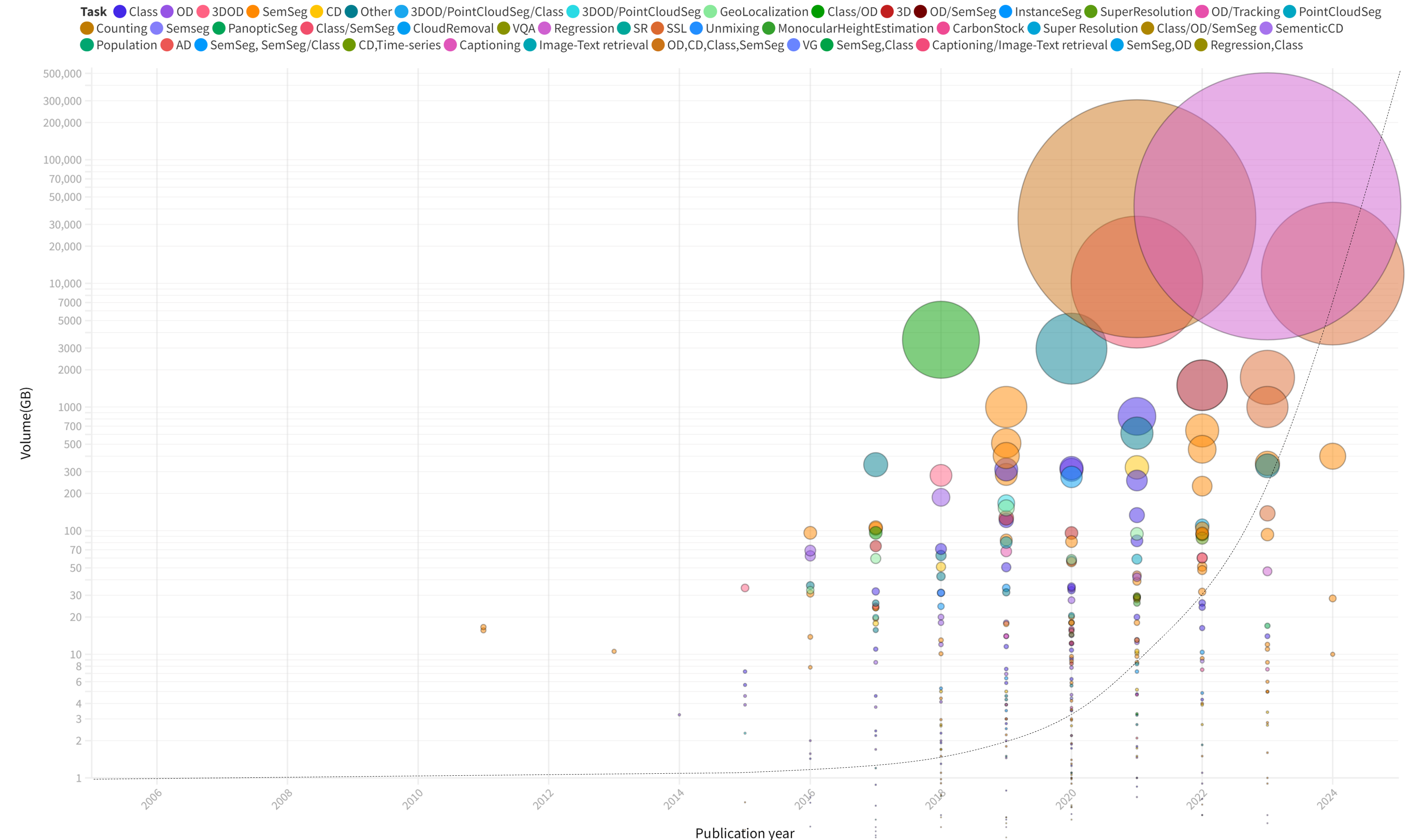}
	\end{center}
	\caption{Chronological overview of the volumes (in logarithmic scale) of over 500 existing datasets. As can be seen, increasingly numerous and larger datasets have been constructed and published over time. (Best viewed by zooming in.)}
	\label{year-vol}
\end{figure*}

With the development of EO technology, more and more satellites with diverse imaging sensors have been launched for different missions. A huge amount of RS data with global coverage and high resolution is received every day for automatic processing and analysis. To deal with large-scale data, deep learning techniques \cite{krizhevsky2017imagenet} have been proven effective for different research areas. In this context, recent RS datasets are constructed with larger and larger volumes of data. In Fig. \ref{year-vol}, we show a chronological overview of the volumes of more than 500 existing datasets. This figure shows that more numerous and larger datasets have been constructed and published during the past decade. Although considerable progress has been made with the overwhelming success of deep learning techniques \cite{zhang2016deep,xiong2018ai}, there are still problems requiring more research efforts to handle.

There is a lack of comprehensive dataset review for EO tasks. Owing to the efforts made by EO researchers, there are numerous datasets in the RS community with different modalities, resolutions, and application domains. Some of these RS data modalities include optical (RGB), Synthetic Aperture Radar (SAR), multispectral, hyperspectral, and point cloud data. Regarding the application domains, the published datasets may be designed for Land Use and Land Cover (LULC) \cite{marchisio2021rapidai4eo}, change monitoring\cite{hecheltjen2014recent}, disaster monitoring\cite{rahnemoonfar2021floodnet}, scene recognition, semantic segmentation, ground object detection, object tracking, agriculture, climate change, and weather forecasting. A comprehensive review of the RS datasets can provide researchers with a holistic view of the status of the research community. Bakula et al. \cite{bakula2019review} review benchmarking in photogrammetry and remote sensing related to geodata. They provide dataset collections and a bibliographic analysis, which are useful for future research. Schmitt et al. \cite{schmitt2021there,schmitt2023there} provide a historic review of RS datasets. They discuss dataset features based on a few examples and present some important criteria for the establishment of a standard database. Although a few works \cite{long2021creating, li2020object, abdollahi2020deep} attempt to review existing RS datasets, they are not sufficient and comprehensive to cover a large range of research domains, like datasets for vision-language models and RS foundation models.

There is no systematic summary and analysis of RS datasets. The ever-growing quantity of RS datasets makes it difficult to find the proper one for a specific application in the jungle of remote sensing datasets. For example, there are more than 20 datasets related to building extraction \cite{tomljenovic2015building}. Finding, assessing, and selecting the one most suitable for a given application can be laborious and time-consuming. Thus, it is vitally important to summarize and categorize RS datasets to provide valuable guidance and reference for researchers. Both the EOD\footnote{\url{https://eod-grss-ieee.com/dataset-search}} \cite{schmitt2022eod} and the AiTLAS Semantic Data Catalog\footnote{\url{http://eodata.bvlabs.ai/##/}} have been designed as search engines for RS datasets, and are valuable and helpful for the RS community. However, the information about RS datasets collected in these existing databases is still limited. Beyond summarization and categorization, insightful analysis of existing datasets can help researchers understand the current research state and trends of the whole community. Systematic analysis of RS datasets is thus crucial to the future development of different research fields.

There is no unified benchmark for fair comparisons of remote sensing methods. In the Computer Vision (CV) community, a large-scale dataset like ImageNet~\cite{deng2009imagenet} is usually used to evaluate newly developed deep learning models. Compared with small-scale datasets, large-scale datasets with rich semantic annotations align better with complex real-world scenarios \cite{long2021creating}. Thus they can be more reliable for performance validation and comparison of deep learning algorithms. Although several large-volume RS datasets have been published \cite{zhu2019so2sat,christie2018functional, kossmann2022seasonet}, many of the currently developed methods are still evaluated on small-scale datasets. However, datasets with a small scale or limited geographic coverage may be biased to a specific data distribution that is not representative of real-world scenarios. Moreover, many RS datasets are published with no standard train/validation/test splits. This increases uncertainty during the evaluation of algorithms. Thus, it is urgent to build new RS benchmarks to enable a fair comparison of different methods. 

Recently, foundation models in RS have demonstrated significant achievements across various EO applications~\cite{jakubik2023foundation}. The key to further enhancing these models involves a thorough investigation of the relationships between existing RS datasets. Such an investigation is important for assembling similar datasets to form a more comprehensive large-scale data archive. This amalgamation of analogous datasets is crucial for effectively pre-training and evaluating large-scale RS foundation models. Despite its importance, the relationships among various RS datasets remain insufficiently explored in current studies.

Currently, there is a lack of an open platform for different EO tasks. For deep learning methods, backbone networks, hyper-parameters, and training tricks are influential factors that should be considered for fair performance comparison. However, existing works usually evaluate the performance with different dataset splits, which makes it difficult to fairly and reliably compare different algorithms. Due to the large variance in data collection sensors and pre-processing pipelines, it is non-trivial to adapt modern deep learning models to RS datasets \cite{stewart2021torchgeo}. As a result, many cutting-edge and off-the-shelf deep learning methods from the machine learning community are not evaluated and compared on RS data.

To tackle the previously mentioned issues, in this study, we first make an exhaustive and comprehensive review of the publicly accessible RS datasets. Next, a systematic analysis is undertaken based on the information about these datasets. Based on the attribute information, we filter, rank, and select five large-scale datasets designed for general purposes to build a new benchmark for model evaluation. To further enable a fair and reproducible comparison of different algorithms, we construct a new deep learning platform, EarthNets, as a foundation for future work. Our main contributions are summarized below. 

\begin{enumerate}
    \item We review over 500 RS datasets and summarize them into different research tasks and domains. Beyond summarization and categorization, EarthNets provides insightful analysis of existing datasets that can help researchers understand the current research state and trends of the whole community.
    
    \item Systematic analyses are made concerning four aspects of these datasets to provide insights for future research within the RS community. Specifically, the volume, resolution distributions, research domains, and dataset relationship are considered for a comprehensive dataset analysis.
    
    \item To our knowledge, we are the first to measure and rank existing RS datasets using the dataset attributes provided in this study. In addition, we provide the affinity relationships between existing datasets, which is useful in assembling similar datasets to support the training of large-scale RS foundation models.
    
    \item We release an open platform, EarthNets, for EO tasks. EarthNets aims to enable fair comparisons, efficient development of methods, and the greater availability of the RS data to a larger research community.
\end{enumerate}
The rest of this paper is organized as follows. Section \ref{Review} reviews existing RS datasets for different tasks. Section \ref{ANA} presents the analyses of the reviewed datasets from four different perspectives. Section \ref{RANK} introduces the proposed dataset ranking, selection method, and the benchmark-building process. Section \ref{ETN} describes the newly released EarthNets platform. Section \ref{EXP} presents the benchmarking results and analysis on the five selected RS datasets. Section \ref{CONC} concludes the paper.
\begin{table*}[htbp]
\centering
\caption{Detailed Information for some representative RS Image Classification Datasets. These datasets are grouped into 27 different research domains in alphabetical order. Note that / denotes the missing information and MS denotes the multispectral data. The complete information for these datasets can be found at \url{https://earthnets.github.io}.}
\small
\label{RSICLS}
\scalebox{0.64}{
\begin{tabular}{ccccccccc}
\toprule
Domain                           & Name                            & Year & \#Samples  & Sample Size         & \#Classes & Modailty                       & Resolution        & Vol.(GB) \\  \midrule
\multirow{13}{*}{Agriculture}    & Brazilian Coffee Scene \cite{Braziliancoffe}         & 2015 & 2876       & 64                  & 2         & RGB                            & 20m               & 0.004      \\
                                 & Indian Pines~\cite{baumgardner2015220}                    & 2015 & 1          & 145                 & 16        & Hyperspectral                  & 20m               & 0.0059         \\
                                 & Salinas~\cite{Hyperspectral}                         & 2015 & 1          & 365                 & 16        & Hyperspectral                  & 3.7m              & 0.026    \\
                                 & Crop Type Mapping Ghana~\cite{m2019semantic}         & 2019 & /          & /                   & 18        & Sentinel-1,Sentinel-2,Planet   & 3$\sim$10m        & 312.54     \\
                                 & CV4A Kenya~\cite{kerner2020field}                      & 2020 & 4688       & 2016x3035           & 7         & Sentinel-2                     & 10m               & 3.5      \\
                                 & BreizhCrops~\cite{breizhcrops2020}                     & 2020 & 610000     & /                   & 9         & Sentinel-2,MT                  & 10$\sim$60m       & /        \\
                                 & CaneSat~\cite{vzbn-qj64-20}                         & 2020 & 1627       & 10                  & 2         & Sentinel-2,MT                  & 10m               & 0.006    \\
                                 & CropHarvest~\cite{tseng2021cropharvest}                     & 2021 & 90,480     & 12 ts               & 343       & Sentinel-1,Sentinel-2,ERA5,DEM & 10$\sim$60m       & 20       \\
                                 & South Africa Crop Type~\cite{South}          & 2021 & 122736     & /                   & 9         & Sentinel-1,Sentinel-2          & /                 & 82.77    \\
                                 & DENETHOR~\cite{kondmann2021denethor}                        & 2021 & /          & /                   & 9         & Sentinel-1,Sentinel-2          & 3m                & 254.5    \\
                                 & The Canadian Cropland\cite{Canadian}          & 2022 & 78536      & 64                  & 10        & Sentinel-2                     & 10m               & 26    \\
                                 & Space2Ground\cite{Space-to-Ground}                    & 2022 & 10102      & 260                 & 2         & Sentinel-1,Sentinel-2,RGB      & 10$\sim$60m       & 0.501    \\
                                 & Sen4AgriNet\cite{sykas2021sen4agrinet}                     & 2021 & 225000     & 366                 & 158       & Sentinel-2                     & 10$\sim$60m       & 10240    \\  \midrule
\multirow{2}{*}{Aircraft}        & SAR-ACD\cite{SAR-ACD}                         & 2022 & 4322       & 64                  & 20        & SAR      & /          & /     \\ 
                                 & MAR20~\cite{wenqi2024mar20} & 2023  &3842   & 800$\times$800     & 20  & RGB &1m   & 1.1 \\ \midrule
\multirow{4}{*}{Cloud}
                                & SPARCS \cite{hughes2014automated}                         & 2016 & 80         & 1000                & 7         & MS,Landsat                     & 30m               & 1.43   \\
                                 & Kaggle Cloud Detection\cite{kagglecloud}         & 2019 & 9244       & 1750                & 4         & RGB                            & /                 & 5.86    \\
                                 & CloudCast\cite{nielsen2020cloudcast}                       & 2020 & 70080      & 1229                & 10        & NWP                            & 3km               & 320.31     \\
                                 & Sentinel-2 Cloud Mask Catalogue\cite{cmc} & 2020 & 513        & 1022                & 3         & Sentinel-2                     & 20m               & 15.38    \\  \midrule
Event                            & ERA\cite{eradataset}                             & 2020 & 343680     & 640                 & 25        & RGB Video                      & /                 & 6.3   \\  \midrule
\multirow{5}{*}{Flood}           & Hurricane Damage\cite{cao2020building}                & 2019 & 16000      & 128                 & 2         & RGB                            & 1m                & 0.064  \\
                                 & SEN-12-FLOOD\cite{rambour2020flood}                    & 2020 & 336        & 512                 & 2         & RGB,SAR,MS                     & 10m               & 12.2    \\
                                 & Sen1Floods11\cite{bonafilia2020sen1floods11}                    & 2020 & 4831       & 512                 & 1         & Sentinel-1                     & 10m               & 14.3     \\
                                 & OMBRIA \cite{drakonakis2022ombrianet}                          & 2022 & 3376       & 256                 & 2         & Sentinel-1,Sentinel-2          & 10m$\sim$20m      & 0.19     \\
                                 & FloodNet\cite{rahnemoonfar2021floodnet}                        & 2021 & 2343       & 4000                & 9         & RGB                            & 0.015m            & 2.1      \\ \midrule
Forest                           & Kaggle Planet Forest\cite{chandak2017understanding}            & 2017 & 150000     & 256                 & 17        & RGB-NIR                        & 5m                & 32.23  \\   \midrule
\multirow{16}{*}{General Scenes} & OVERHEAD MNIST\cite{noever2021overhead}  & 2020 & 1000       & 28                  & 9         & Grayscale                      & /                 & 0.017  \\
                                 & fMoW\cite{christie2018functional}        & 2018 & 523846     & /                   & 63        & RGB,MS                         & 0.3m              & 3500     \\
                                 & UC Merced\cite{UCMerced}                      & 2010 & 2100       & 256                 & 21        & RGB                            & 0.3m              & 0.3    \\
                                 & WHU-RS19\cite{Xia2010WHURS19}                        & 2012 & 1013       & 600                 & 19        & RGB                            & 0.5m              & 0.1    \\
                                 & RSSCN7\cite{zou2015deep}                          & 2015 & 2800       & 400                 & 7         & RGB                            & /                 & 0.086  \\
                                 & NWPU-RESISC45\cite{cheng2017remote}                   & 2016 & 31500      & 256                 & 45        & RGB                            & 0.2$\sim$30m      & 0.404    \\
                                 & RSC11~\cite{zhao2016feature}                           & 2016 & 1232       & 512                 & 11        & RGB                            & /                 & 0.63      \\
                                 & AID~\cite{xia2017aid}                            & 2017 & 10000      & 600                 & 30        & RGB                            & 3m                & 2.4    \\
                                 & RSD46-WHU~\cite{xiao2017high}                       & 2017 & 117000     & 256                 & 46        & RGB                            & 0.5$\sim$2m       & 11    \\
                                 & PatternNet~\cite{zhou2018patternnet}                     & 2018 & 30,400     & 256                 & 38        & RGB                            & 0.062$\sim$4.693m & 1.3     \\
                                 & OPTIMAL-31~\cite{wang2018scene}                      & 2019 & 1860       & 256                 & 31        & RGB                            & /                 & 0.024  \\
                                 & MLRSNet\cite{qi2020mlrsnet}                         & 2020 & 109,161    & 256                 & 46        & RGB                            & 0.1$\sim$10m      & 1.254  \\
                                 & CLRS\cite{li2020clrs}                            & 2020 & 15000      & 256                 & 25        & RGB                            & 0.26$\sim$8.85m   & 1.735   \\
                                 & Million AID\cite{long2021creating}                     & 2021 & 1,000,000  & 150$\sim$550        & 28        & RGB                            & 0.5$\sim$153m     & 133.5    \\
                                 & NaSC-TG2\cite{zhou2021nasc}                        & 2021 & 20000      & 256                 & 10        & RGB-NIR                        & 100m              & /      \\
                                 & Satellite Image Classification \cite{sic}  & 2021 & 5631       & 256                 & 4         & RGB                            & /                 & 0.023  \\  \midrule
Multi-label Scenes               & MultiScene\cite{hua2021multiscene}                     & 2021 & 100,000    & 512                 & 36        & RGB                            & 0.3$\sim$0.6m     & 0.85   \\  \midrule
Geophysical                      & Hephaestus\cite{bountos2022hephaestus}                      & 2022 & 216106     & 224                 & 6         & InSAR                          & /                 & 93.71 \\  \midrule
Golf Course                      & MUSIC4GC (Golf Course)\cite{uehara2017object}        & 2017 & 83431      & 16                  & 2         & MS,Landsat                     & 30m               & 0.37  \\  \midrule
Hot Area                         & MUSIC4HA (Hot Area)\cite{uehara2017object}             & 2022 & 2511       & 16                  & 6         & Sentinel-2                     & 10m               & 0.01  \\  \midrule
Iceberg                          & Iceberg Detection\cite{iceberg}               & 2018 & 10028      & 75                  & 2         & SAR                            & /                 & 0.295  \\  \midrule
\multirow{10}{*}{Land Cover}     & SAT-4\cite{basu2015deepsat}                           & 2015 & 500000     & 28                  & 4         & RGB-NIR                        & 1$\sim$6m         & 7.25  \\
                                 & SAT-6\cite{basu2015deepsat}                          & 2015 & 405000     & 28                  & 6         & RGB-NIR                        & 1m                & 5.65   \\
                                 & Botswana~\cite{Hyperspectral}                        & 2015 & 1          & 875                 & 14        & Hyperspectral                  & 30m               & 0.077    \\
                                 & TiSeLaC~\cite{Tiselac}                        & 2017 & 23         & 2866x2633           & 9         & RGB-NIR,MT                     & 30m               & /     \\
                                 & Gaofen Image Dataset (GID) \cite{GID2020}      & 2018 & 150        & 7200                & 15        & RGB-NIR                        & 4m                & 71.1   \\
                                 & MSLCC~\cite{bahmanyar2018multisensor}                           & 2018 & 2          & 5596x6031,8149x5957 & 4         & SAR,MS                         & 10m               & 0.5      \\
                                 & BigEarthNet~\cite{sumbul2019bigearthnet}                    & 2019 & 590326     & 120                 & 43        & Sentinel-1,Sentinel-2          & 10m, 20m, 60m     & 121      \\
                                 & Slovenia Land Cover~\cite{Slovenia}             & 2019 & 940        & 500                 & 10        & Sentinel-2                     & 10m               & 11.55  \\
                                 & So2Sat LCZ42~\cite{zhu2019so2sat}                    & 2019 & 400673     & 32                  & 17        & Sentinel-1,Sentinel-2          & 10m               & 50.59     \\
                                 & TG1HRSSC~\cite{TG1HRSSC}                      & 2021 & 204        & 512                 & 9         & Hyperspectral                  & 5m, 10m, 20m      & 0.277   \\   \midrule
\multirow{9}{*}{Land Use}        & SIRI-WHU (Google+USGS)~\cite{zhong2015scene}        & 2016 & 2400       & 200                 & 12        & RGB                            & 2m                & 0.7   \\
                                 & RSI-CB256~\cite{li2020RSI-CB}                      & 2017 & 24000      & 256                 & 35        & RGB                            & 0.3$\sim$3m       & 2.2   \\
                                 & RSI-CB128~\cite{li2020RSI-CB}                      & 2017 & 36000      & 128                 & 45        & RGB                            & 0.3$\sim$3m       & 0.88   \\
                                 & Austin Zoning~\cite{Austin}                  & 2017 & 3,666      & 773x961             & 5         & RGB                            & /                 & 0.596  \\
                                 & HistAerial~\cite{ratajczak:hal-02003932}                      & 2019 & 42000      & 25,50,100           & 7         & Grayscale                      & /                 & 7.6      \\
                                 & AiRound~\cite{machado2020airound}                        & 2020 & 11753      & 300                 & 11        & RGB,Sentinel-2,Ground,Aerial   & /                 & 33       \\
                                 & CV-BrCT~\cite{machado2020airound}                         & 2020 & 24000      & 500                 & 9         & RGB                            & /                 & 9.2     \\
                                 & EuroSAT~\cite{helber2019eurosat}                         & 2018 & 27000      & 64                  & 10        & Sentinel-2                     & 10m               & 1.92     \\
                                 & SenseEarth classify~\cite{SENCLY}            & 2020 & 70000      & 100$\sim$12655      & 51        & RGB                            & 0.2$\sim$153m     & 10.8   \\   \midrule
Landslide                        & Bijie Landslide\cite{ji2020landslide}                 & 2020 & 2773       & 200                 & 2         & RGB                            & 0.68m             & 0.51   \\  \midrule
Military                         & MSTAR-8class\cite{ranjan2002classification}                    & 1996 & 9466       & 368                 & 8         & SAR                            & 0.3m              & 0.444   \\  \midrule
\multirow{5}{*}{Plant/Tree}      & TalloS~\cite{bountos2024fomobench}  & 2024  &190,481 & /    & 1160  &                                  SAR,MS   & 10m   & / \\
                                 & WiDS Datathon 2019\cite{Datathon}              & 2019 & 11000      & 256                 & 2         & RGB                            & 3m                & 0.46    \\
                                 & The Auto Arborist Dataset\cite{beery2022auto}       & 2022 & 2,637,208  & 1,024               & 344       & RGB,MS                         & /                 & 24    \\
                                 & TreeSatAI\cite{ahlswede2022treesatai}                       & 2022 & 50381      & 304x304,6x6         & 47        & Sentinel-1,Sentinel-2,RGB      & 10m,0.2m          & 16.3     \\
                                 & Forest Damages Larch Casebearer\cite{Casebearer} & 2021 & 1543       & 1500                & 5         & RGB                            & UAV               & 3.3   \\   \midrule
Sea Lion                         & NOAA Sea Lion Population Count\cite{NOAA}  & 2017 & 950        & 4900                & 4         & RGB                            & /                 & 96 \\  \midrule
\multirow{5}{*}{Ship}            & Ships in Satellite Imagery\cite{shipsi}      & 2017 & 4000       & 80                  & 2         & RGB                            & 3m                & 0.343 \\
                                 & MASATI\cite{Gallego2018}                          & 2018 & 7389       & 512                 & 7         & RGB                            & 0.08$\sim$2m      & 2.3    \\
                                 & DSCR\cite{di2019public}                           & 2019 & 20,675     & 150$\sim$800        & 7         & RGB                            & /                 & /     \\
                                 & FGSCR-42\cite{di2019public}                        & 2021 & 9320       & 140$\sim$800        & 42        & RGB                            & /                 & 4.76   \\
                                 & SynthWakeSAR\cite{rizaev2022synthwakesar}                    & 2022 & 46080      & 96000               & 10        & SAR                            & 3.3m              & 4.3     \\   \midrule
Smoke                            & USTC\_SmokeRS \cite{ba2019smokenet}                  & 2019 & 6225       & 256                 & 6         & RGB                            & 1000m             & 0.79  \\   \midrule
Solar Power Plants               & MUSIC4P3\cite{imamoglu2017solar}                       & 2017 & 1280000    & 16                  & 2         & MS,Landsat                     & 30m               & 4.6   \\   \midrule
Species                          & GeoLifeCLEF 2021\cite{cole2020geolifeclef}                & 2021 & 19,000,000 & 256                 & 31,435    & RGB-IR,MS,LC,DEM               & 1m,0.3m,0.1m      & 840  \\   \midrule
Tailings Dam                     & BrazilDAM\cite{ferreira2020brazildam}                       & 2020 & 769        & 384                 & 2         & RGB                            & 10$\sim$60m       & 57     \\   \midrule
Urban Village                    & S2UC \cite{chen2022multi}                           & 2021 & 1714       & 224                 & 2         & RGB                            & 2m                & 1.8   \\   \midrule
\multirow{2}{*}{Vegetation}      & Kennedy Space Center\cite{Kennedy}            & 2015 & 1          & 550                 & 13        & Hyperspectral                  & 0.18m             & 0.055  \\
                                 & Brazilian Cerrado-Savanna \cite{nogueira2016towards}      & 2016 & 1311       & 64                  & 4         & MS                             & 5m            & 0.011   \\   \midrule
\multirow{3}{*}{Vehicles}        & WAMI DIRSIG \cite{uzkent2017tracking}                    & 2017 & 55226      & 64                  & 2         & Hyperspectral                            & 0.3m              & 0.33   \\
                                 & Kaggle Find a Car Park\cite{carpark}          & 2019 & 3262       & 1296                & 2         & RGB                            & /                 & 2.75   \\
                                 & MAFAT-Fine-Grained \cite{dahan2021cofga}                & 2021 & 4216       & /                   & 37        & RGB                            & 0.05$\sim$ 0.15m   & /     \\    \midrule
Wind                             & Airbus Wind Turbine Patches\cite{Turbines}     & 2021 & 155,000    & 128                 & 2         & RGB,MS                         & 1.5m              & 1    \\ \bottomrule
\end{tabular}}
\end{table*}

\section{Remote Sensing Dataset Review}
\label{Review}
In this section, we review and organize over 500 existing public RS datasets into five parts concerning their tasks: image classification, object detection, semantic segmentation, change detection, vision-language understanding, and foundation models. For an in-depth analysis of these datasets, we collect as much detailed information as possible for each dataset. Compared with existing review articles, our work provides richer information on the attributes of the datasets. Specifically, the following 9 aspects are considered.

    \textbf{1) Research domain}. To present these reviewed datasets clearly, we organize them according to the specific domains in which they are created. Some typical research domains include agriculture, building, road, cloud, LULC, general scenes or objects, etc. 
    
    \textbf{2) Publication year}. We provide the year of publication of the dataset, which is useful for the chronological analysis of these datasets.
    
    \textbf{3) Number of samples}: To measure the dataset scale, we list the number of samples for each dataset. Note that it could be the number of images (for image-based datasets), the number of video clips (for video-based datasets), the number of points (for point cloud datasets), or the number of image pairs (for change detection datasets).
    
    \textbf{4) Size of sample}: The size of each sample in the dataset is also an important factor for measuring the dataset scale. For images, the size is the height and width. For datasets designed for 3D understanding, the sample size could be the covered area. For vision-language datasets, the size could be the image-text pairs.
    
    \textbf{5) Number of Classes}: For image classification, object detection, semantic segmentation, and vision-language tasks, we provide the number of classes annotated in each dataset.
    
    \textbf{6) Data modality}: A wide range of imaging sensors can be used to build datasets for different EO tasks. For example, some of the data sources of RS datasets may include optical images, multispectral images, hyperspectral images \cite{Hyperspectral}, SAR\cite{SAR-ACD}, point cloud \cite{Columbia_point_cloud}, and Digital Surface Model (DSM)~\cite{ISPRS-Contest1, ISPRS-Contest2}.
    
    \textbf{7) Resolution range}: The spatial resolutions of RS images are highly correlated with the image content. The resolution range is also highly relevant to specific EO tasks to which it can be applied.

    \textbf{8) Volume}: The volume of each RS dataset is also factored in as a measurement of the dataset scale.
    
    \textbf{9) Dataset link}: To facilitate the research, we provide the download link of each dataset. More detailed information can be found at \url{https://earthnets.github.io/}.

\subsection{RS Image Classification Datasets}
Image classification is a fundamental task in both the CV and RS communities. With image- or patch-level annotations, RS image classification has been employed in different real-world applications. Table \ref{RSICLS} presents some representative RS image classification datasets. To facilitate researchers to search and index, we organize them into different research domains alphabetically. To be specific, Table \ref{RSICLS} contains 13 agriculture-related datasets \cite{Braziliancoffe,kondmann2021denethor,sykas2021sen4agrinet,Canadian} that are constructed from the year 2015 to 2022. For agriculture-related applications, the images or patches in the datasets are labeled with different granularity levels, from binary labels (crop/non-crop) to fine-grained crop-type labels (up to 348 granular labels).
For general scene classification, 16 datasets are presented in the table. Million-AID, the largest of these, contains a million instances for training and evaluating scene classification methods. MLRSNet \cite{qi2020mlrsnet}, RSD46-WHU \cite{xiao2017high}, and NWPU-RESISC45 \cite{cheng2017remote} are annotated with more than 40 class labels. These datasets are also widely used for evaluating RS foundation models. There are 19 datasets for LULC applications in the table. Among these datasets, multiple types of data sources are used, including hyperspectral \cite{Hyperspectral}, multispectral \cite{sumbul2019bigearthnet,zhu2019so2sat,machado2020airound,helber2019eurosat}, SAR \cite{bahmanyar2018multisensor} and RGB data \cite{zhong2015scene, li2020RSI-CB}. 

There are 5 ship-related \cite{shipsi,Gallego2018,di2019public,rizaev2022synthwakesar} and 5 flood-related \cite{cao2020building,rambour2020flood,bonafilia2020sen1floods11} datasets for RS image classification task. For the cloud-related research domain, 4 datasets are reviewed in this table. Some specific domains with fewer datasets are also presented, like smoke \cite{ba2019smokenet}, sea lion \cite{NOAA}, solar power plants \cite{imamoglu2017solar} and wind datasets \cite{Turbines}. It is worth noting that species classification datasets are annotated with the most semantic labels.

Compared with object detection (with object-level annotation) and pixel-level segmentation tasks, agriculture-related applications are mainly modeled as image classification tasks. In contrast, aircraft \cite{SAR-ACD} and ship-related datasets are built primarily for object detection tasks. For agriculture-related datasets, data from the Sentinel satellites is mostly used with a lower spatial resolution to cover larger crop areas. General scene classification and LULC are dominant domains in RS image classification datasets. 

\subsection{RS Object Detection Datasets}
Object detection is closely related to real-world applications like autonomous driving, video surveillance, and many other high-level scene understanding tasks. Thus, several widely-read works have been published in the CV community, like Faster RCNN \cite{ren2015faster}, SSD \cite{liu2016ssd}, YOLO \cite{redmon2016you}, and Transformer-based detectors \cite{carion2020end}. For the RS object detection task, more and larger datasets are also being published for different EO applications, including aircraft detection \cite{CASIA-aircraft}, building detection \cite{DeepGlobe18,USBuilding,kagglebuilding1,weir2019spacenet}, ship detection \cite{huang2017opensarship,li2017ship,AirbusSDC,ShipsGoogleEarth}, vehicle detection \cite{heitz2008learning,razakarivony2016vehicle,mundhenk2016large,DLR3K,yang*18:icip}, general ground object detection \cite{cheng2016survey,long2017accurate,zhang2019hierarchical,xview2018,fmow2018,Xia_2018_CVPR} and other research domains \cite{HAN2022102966,dstl10bands,AirbusOilStorage}.

Object detection from satellite and Unmanned Aerial Vehicles (UAV) perspectives is crucial for applications such as traffic analysis, wildlife protection, and search and rescue operations. Consequently, numerous datasets have been developed for tasks like object counting, person detection, ship detection, and vehicle detection.
Table \ref{RSIOD} presents some representative RS object detection datasets organized into 18 different research domains. Some popular research domains for RS object detection tasks are general object detection with 12 datasets, building detection with 13 datasets, aircraft detection with 7 datasets, ship detection with 14 datasets, and vehicle-related object detection with 20 datasets. Since object detection is a task with object-level annotations, the spatial resolutions of these datasets are usually higher than those of image classification datasets. However, for objects with large sizes, like ships, the data from Sentinel-1 and Sentinel-2 satellites are also used \cite{LS_SSDD,wei2020hrsid}. For the detection of traffic objects \cite{AI-TOD_2020_ICPR} or other small objects \cite{cheng2022towards}, images captured from UAV are usually used.

\subsection{RS Semantic Segmentation Datasets}
Pixel-level semantic segmentation aims to assign semantic labels to each image pixel. Compared with image-level and object-level tasks, interpreting the image data with semantic maps can provide a more complete understanding of the scene at a fine-grained level. Table \ref{RSISEG} presents some representative datasets for RS semantic segmentation tasks.
Given the close association of agriculture, buildings, roads, LULC, and urban scenes with essential applications such as urban growth, urban planning, and city living, these areas emerge as primary domains for semantic segmentation tasks. 

Among them, 21 datasets are built for LULC segmentation tasks \cite{WDCmall,kemker2018algorithms,boguszewski2021landcover,schmitt2019sen12ms,kaggleaerial,wang2022air,ISPRS-Contest1,ISPRS-Contest2,kossmann2022seasonet}. There are 17 additional datasets constructed for the segmentation of general scenes. 
\begin{table*}[]
\centering
\caption{Detailed Information for some representative RS Object Detection Datasets. These datasets are grouped into 18 different research domains in alphabetical order. Note that / denotes the missing information and MS denotes the multispectral data. The download links for all these datasets can be found at \url{https://earthnets.github.io}.}
\small
\scalebox{0.65}{
\begin{tabular}{ccccccccc}
\toprule
Domain                             & Name                                 & Year & \# samples  & Size                  & \# classes & Modality            & Resolution      & Volume (GB) \\ \midrule
Agriculture                        & PASTIS \cite{garnot2021panoptic}                              & 2021 & 2433        & 128                   & 18         & Sentinel-2          & 10m             & 29     \\ \midrule
\multirow{7}{*}{Aircraft}          & MAR20~\cite{wenqi2024mar20} & 2023  &3842   &                                                800$\times$800     & 20  & RGB &1m   & 1.1 \\
                                   & MTARSI (Aircraft) \cite{zenodoMTARSII}                    & 2019 & 9385        & 256                   & 2          & RGB                 & /               & 0.48     \\
                                   & RarePlanes \cite{rareplanes}                           & 2020 & 713348      & 512                   & 110        & MS,WorldView3       & 0.3$\sim$1.5m   & 310.55        \\
                                   & CGI Planes \cite{kaggleCGI}         & 2021 & 500         & /                     & 2          & RGB                 & /               & 0.7           \\
                                   & CASIA-aircraft \cite{CASIA-aircraft}                      & 2021 & 58,121      & 399          & 2          & RGB                 & /               & /           \\
                                   & Airbus Aircraft Detection \cite{Airbus_Aircraft_Detection}            & 2021 & 109         & 2560                  & 2          & RGB                 & 0.5m            & 0.092     \\
                                   & SAR Aircraft \cite{9761751}         & 2022 & 2966        & 224                   & 2          & SAR                 & 0.5m$\sim$3m    & 0.18      \\ \midrule
Bridge                             & Bridges Dataset \cite{nogueira2016pointwise}                     & 2019 & 500         & 4800x2843             & 2          & RGB                 & 0.5m            & 1.45    \\ \midrule
\multirow{13}{*}{Building}         & SpaceNet-4 (Multi-View)  \cite{weir2019spacenet}           & 2018 & 60000       & 900                   & 1          & MS,WorldView2       & 0.3m            & 186       \\
                                   & DeepGlobe (Building)  \cite{DeepGlobe18}               & 2018 & 24586       & 650                   & 2          & Panchromatic,RGB,MS & 0.5m            & /         \\
                                   & WHU Building  \cite{ji2018fully}     & 2018 & 25577       & 512                   & 2          & RGB                 & 0.3m            & 24.41    \\
                                   & CrowdAI Mapping  \cite{mohanty2020deep}                   & 2018 & 401,755     & 300                   & 1    & RGB                 & /               & 5.3     \\
                                   & Map Challenge \cite{mohanty2020deep}                      & 2018 & 341,058     & 300                   & 2          & RGB                 & /               & /     \\
                                   & TBF~\cite{TBF}        & 2018 & 13          & 40,000                & 2          & RGB                 & /               & /      \\
                                   & Microsoft Building (Australia) \cite{AustraliaBuilding}   & 2019 & 11,334,866  & /    & 2          & /                & /               & 6.4    \\
                                   & Microsoft Building (Uganda/Tanzania) \cite{UgandaTanzaniaBuilding} & 2019 & 17,942,345  & /                     & 2          & /                   & /               & 3.5    \\
                                   & Microsoft Building (USA) \cite{USBuilding}            & 2019 & 129,591,852 & /                     & 2          & /                   & /               & 34.4    \\
                                   & Microsoft Building (Canada) \cite{USCanada}         & 2019 & 11,842,186  & /                     & 2          & /                   & /               & 2.5    \\
                                   & Urban Building Classification \cite{UBCBuilding}       & 2022 & 800         & 600                   & 61         & RGB                 & 0.5$\sim$0.8m   & 0.675    \\
                                   & BONAI ~\cite{wang2022bonai}                              & 2022 & 3,300       & 1,024                 & 1          & RGB                 & 0.3m$\sim$0.6m  & 4.86  \\
                                   &MSNet~\cite{zhu2021msnet} &2020 &1030 & \ &3  & RGB  & \ & 0.43 \\
                                   \midrule
\multirow{13}{*}{General}          & NWPU-VHR10 \cite{cheng2016survey}    & 2014 & 800         
                                   & 1000                  & 10         & RGB,IRRG            & 0.08$\sim$2m    & 0.07   \\
                                   & RSOD  \cite{long2017accurate}                             & 2017 & 976         & 1000                  & 4          & RGB                 & 0.3$\sim$3m     & 0.077    \\
                                   & TGRS HRRSD \cite{zhang2019hierarchical}                   & 2017 & 21761       & 10569                 & 13         & RGB                 & 0.15$\sim$1.2 m & 8.6          \\
                                   & xView \cite{xview2018}       & 2018 & 1,413       & 3,000 & 60         & RGB,MS              & 0.3m            & 20     \\
                                   & fMoW  \cite{fmow2018}     & 2018 & 523846      & /        & 63         & RGB,MS              & 0.3m            & 3500 \\
                                   & DOTA v1.0 \cite{Xia_2018_CVPR}                           & 2018 & 2806        & 4000                  & 15         & RGB                 & /               & 18     \\
                                   & DOTA v1.5 \cite{Ding_2019_CVPR}                           & 2019 & 2806        & 4000                  & 16         & RGB                 & /               & 18      \\
                                   & DIOR  \cite{li2020object}                              & 2019 & 23463       & 800                   & 20         & RGB                 & 0.5$\sim$30 m   & 6.93    \\
                                   & DOTA v2.0  \cite{9560031}                          & 2020 & 11268       & 4000                  & 18         & RGB                 & 0.1$\sim$0.81   & 34.3       \\
                                   & iSAID \cite{waqas2019isaid}                                & 2020 & 2806        & 4000                  & 15         & RGB                 & /               & 18     \\
                                   & VALID \cite{valid2020}                                & 2020 & 6690      & 1024          & 30        & RGBD                       & /                & 15.7    \\
                                   & UAVOD10 \cite{HAN2022102966}                             & 2022 & 844         & 1000$\sim$4800        & 10         & RGB                 & 0.15m           & 0.9 \\  \midrule
Human/Animals                      & BIRDSAI \cite{bondi2020birdsai}                             & 2020 & 162000      & 640                   & 10         & Thermal             & UAV@60-120m     & 3.7   \\ \midrule
Land Covers                        & Dstl Satellite Imagery \cite{dstl10bands}              & 2017 & 57          & 3,348                 & 10         & RGB,MS              & 0.3m$\sim$7.5m  & 21.7    \\ \midrule
Object Counting                    & RSOC (Object Counting) \cite{gao2020counting}              & 2020 & 3057        & 2500                  & 4          & RGB                 & /               & 0.082     \\ \midrule
\multirow{3}{*}{Oil Storage Tanks} & Oil and Gas Tank (OGST) \cite{OGST}             & 2020 & 10000       & 512                   & 2          & RGB                 & 0.3m            & 1.87  \\
                                   & Airbus Oil Storage Detection \cite{AirbusOilStorage}        & 2021 & 103         & 2560                  & 2          & MS                  & 1.2m            & 0.102   \\
                                   & Oil Storage Tanks \cite{OilStorage}                   & 2019 & 10000       & 512                   & 2          & RGB                 & 0.5m            & 3     \\  \midrule
Volcanoes                          & Hephaestus \cite{Bountos_2022_CVPR}                          & 2022 & 216106      & 224                   & 6          & InSAR               & /               & 93.71    \\  \midrule
\multirow{2}{*}{Person}                             & Semantic Drone-OD \cite{SDD}                   & 2019 & 400         & 5000                  & 2          & RGB                 & /               & 3.91     \\ 
                        & Stanford Drone  \cite{robicquet2020learning}                     & 2016 & 100         & 1400x1904             & 6          & RGB Video           & 0.025m       & 69  \\  \midrule
\multirow{4}{*}{Sea}                                & NOAA Sea Lion Count \cite{NOAA}                 & 2017 & 950         & 4,900                 & 4          & RGB                 & /               & 96    \\ 
                                   & Aerial Maritime Drone \cite{Jacob_AMDD}               & 2020 & 508         & 800x600               & 5          & RGB                 & /               & 0.038    \\
                                   & SeaDronesSee \cite{varga2022seadronessee}                        & 2022 & 5630        & 3,840$\sim$5,456      & 6          & RGB                 & /               & 60.3  \\  
                                   & AFO-Floating objects \cite{Gasienica_AFO}                & 2020 & 3647        & 720$\sim$3840         & 6          & RGB                 & /               & 4.7  \\ \midrule
Search/Rescue                      & Search And Rescue \cite{thoreau2021sarnet}                   & 2021 & 2552        & 1000                  & 1          & RGB                 & 0.5m            & /    \\  \midrule
\multirow{14}{*}{Ship}             & OpenSARShip  \cite{huang2017opensarship}                        & 2017 & 11346       & 900                   & 1          & Sentinnel-1         & 10m             & 1.7   \\
                                   & SSDD  \cite{li2017ship}                               & 2017 & 1160        & 500                   & 2          & SAR                 & 1$\sim$15m      & /   \\
                                   & HRSC2016 (Ship) \cite{liu2017high}                      & 2017 & 1061        & 300x300$\sim$1500x900 & 26         & RGB                 & 0.4$\sim$2m     & 3.74   \\
                                   & Kaggle Airbus Ship \cite{AirbusSDC}                  & 2018 & 192556      & 768                   & 2          & /                   & 1.5m            & 31.41  \\
                                   & Airbus Ship Detection \cite{AirbusSDC}                & 2018 & 40,000      & 768                   & 2          & RGB                 & /               & 31.4  \\
                                   & Ships in Google Earth \cite{ShipsGoogleEarth}               & 2018 & 794         & 2000                  & 2          & RGB                 & /               & 2  \\
                                   & SAR Ship Detection \cite{rs11070765}                  & 2019 & 43819       & 256                   & 2          & SAR                 & 3m, 5m, 8m,10m  & 0.4  \\
                                   & AIR-SARShip-1.0 \cite{xian2019air}                     & 2019 & 31          & 3000                  & 2          & SAR                 & 1$\sim$3m       & 0.24  \\
                                   & AIR-SARShip-2.0 (GF-3) \cite{wang2022air}              & 2020 & 300         & 1,000                 & 2          & SAR                 & 1$\sim$3m       & 0.22  \\
                                   & LS-SSDD (Large Scale) \cite{LS_SSDD}               & 2020 & 15          & 20,000                & 2          & Sentinel-1,SAR      & 0.5,1,3m        & 7.8   \\
                                   & HRSID (Ship) \cite{wei2020hrsid}                        & 2020 & 5,604       & 800                   & 2          & Sentinel-1,SAR      & 0.5$\sim$3m     & 0.58   \\
                                   & SWIM-Ship \cite{xue2021rethinking}                           & 2021 & 14610       & 768                   & 2          & RGB                 & 0.5$\sim$2.5m   & 12.5  \\
                                   & CASIA-Ship  \cite{CASIA_Ship}                         & 2021 & 1,118       & 1,680                 & 2          & RGB                 & /               & /  \\
                                   & xView3-SAR \cite{xview3}                          & 2022 & 1000        & $\sim$29400x24400     & 2          & Sentinel-1          & 10m             & 1500  \\  \midrule
\multirow{2}{*}{Small Objects}     & AI-TOD \cite{AI-TOD_2020_ICPR}                              & 2021 & 28036       & 800                   & 8          & RGB                 & 0.3m$\sim$30m   & 42     \\
                                   & SODA-A \cite{cheng2022towards}                              & 2022 & 2510        & 4761×2777             & 9          & RGB                 & /               & /    \\  \midrule
\multirow{5}{*}{Traffic Objects}   & AU-AIR \cite{bozcan2020air}                              & 2020 & 32823       & 1920                  & 8          & RGB                 & UAV@30m         & 2.2    \\
                                   & HighD \cite{highDdataset}                               & 2018 & 110000      & 4096x2160             & 2          & RGB                 & /               & /   \\
                                   & Interaction Dataset \cite{interactiondataset}                 & 2019 & 10,933      & /                     & 1          & RGB                 & /               & /   \\
                                   & Intersection Drone \cite{inDdataset}                  & 2020 & 11,500      & 4096x2160             & 5          & RGB                 & /               & /   \\
                                   & Roundabouts Drone \cite{rounDdataset}                   & 2020 & 13,746      & 4096x2160             & 8          & RGB                 & /               & /   \\  \midrule
\multirow{2}{*}{Tree}              & NEON Tree Crowns \cite{ben_weinstein_2020_3765872}                                           & 2020 & 11,000      & 100 million trees     & 2          &                                    RGB                 & /               & 27.4   \\
                                   & Oil Palm Plant~\cite{widsdatathon2019} &2019 &20,000 & / & 2 & RGB  & 3m  & 0.5 \\
                                   & Forest Damages \cite{SwedishFA}                      & 2021 & 1543        & 1500                  & 5          & RGB                 & UAV             & 3.3  \\   \midrule
\multirow{20}{*}{Vehicles}         & Things And Stuff (TAS) \cite{heitz2008learning}              & 2008 & 30          & 792                   & 2          & RGB                 & 0.5m            & 0.01   \\
                                   & OIRDS  \cite{OIRDS}                              & 2009 & 900         & 256$\sim$640          & 5          & RGB                 & 0.15m           & 0.153   \\
                                   & UCAS\_AOD \cite{zhu2015orientation}                           & 2014 & 976         & 1,000                 & 2          & RGB                 & /               & 3.24    \\
                                   & VEDAI(Vehicle) \cite{razakarivony2016vehicle}                      & 2015 & 1,250       & 1,024                 & 9          & IRGB                & 0.125m          & 3.9   \\
                                   & PKLot(Parking Lot) \cite{de2015pklot}                  & 2015 & 12417       & 1280                  & 2          & RGB                 & UAV             & 4.6  \\
                                   & COWC \cite{mundhenk2016large}                                & 2016 & 388435      & 256                   & 2          & RGB                 & 0.15m           & 62.5  \\
                                   & Car Parking Lot (CARPK) \cite{hsieh2017drone}             & 2016 & 1448        & 1280                  & 2          & RGB                 & UAV             & 2    \\
                                   & DLR3k/DLR-MVDA \cite{DLR3K}                      & 2016 & 20          & 3744                  & 7          & RGB                 & 0.13m           & 0.162   \\
                                   & ITCVD (Vehicle) \cite{yang*18:icip}                      & 2018 & 173         & 5616                  & 2          & RGB                 & 0.1m            & 12  \\
                                   & VisDrone2019-DET \cite{9573394}                    & 2019 & 10209       & 2000x1500             & 10         & RGB                 & UAV             & 2    \\
                                   & VisDrone2019-VID \cite{9573394}                    & 2019 & 40000       & 3840x2160             & 5          & RGB                 & UAV             & 14  \\
                                   & VisDrone2019-SOT \cite{9573394}                    & 2019 & 139300      & 3840x2160             & 3          & RGB Video           & UAV       & 68  \\
                                   & VisDrone2019-MOT \cite{9573394}                    & 2019 & 40000       & 3840x2160             & 5          & RGB Video           & UAV             & 14  \\
                                   & SIMD (Multi-vehicles) \cite{9109702}               & 2020 & 5000        & 1024                  & 15         & RGB                 & UAV@150m        & 1  \\
                                   & VisDrone \cite{zhu2021detection}                            & 2020 & 275,437     & 1,400                 & 11         & RGB                 & UAV             & 16   \\
                                   & EAGLE \cite{9412353}                               & 2020 & 8820        & 936                   & 2          & RGB                 & 0.05$\sim$0.45m & /   \\
                                   & MOR-UAV \cite{mandal2020mor}                             & 2020 & 10948       & 1080                  & 1          & RGB                 & UAV             & /   \\
                                   & DroneVehicle \cite{sun2020drone}                        & 2021 & 56,878      & 840                   & 5          & RGB-Infrared        & UAV@100m        & 13.09  \\
                                   & Swimming pool/car \cite{SwimmingPool_Car}                   & 2019 & 3750        & 224                   & 2          & RGB                 & /               & 0.12   \\
                                   & ArtifiVe-Potsdam \cite{weber2021artifive} & 2021 & 4800  & 600  &  1 & MS  & 0.05m & 15.6 \\
                                   \bottomrule
\end{tabular}}
\label{RSIOD}
\end{table*}
There are 13 datasets \cite{mohanty2020deep,TBF,AustraliaBuilding,UgandaTanzaniaBuilding,USBuilding,USCanada,UBCBuilding} that are designed for building extraction with pixel-level annotations. Note that some of them are constructed for building instance segmentation tasks. In those cases, the buildings are annotated with both the object-level and pixel-level labels.
For road extraction, 10 datasets \cite{liu2019roadnet,maggiori2017can,abdollahi2020deep,mattyus2016hd,mattyus2015enhancing,bastani2018roadtracer,mrd} are constructed. Datasets designed for LULC, general scenes, buildings, and road segmentation dominate the RS semantic segmentation tasks. We present 10 datasets built with lower spatial resolution RS images than other domains \cite{foga2017cloud,38-cloud-1,95-cloud,li2018deep,li2021lightweight} for cloud-related applications. Furthermore, 9 agriculture datasets are annotated with pixel-level labels towards a fine-grained understanding of the crops~\cite{barsi2014spectral,ji2020learning,weikmann2021timesen2crop,garnot2021panoptic,chiu2020agriculture}. 
\begin{table*}[]
\centering
\caption{Detailed Information for some representative RS Semantic Segmentation Datasets. These datasets are grouped into 19 different research domains in alphabetical order. Note that / denotes the missing information and MS denotes the multispectral data. More information can be found at \url{https://earthnets.github.io}.}
\small
\scalebox{0.63}{
\begin{tabular}{ccccccccc}
\toprule
Domain                           & Name                                     & Year & \#Samples & Sample Size                    & \#Classes & Modailty              & Resolution     & Vol.(GB) \\  \midrule
\multirow{9}{*}{Agriculture}     & Agricultural Crop Cover\cite{barsi2014spectral}                & 2018 & 40        & /                              & 2         & MS                    & 30m            & 4.4    \\
                                 & GF2 Dataset for 3DFGC \cite{ji2020learning}                   & 2019 & 11        & 2,652                          & 5         & RGB-NIR               & 4m             & 0.056      \\
                                 & TimeSen2Crop \cite{weikmann2021timesen2crop}                            & 2020 & 1000000   & 10980                          & 16        & Sentinel-2            & 10m            & 1.1   \\
                                 & Agriculture-Vision \cite{chiu2020agriculture}                      & 2020 & 94986     & 512                            & 9         & RGB-NIR               & 0.1$\sim$0.2m  & 4.4      \\
                                 & WHU-Hi-LongKou \cite{zhong2020whu}                          & 2020 & 1         & 550x400                        & 9         & Hyperspectral         & 0.463m         & /   \\
                                 & WHU-Hi-HanChuan \cite{zhong2020whu}                         & 2020 & 1         & 1217x303                       & 16        & Hyperspectral         & 0.109m         & /     \\
                                 & WHU-Hi-HongHu \cite{zhong2020whu}                           & 2020 & 1         & 940x475                        & 22        & Hyperspectral         & 0.043m         & /     \\
                                 & ZueriCrop  \cite{turkoglu2021crop}                              & 2021 & 28000     & 24                             & 48        & Sentinel-2            & 10m            & 39    \\
                                 & EuroCrops \cite{Schneider2021EPE}                               & 2021 & 805,401   & 0.5 ha                         & 43        & Sentinel-2            & 10m            & 8.6 \\  \midrule
Arctic                           & Arctic Sea Ice Image Masking \cite{wang2021arctic}            & 2021 & 3392      & 357x306                        & 8         & RG-NIR                & 10m            & 0.092   \\  \midrule
\multirow{13}{*}{Building}       & SpaceNet-1 (Building)\cite{van2018spacenet}                    & 2016 & 9735      & 650     & 2         & RGB,MS                & 0.5$\sim$1m    & 31   \\
                                 & SpaceNet-2 (Building)\cite{van2018spacenet}                    & 2017 & 24586     & 650                            & 2         & RGB,MS                & 0.3m           & 104  \\
                                 & INRIA Aerial Image Labeling \cite{maggiori2017can}             & 2017 & 360       & 1500                           & 1         & RGB                   & 0.3m           & 19.5     \\
                                 & built-structure-count dataset\cite{shakeel2019deep}            & 2019 & 5364      & 512                            & 1         & RGB                   & 0.3m           & 2.23  \\
                                 & SpaceNet-6 (Multi-Sensor All Weather)\cite{shermeyer2020spacenet}    & 2020 & 3401      & 900                            & 2         & SAR,RGB               & 0.5m           & 55.9   \\
                                 & SpaceNet-7 (Multi-Temporal Urban)\cite{van2021multi}        & 2020 & 1525      & 1024                           & 2         & MS,MT                 & 4m             & 20.1 \\
                                 & Synthinel-1 \cite{kong2020synthinel}                             & 2020 & 2108      & 572                            & 2         & RGB                   & 0.3m           & 0.977  \\
                                 & Kaggle buildings segmentation\cite{kagglebuilding1}            & 2020 & 6038      & 256                            & 2         & RGB                   & /              & 0.899  \\
                                 & Kaggle Massachusetts Buildings\cite{MnihThesis}           & 2020 & 151       & 1500                           & 2         & RGB                   & 1m             & 2.93   \\
                                 & Open Cities AI Challenge\cite{OpenAIChallenge}                 & 2020 & 11,000    & 1,024                          & 2         & RGB                   & 0.03$\sim$0.2m & 81.5   \\
                                 & Mini Inria Aerial Image Labeling Dataset\cite{maggiori2017can} & 2021 & 32,500    & 512                            & 2         & RGB                   & 0.3m           & /    \\
                                 & High-speed Rail Line Building Dataset\cite{Building_Dataset}    & 2021 & 336       & 2000                           & 2         & RGB                   & 0.5m           & /   \\
                                 & AIS From Online Maps \cite{kaiser2017learning}                    & 2017 & 1671      & 3000                           & 2         & RGB                   & 0.5m           & 23.8   \\  \midrule
\multirow{10}{*}{Cloud}          & Biome: L8 Cloud Cover\cite{foga2017cloud}                    & 2016 & 96        & /          & 4         & RGB           & 30m            & 96   \\
                                 & 38-Cloud \cite{38-cloud-1}                            & 2018 & 17601     & 384                            & 2         & RGB                   & 30m            & 13   \\
                                 & Sentinel-2 Cloud Detection (ALCD)\cite{baetens2019validation}        & 2019 & 38        & 1830                           & 2         & MS,Sentinel-2         & 10$\sim$60m    & 0.234  \\
                                 & HRC\_WHU \cite{li2018deep}                                & 2019 & 150       & 1280x720                       & 2         & RGB                   & 0.5$\sim$15m   & 0.17  \\
                                 & WHU Cloud Dataset\cite{ji2020simultaneous}                        & 2020 & 859       & 512                            & 2         & RGB                   & 30m            & 3.56   \\
                                 & 95-Cloud \cite{95-cloud}                                & 2020 & 34701     & 384                            & 2         & RGB                   & 30m            & 18  \\
                                 & WHUS2-CD+ \cite{li2021lightweight}                                & 2021 & 36        & 10980                          & 2         & Sentinel-2            & 10m            & 27.8   \\
                                 & AIR-CD  \cite{AIR-CD}                                 & 2021 & 34        & 7300                           & 2         & RGB-NIR               & 4m             & 13   \\
                                 & The Azavea Cloud Dataset\cite{azavea}                 & 2021 & 32        & /                              & 2         & Sentinel-2            & 10m$\sim$60m   & /   \\
                                 & Sentinel-2 Cloud Cover \cite{s2cloudcover}                  & 2022 & 22728     & /                              & 2         & MS                    & 10m$\sim$60m   & 51.2   \\   \midrule
\multirow{3}{*}{General Objects} & DLRSD \cite{shao2018performance}                                & 2018 & 2100      & 256                            & 17        & RGB                   & 0.3m           & 0.004   \\
                                 & Kaggle aerial segmentation \cite{kaggleaerial}    & 2020 & 72        & 800                            & 6         & RGB                   & /              & 0.033    \\
                                 & AIR-PolSAR-Seg \cite{wang2022air}                          & 2022 & 2000      & 512                            & 6         & SAR                   & 8m             & 0.609  \\   \midrule
\multirow{18}{*}{General Scenes} & ISPRS 2D - Potsdam \cite{ISPRS-Contest1}                      & 2011 & 38        & 6000                           & 6         & RGB,nDSM              & 0.05m          & 15.625   \\
                                 & ISPRS 2D - Vaihingen \cite{ISPRS-Contest2}                    & 2011 & 33        & 2200                           & 6         & RGB,nDSM              & 0.09m          & 16.6    \\
                                 & Aerial Image Segmentation \cite{yuan2013systematic}               & 2013 & 80        & 512                            & 2         & RGB                   & 0.3$\sim$1m    & 0.007   \\
                                 & DFC2015 Zeebruges  \cite{moser20152015}                      & 2015 & 7         & 100,000                        & 8         & RGB,DSM,LiDAR         & 0.05m          & 0.0024  \\
                                 & Zurich Summer Dataset  \cite{volpi2015semantic}                  & 2015 & 20        & 1000                           & 8         & RGB-NIR               & 0.61m          & 0.38  \\
                                 &Ticino~\cite{barbato2024ticino}  &2024 &1502  & / & 10  & RGB,DTM,Pan.,Hyperspectral  & 186m-30m & / \\
                                 & EvLab-SS Dataset \cite{zhang2017learning}                        & 2017 & 60        & 4500                           & 11        & RGB                   & 0.1m,0.25m     & /    \\
                                 & SynthAer \cite{phdthesis}                                & 2018 & 765       & 1280                           & 8         & RGB                   & /              & 0.977  \\
                                 & Aeroscapes \cite{nigam2018ensemble}                              & 2018 & 3269      & 1280                           & 11        & RGB                   & UAV@5-50m      & 0.73  \\
                                 & Urban Drone Dataset (UDD) \cite{chen2018large}               & 2018 & 301       & 4,096                          & 6         & RGB                   & UAV            & 1.1  \\
                                 & RIT-18 \cite{kemker2018algorithms}                                  & 2018 & 3         & 9393x5642,8833x6918,12446x7654 & 18        & MS                    & 0.047m         & 1.5  \\
                                 & Semantic Drone Dataset-SemSeg \cite{SDD}           & 2019 & 400       & 5000                           & 20        & RGB                   & /              & 3.91   \\
                                 & DroneDeploy  \cite{DDD}                            & 2019 & 55        & 6,000                          & 7         & RGB                   & 0.1m           & /    \\
                                 & MidAir  \cite{fonder2019mid}                                 & 2019 & 420000    & 1024                           & 12        & RGBD,Odometry         & /              & 1000  \\
                                 & AeroRIT  \cite{rangnekar2020aerorit}                                & 2019 & 1         & 3975x1973                      & 6         & RGB,Hyperspectral     & 0.4m           & 1.8  \\
                                 & SemCity Toulouse  \cite{roscher2020semcity}                       & 2020 & 16        & 3500                           & 8         & MS                    & 0.5$\sim$2m    & 8.8  \\
                                 & UAVid  \cite{LYU2020108}                                  & 2020 & 420       & 4000                           & 8         & RGB                   & UAV            & 5.88 \\   \midrule
Settlements                      & DFC21-DSE  \cite{DFC2021}                              & 2021 & 98        & 800                            & 2         & SAR,MS,Hyperspectral  & 10$\sim$750m   & 18    \\   \midrule
\multirow{18}{*}{Land Cover}     
                                 & DeepGlobe (LandCover) \cite{demir2018deepglobe}                   & 2018 & 1146      & 2448                           & 7         & RGB                   & 0.5m           & 2.96  \\
                                 & HyRANK \cite{HyRANK}                                  & 2018 & 5         & 1,000                          & 14        & Hyperspectral         & 30m            & 0.4    \\
                                 & WHDLD  \cite{shao2018performance}                                  & 2018 & 4940      & 256                            & 6         & RGB          & 2m           & 0.102   \\
                                 & SEN12MS \cite{schmitt2019sen12ms}                                 & 2019 & 541,986   & 256                            & 17        & MS,SAR                & 10m            & 510  \\
                                 & Urban Semantic 3D (DFC19) \cite{le20192019}               & 2019 & 2783      & 1024                           & 6         & MS,LiDAR              & 0.3$\sim$1.3m  & 285 \\
                                 & Chesapeake Land Cover \cite{robinson2019large}                   & 2019 & 100000    & 224                            & 6         & RGB,MS            & 1m         & 404.95 \\
                                 & DFC20 \cite{yokoya20202020}                                   & 2020 & 180662    & 256                            & 10        & Sentinel-1,Sentinel-2 & 10m            & 9.6  \\
                                 & BDCI2020  \cite{BDCI}                               & 2020 & 145,981   & 256                            & 7         & RGB         & 2m          & 1.3     \\
                                 & LandCoverAI \cite{boguszewski2021landcover}                             & 2020 & 41        & 9000                           & 3         & RGB     & 0.25m,0.5m     & 1.4    \\
                                 & Luojia-HSSR~\cite{xu2023luojia}   & 2023 &6438   & 256           & 23        & RGB          & 0.75m    & /  \\
                                 & LoveDA  \cite{wang2021loveda}                                 & 2021 & 5,987     & 1,024                          & 7         & RGB                   & 0.3m           & 9.6  \\
                                 & MiniFrance-DFC22 \cite{castillo2021semi}                        & 2022 & 2322      & 2000                           & 15        & RGB                   & 0.5m           & 93   \\
                                 & GeoNRW  \cite{baier2021synthesizing}                                 & 2022 & 7783      & 1000                           & 10        & RGB,nDSM         & 1m      & 32  \\
                                 & SEASONET \cite{kossmann2022seasonet}                                & 2022 & 1759830   & 120                            & 33        & Sentinel-2     & 10m            & 229     \\
                                 & TimeSpec4LULC \cite{khaldi2022timespec4lulc}                           & 2022 & /         & 262 months                     & 29        & MS,MT      & 500m        & 60  \\
                                 & Five-Billion-Pixels \cite{2022FBP}                     & 2022 & 150       & 7200×6800                      & 24        & RGB,MS                & 4m             & 104  \\
                                 & GAMUS~\cite{xiong2023gamus}         & 2023 & 9340      & 1024                           & 6         & RGB,nDSM              & 0.33m          & 48   \\
                                 & WHU-OHS \cite{li2022whu}                               & 2022 & 7795      & 512                            & 24        & Hyperspectral         & 10m            & 94.9     \\   \midrule
\multirow{3}{*}{Land Use}        & OpenSentinelMap \cite{johnson2022opensentinelmap}                         & 2022 & 137045    & 192, 96         & 15        & RGB,Sentinel-2        & 10m$\sim$60m   & 455   \\
                                 & DFC18  \cite{le20182018}                                  & 2018 & 10,798    & 2,001                          & 20        & MS,Hyperspectral,RGB  & 0.05$\sim$1m   & 10.1 \\
                                 & MultiSenGE \cite{wenger2022multisenge}                              & 2022 & 8157      & 256                            & 14        & Sentinel-1,Sentinel-2 & 10m        & 530  \\   \midrule
Parking                          & APKLOT \cite{hurst2020}                                  & 2020 & 501       & /                              & 2         & RGB                   & /              & 3    \\  \midrule
Power                            & TTPL \cite{abdelfattah2020ttpla}      & 2020 & 1100      & 3840                           & 3         & RGB                   & UAV            & 4.2       \\    \midrule
\multirow{10}{*}{Road}            & Massachusetts Roads \cite{MnihThesis}                     & 2013 & 1171      & 1500                           & 1         & RGB                   & 1m             & 10.56  \\
                                 & ERM PAIW \cite{mattyus2015enhancing}                                & 2015 & 41        & 4000       & 2         & RGB        & 0.3m        & 0.635  \\
                                 & HD-Maps \cite{mattyus2016hd}                                & 2016 & 20        & 4000                           & 5         & RGB             & 0.3m          & 0.146    \\
                                 & SpaceNet-3 (Road ) \cite{van2018spacenet}                      & 2017 & 3711      & 1300                           & 2         & Panchromatic,RGB,MS   & 0.3$\sim$1.24m & 106    \\
                                 & RoadNet  \cite{liu2019roadnet}                                & 2018 & 20        & /            & 2         & RGB                   & 0.21m          & 0.905   \\
                                 & AerialLanes18 \cite{he2022lane}                          & 2018 & 20        & 5616                           & 1         & RGB                   & 0.125m         & 0.0014 \\
                                 & SpaceNet-5 (Road Network) \cite{van2018spacenet}               & 2019 & 2369      & 1300                           & 2         & Panchromatic,RGB,MS   & 0.3m           & 84   \\
                                 & SpaceNet-8 (Flooded Road) \cite{hansch2022spacenet}               & 2022 & /      & 1300                           & 4         & Panchromatic,RGB   & 0.3$\sim$0.8m           & /   \\
                                 & RoadTracer \cite{bastani2018roadtracer}                           & 2019 & 3,000     & 4,096                          & 1         & RGB                   & 0.6m           & /     \\
                                 & Microsoft RoadDetections \cite{mrd}                & 2022 & 20000     & 1088                           & 1         & RGB                   & 1m             & 9.25   \\  \midrule
\multirow{3}{*}{Roof}            & AIRS \cite{chen2019}                              & 2019 & 1047      & 10000                          & 1         & RGB                   & 0.075m         & 17.6   \\
                                 & Open AI Challenge: Caribbean\cite{caribbean}             & 2019 & 7         & 52,318                         & 5         & RGB                   & 0.04m          & /  \\
                                 & RID \cite{krapf2022rid}                                    & 2022 & 2000      & /                              & 16        & RGB           & 0.1m         & 1.5      \\   \midrule
\multirow{2}{*}{Salient Objects} & ORSSD \cite{zhang2020dense}                                   & 2019 & 800       & 500                            & 8         & RGB                   & /              & 0.026  \\
                                 & EORSSD \cite{zhang2020dense}                                 & 2020 & 2,000     & 500                            & 2         & RGB        & /            & 0.06      \\  \midrule
Shadow                           & AISD \cite{luo2020deeply}                                   & 2020 & 514       & 512                            & 2         & RGB                   & /              & 0.29   \\  \midrule
Traffic Scenes                   & DLR-SkyScapes\cite{azimi2019skyscapes}                           & 2019 & 16        & 4680                           & 31        & RGB                   & 0.13m          & /   \\   \midrule
Water Body                       & Kaggle Water Bodies\cite{swt}                   & 2020 & 2841      & 1000                           & 2         & RGB                   & /              & 0.28  \\ \bottomrule
\end{tabular}}
\label{RSISEG}
\end{table*} 
\subsection{RS Change Detection Datasets}
RS change detection focuses on quantitatively analyzing surface alterations using data obtained from remote sensors. This research is vital for practical applications, including damage monitoring and urban planning. Table \ref{RSCD} presents some representative datasets built for RS change detection tasks. Flood events and other natural disasters can inflict damage on structures. In these scenarios, change detection plays a crucial role in assessing the extent of the damage and supplying data for analysis and rescue operations. Therefore, datasets related to buildings~\cite{shen2021s2looking} and floods~\cite{luppino2019unsupervised} have been developed to facilitate these applications. The majority of change detection datasets primarily concentrate on changes in land cover or land use. Detecting these changes is crucial for analyzing urban development, as well as for urban growth and planning. In addition, the time-series data can aid in monitoring crops to ensure food security~\cite{liu2022cnn}. 

Several other datasets are constructed for 3D change detection \cite{de2021change}. Many different data modalities are used for constructing these datasets, including optical (RGB), point cloud, multispectral, hyperspectral, and SAR data. Note that some datasets like SECOND \cite{yang2021asymmetric} and Dynamic World \cite{brown2022dynamic} are annotated with pixel-wise semantic labels, not only the change/non-change binary label.

\begin{table*}[]
\centering
\caption{Detailed Information for some Representative RS Change Detection Datasets. These datasets are grouped into 5 different research domains in alphabetical order. Note that / denotes the missing information and MS denotes the multispectral data. More detailed information can be found at \url{https://earthnets.github.io}.}
\small
\scalebox{0.65}{
\begin{tabular}{ccccccccc}
\toprule
Domain                        & Name                                  & Year & \#Samples & Sample Size                & \#Classes & Modailty             & Resolution                 & Vol.(GB)  \\  \midrule
\multirow{2}{*}{3D}                            
                              & URB3DCD\cite{de2021change}                              & 2021 & 50        & /                          & 2         & PointCloud           & 0.5 pm                     & 1.5  \\ 
                              & 3DCD~\cite{marsocci2023inferring}    & 472  & 400$\times$400  & 2  & RGB,nDSM  & / & / \\
                              \midrule
\multirow{4}{*}{Building}     & AIST Building Change Detection \cite{fujita2017damage}       & 2017 & 16950     & 160                        & 2         & RGB                  & 0.4m       & 17.77     \\
                              & WHU Building change detection \cite{ji2018fully} & 2018 & 2         & 15354×32507                & 2         & RGB                  & 0.075m                     & 5   \\
                              & LEVIR-CD \cite{Chen2020}                             & 2020 & 637       & 1024                       & 2         & RGB                  & 0.5m                       & 2.64   \\
                              & xView2 (xBD) \cite{gupta2019xbd}                         & 2018 & 22068     & 1024                       & 4         & RGB                  & 0.5m                       & 51  \\ \midrule
CropLand               & CropLand Change Dection (CLCD) \cite{liu2022cnn}       & 2022 & 600       & 512          & 2         & RGB        & 0.5$\sim$2 m         & 0.5     \\ \midrule
Flood                  & California flood dataset \cite{luppino2019unsupervised}             & 2019 & 1         & 1534×808                   & 2         & RGB,MS               & 5m,30m        & 0.33  \\  \midrule
\multirow{29}{*}{Land Change} & SZTAKI AirChange \cite{benedek2009change}                     & 2008 & 13        & 800             & 2         & RGB        & 1.5m        & 0.04   \\
                              & Cross-sensor Bastrop \cite{volpi2015spectral}                 & 2015 & 4         & 444x300,1534x808           & 2         & MS      & 30m,120m             & /   \\
                              & GETNET dataset \cite{wang2018getnet}                       & 2018 & 1         & 463x241                    & 2         & Hyperspectral        & 30m             & 0.05      \\
                              & Onera Satellite CD \cite{daudt2018urban}                   & 2018 & 24        & 600            & 2         & Sentinel-2       & 10m           & 0.48    \\
                              & CDD (season-varying)\cite{lebedev2018change}                  & 2018 & 16000     & 256                        & 2         & RGB         & 0.03$\sim$0.1m        & 2.7  \\
                              & Hyperspectral CD\cite{mhcd}                      & 2018 & 3     & 984x740,600x500,390x200    & 5         & Hyperspectral        & 30m          & 1.7   \\
                              & HRSCD \cite{daudt2018urban}                                & 2019 & 291       & 10000            & 5         & RGB             & 0.5m          & 5   \\
                              & MtS-WH \cite{wu2016scene}                               & 2019 & 190       & 150                        & 2         & RGB-NIR   & 1m               & 0.43    \\
                              & SECOND \cite{yang2021asymmetric}                               & 2020 & 4662      & 512                        & 6         & RGB         & 0.5$\sim$3m     & 2.2   \\
                              & Zhang et al. CD dataset \cite{zhang9052762}              & 2020 & 4         & 1431×1431,458×559,1154×740 & 2         & RGB,NIR       & 2m,2.4m,5.8m       & 0.1   \\
                              & DSIFN \cite{zhang2020deeply}                                & 2020 & 3,988     & 512                        & 2         & RGB    & 10m            & 0.46  \\
                              & Hermiston City Oregon \cite{lopez2018stacked}           & 2018 & 1     & 390x200                        & 5         & Hyperspectral                  & 30m                & /    \\
                              & Hi-UCD \cite{tian2020hi}                               & 2020 & 1293      & 1024                       & 9         & RGB                  & 0.1m                       & /    \\
                              & Google Data Set \cite{peng2020semicdnet}                      & 2020 & 19        & 1000$\sim$5000             & 2         & RGB            & 0.55m             & 0.6   \\
                              & DFC21-MSD  \cite{DFC2021}                           & 2021 & 2250      & 4000                       & 15        & MS,MT       & 1$\sim$30m         & 325      \\
                              & Relative Radiometric Normalization \cite{moghimi2020novel}   & 2021 & 7         & 300$\sim$5000              & 2         & MS                   & 0.31m,0.4m,10m,20m,30m,60m & 1   \\
                              & HTCD \cite{shao2021sunet}                                 & 2021 & 2         & 11 K×15 K,1.38 M×1.04 M    & 2         & RGB                  & 0.5971m, 0.074m            & 1.74   \\
                              & S2Looking \cite{shen2021s2looking}                            & 2021 & 5000      & 1,024                      & 2         & RGB                  & 0.5$\sim$0.8m              & 10.21  \\
                              & SYSU-CD \cite{shi2021deeply}                              & 2021 & 20,000    & 256                        & 2         & RGB    & 0.5m            & 5.17  \\
                              & WH-MAVS \cite{yuan2022wh}                              & 2021 & 47,134    & 200                        & 15        & RGB       & 1.2m      & /      \\
                              & S2MTCP \cite{leenstra2021self}                               & 2021 & 1520      & 600                        & /         & MS     & 10m         & 10.6  \\
                              &RapidAI4EO~\cite{marchisio2021rapidai4eo} &2021  &500,000 & / & 44  & MS,RGB  & 3m,10m & / \\
                              & Dynamic EarthNet Challenge \cite{toker2022dynamicearthnet}           & 2021 & 22500     & 1024                       & 7         & RGB       & 3m          & /   \\
                              & MSBC \cite{li2022mscdunet}                                 & 2022 & 3,769     & 256             & 2         & RGB,SAR,MS           & 2m        & 3.9    \\
                              & Dynamic World \cite{brown2022dynamic} &2022  &/  &/  &9 & Sentinel-2  &10m  &/  \\
                              & MSOSCD \cite{li2022mscdunet}                               & 2022 & 5,107     & 256          & 2         & RGB,SAR,MS           & 10$\sim$60m & 2.7  \\   
                              &ChangeNet~\cite{ji2024changenet}  &2024  &31000 &/  &6  & RGB  &0.3m  & / \\
                              &BANDON~\cite{pang2023detecting}   & 2023  &2283  &2048$\times$2048  &2 & RGB  & 0.6m & 3.4 \\
                              \bottomrule
\end{tabular}}
\label{RSCD}
\end{table*}

\subsection{Datasets for RS Foundation Models}
As we reviewed, extensive annotated RS datasets have been constructed to train task-specific algorithms~\cite{camps2021deep}. However, these traditional deep learning models require significant human labor for data collection and annotation, as well as considerable computational costs for model development and evaluation. In contrast, foundation models~\cite{bommasani2021opportunities}, trained on a wide array of data, can be fine-tuned for specific downstream tasks using far fewer annotated examples. This efficiency stems from their capacity to leverage general feature representations learned from vast amounts of unlabeled data. To foster research on foundation models, many RS datasets have been built for pre-training and evaluating them, as presented in Table \ref{RSFoundation}. Among them, SatlasPretrain~\cite{bastani2023satlaspretrain} is a large-scale RS dataset containing data from different sensors with 302,222,000 annotated labels, which is useful for training RS foundation models. Metadataset is a collection of existing datasets, which can be used for pre-training or evaluating RS foundation models. For instance, DOFA-data~\cite{xiong2024neural}, GeoPile(GFM)~\cite{mendieta2023building}, and Prithvi~\cite{jakubik2023foundation} construct large-scale dataset by assembling different sub-datasets. This concept has also been adopted by GEO-Bench~\cite{lacoste2024geo}, SATIN~\cite{roberts2023satin}, and FoMo-Bench~\cite{bountos2024fomobench} for the evaluation of RS foundation models across various tasks. 

\subsection{RS Vision-Language Datasets}
Large Language Models (LLMs) and Vision-Language Models (VLMs) have attracted significant attention in research on Natural Language Processing (NLP) and CV. These models leverage large-scale Transformer networks to enhance natural language understanding, multi-modal learning, and reasoning ability, achieving significant performance improvements across various tasks. Compared with the reviewed image classification, object detection, semantic segmentation, and change detection tasks, research on VLMs still signifies an emerging and innovative area in the RS domain. Various vision-language datasets have been constructed to accelerate research on VLMs, as presented in Table \ref{RSFoundation}.
\begin{table*}[htbp]
\centering
\caption{Detailed information of some representative datasets for RS foundation models and vision-language models. Note that / denotes the missing information and MS denotes the multispectral data. More information can be found at \url{https://earthnets.github.io}.}
\small
\scalebox{0.65}{
\begin{tabular}{cccccccc}
\toprule
Domain                                           & Name                   & Publication year & \# samples         & Task                            & \# classes & Modality                  & Resolution  \\ \midrule
\multirow{5}{*}{Foundation Models Evaluation}    & GEO-Bench~\cite{lacoste2024geo}              & 2023             & 12 subdatasets     & SemSeg,Class                    & /          & RGB,SAR,MS,Hyperspectral  & 0.5m - 30m  \\
                                                 & SATIN~\cite{roberts2023satin}                  & 2023             & 27 subdatasets      & SemSeg,Class                    & 250        & RGB                       & /           \\
                                                 & PhilEO~\cite{fibaek2024phileo}                 & 2024             & 3 subdatasets                 & SemSeg                          & 11         & RGB,MS         & 10m         \\
                                                 & FoMo-Bench~\cite{bountos2024fomobench}             & 2024             & 15 subdatasets                  & SSL                             & /          & RGB,MS,SAR     & 0.5m - 20m  \\
                                                 & SustainBench~\cite{yeh2021sustainbench}           & 2021             & 15 subdatasets                 & Regression,Class                & /          & RGB,Time-seris            & 30m         \\ \midrule
\multirow{10}{*}{Foundation Models Pre-training} & SatlasPretrain~\cite{bastani2023satlaspretrain}         & 2021             & 302,222,000 labels & OD,CD,Class,SemSeg              & 137        & NAIP,RGB,MS               & 1m,10m      \\
                                                 & SSL4EO-L~\cite{stewart2024ssl4eo}               & 2023             & 5,000,000          & SSL                             & /          & MS                        & 30m         \\
                                                 & CACo1M~\cite{mall2023change}                 & 2023             & 1,000,000          & SSL                             & /          & Sentinel,MS    & 10m         \\
                                                 & TOV-RS-Balanced~\cite{tao2023tov}        & 2023             & /                  & SSL                             & /          & RGB,SAR,Hyperspectral     & /           \\
                                                 & GeoPile(GFM)~\cite{mendieta2023building}           & 2023             & 600,000            & SSL                             & /          & RGB                       & 0.1m - 30m  \\
                                                 & DOFA-data(Metadataset)~\cite{xiong2024neural} & 2024             & 8,081,411          & SSL                             & /          & RGB, SAR,MS,Hyperspectral & 0.5m - 30m  \\
                                                 & GRAFT~\cite{mall2023remote}                  & 2024             & 189,000,000        & SSL                             &            & RGB,MS                    & /           \\
                                                 & SAMRS(Metadataset)~\cite{wang2024samrs}     & 2023             & /                  & SemSeg,OD                       & /          & RGB                       & /           \\
                                                 & Prithvi~\cite{jakubik2023foundation}                & 2023             & /                  & SSL                             & /          & HLS-2                     & /           \\
                                                 & Skysense~\cite{guo2023skysense}         & 2024             & 215,000,000        & SSL                             & /          & RGB,SAR,MS                & 0.31m - 20m \\ \midrule
\multirow{15}{*}{Vision-Language Datasets}       & LuojiaHOG~\cite{zhao2024luojiahog}              & 2024             & 565231 captions    & Captioning/Retrieval & 131        & RGB                       & 0.5m        \\
                                                 & RSVG~\cite{zhan2023rsvg}                   & 2023             & 38320 captions     & VG                              & /          & RGB                       & 0.5m-1m     \\
                                                 & RSVQA-LR \cite{lobry2020rsvqa}   & 2020 & 772  &VQA  & 9   & MS                 & 10m     \\
                                                & RSVQA-HR \cite{lobry2019visual}  & 2020 & 10659  &VQA   & 89  & RGB                        & 0.15m         \\
                                                & RSVQAxBEN \cite{lobry2020rsvqa}    & 2021 & 590,326 &VQA     & 26,875    & MS          & 10m    \\
                                                & RSIVQA \cite{lobry2020rsvqa}                              & 2021 & 37,264    & VQA & 864       & RGB                        & 0.1$\sim$8m       \\
                                                 & RS5M~\cite{zhang2023rs5m}                   & 2023             & 5,000,000          & Captioning                      & /          & RGB,Text                  & 0.5m-30m    \\
                                                 & RSICD~\cite{lu2017exploring}                  & 2017             & 10921              & Captioning                      & 31         & RGB,Text                  & 0.5m-30m    \\
                                                 & NWPU-Captions~\cite{cheng2022nwpu}          & 2022             & 31500              & Captioning                      & 45         & RGB,Text                  & 0.5m-30m    \\
                                                 & RSITMD~\cite{yuan2022exploring}                 & 2022             & 4743               & Retrieval                       & 32         & RGB,Text                  & 0.5m-30m    \\
                                                 & RSICap~\cite{hu2023rsgpt}                 & 2023             & 2585               & Captioning                      & 16         & RGB,Text                  & 0.5m-30m    \\
                                                 & SkyScript~\cite{wang2024skyscript}              & 2023             & 2600000            & Captioning                      & /          & RGB,Text                  & 0.5m-30m    \\
                                                 & ChatEarthNet~\cite{yuan2024chatearthnet}           & 2024             & 163488             & Captioning                      & /          & SAR,MS,Text               & 10m-20m   \\
                                                 & LHRS-Align~\cite{muhtar2024lhrs}             & 2024             & 1150000            & Captioning                      & /          & RGB,Text                  & 0.5m-30m    \\
                                                 & LAION-EO~\cite{czerkawski2023laion}              & 2023             & 24933              & Captioning                      & /          & RGB,Text                  & 0.5m-30m   \\ \bottomrule
\end{tabular}}
\label{RSFoundation}
\end{table*}

Among these datasets, image captioning \cite{lu2017exploring,RSICC,zhao2024luojiahog} and Visual Question Answering (VQA) \cite{lobry2020rsvqa,yuan2022change} combine natural language and image data to enable a more natural interaction between end-users and artificial intelligence systems. Datasets for RS Visual Grounding (VG)~\cite{zhan2023rsvg} are constructed to enable specific object detection guided by natural language. In the era of VLMs, large-scale vision-language datasets including RS5M~\cite{zhang2023rs5m}, Skyscript~\cite{wang2024skyscript} and ChatEarthNet~\cite{yuan2024chatearthnet} are built for aligning natural language with RS imagery and fine-tuning VLMs.
\begin{figure}
	\centering
	\includegraphics[width=0.495\textwidth]{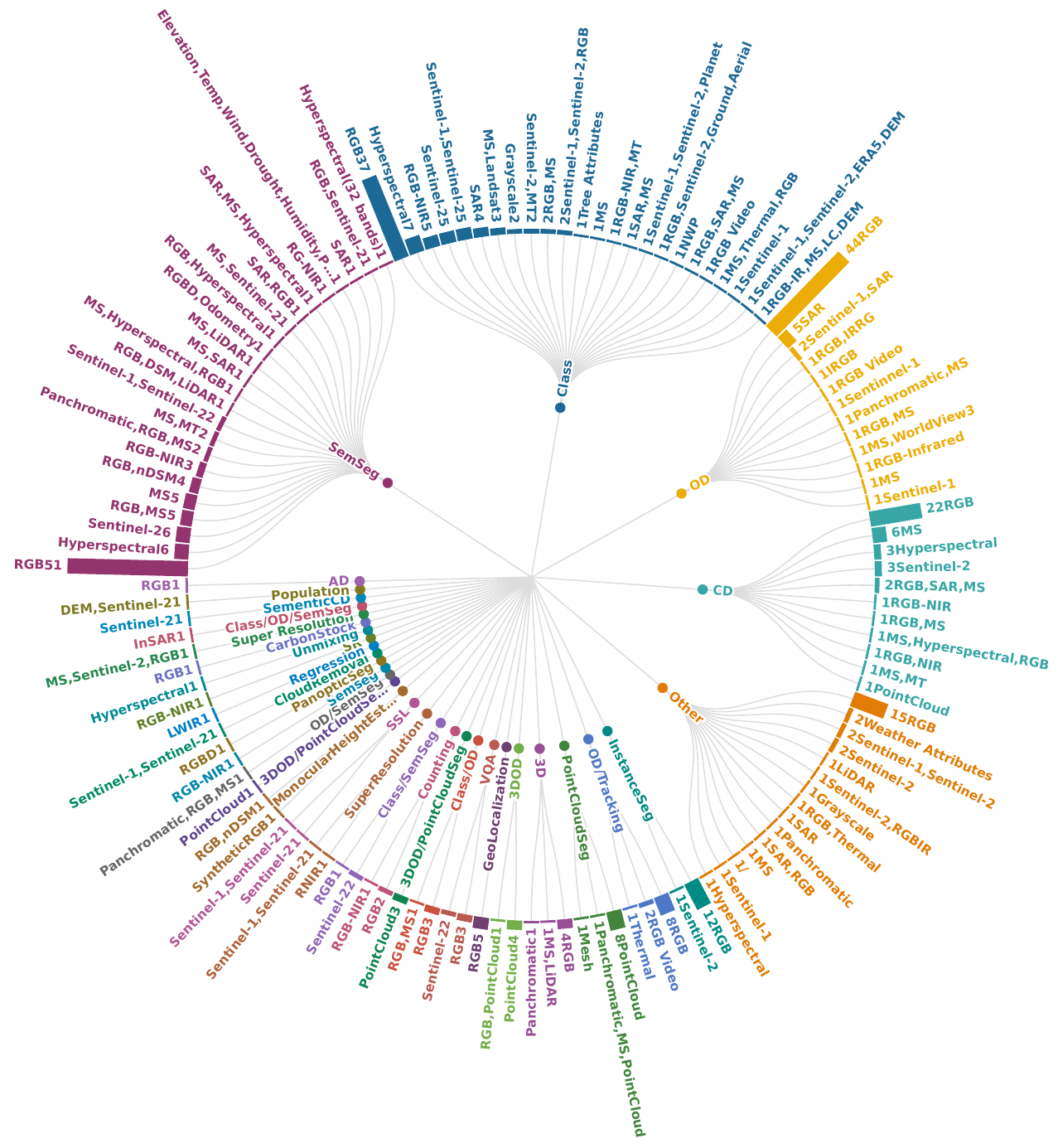}
	\caption{Data modalities (outer perimeter) organized by EO tasks (inner labels). Although a wide range of data sources are used for EO, optical data (RGB) remains the most commonly used modality for the majority of RS tasks.}
	\label{HSB}
\end{figure}

Apart from these reviewed RS tasks, we list some datasets constructed for other tasks in the supplementary materials. Multi-view stereo datasets \cite{SatStereo,liu2020novel,kolle2021hessigheim} are used for 3D reconstruction. There are also datasets used for more sporadic tasks like geo-localization \cite{workman2015wide,zheng2020university}, weather forecasting \cite{kashinath2021climatenet,grnquist2020deep}, soil parameter estimation \cite{Hyperview_Challenge}, and wind speed estimation \cite{Turbines,Tropical}. 

\section{Remote Sensing Dataset Analysis}
\label{ANA}
In this section, we analyze the reviewed over 500 RS datasets and provide statistics related to four different aspects of these datasets.
\begin{figure}
	\centering
	\includegraphics[width=0.494\textwidth]{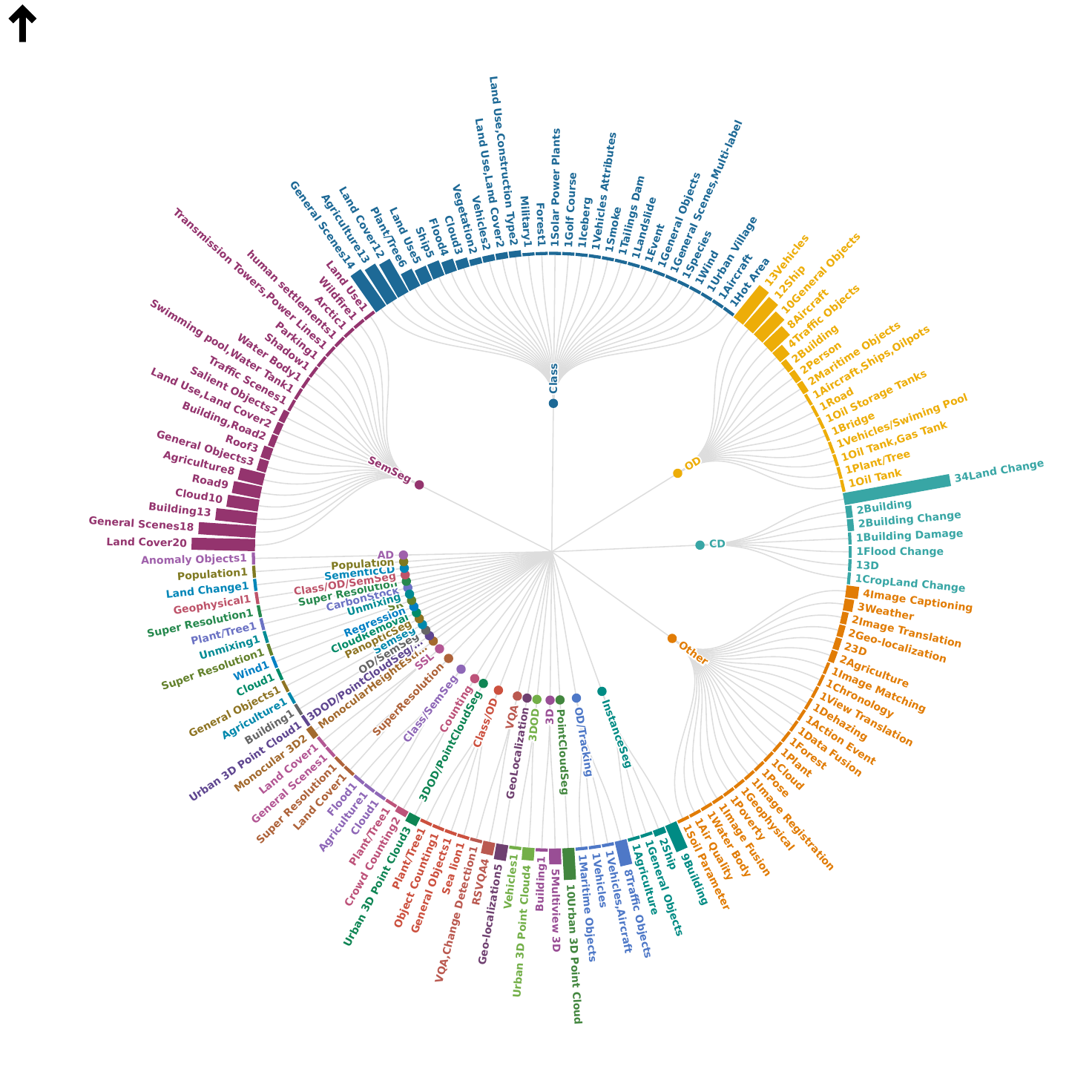}
	\caption{Research Domains (outer perimeter) Organized by EO Tasks (inner labels). It can be seen that there are strong correlations between the research domains and EO tasks.}
	\label{TDomain}
\end{figure}
\begin{figure*}[t]
	\centering
	\includegraphics[width=0.824\textwidth]{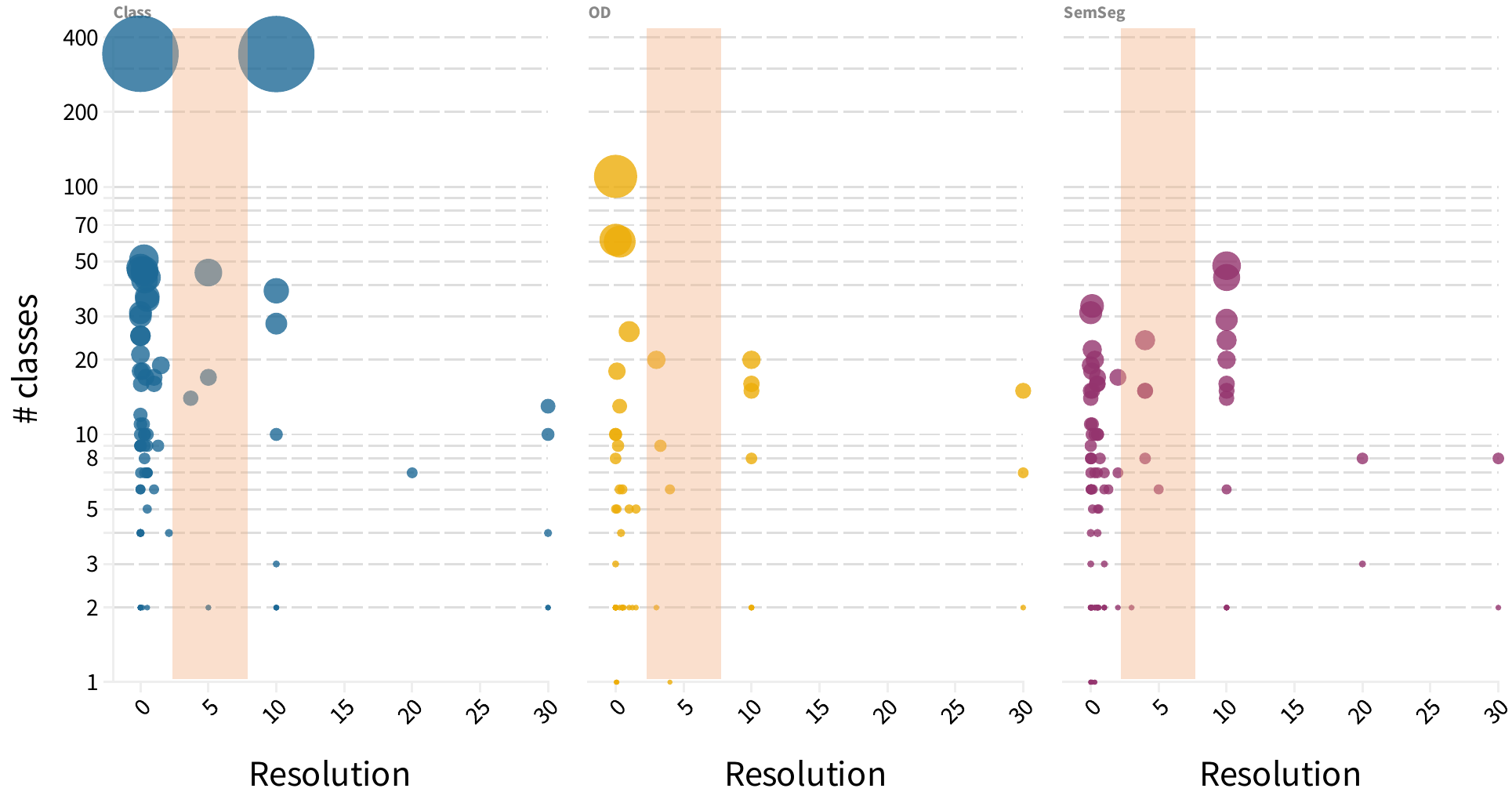}
	\caption{Visualization of the relationships between data resolution and the number of annotated classes. An interesting finding is that most datasets have a resolution that is either smaller than 1m or larger than 10m. Datasets with a resolution range of between 1 to 10m are scarce.}
	\label{RES1}
\end{figure*} 

\subsection{The Volume Trend}
\label{Unit}
Thanks to their powerful representation learning capabilities, deep learning networks trained with large-scale datasets have shown superior performance to classical machine learning methods. In the deep learning era, large-scale datasets are important in training deep models that yield better performance and generalizability. Another advantage of large-scale datasets is that they align better with real-world scenarios. In Fig. \ref{year-vol}, we visualize a chronological overview of the volume of over 500 RS datasets. Note that the volume (in GBs) shown in the figure is transformed into the logarithmic scale, and larger circles indicate larger volumes. It can be seen that datasets before the year 2015 are usually smaller in volume. Similar to the CV community, after deep learning became the mainstream technique, both the number and the volume of RS datasets significantly increased. For example, the volumes of fMoW \cite{fmow2018} and Sen4AgriNet \cite{sykas2021sen4agrinet} are greater than 4000 GB. Well-annotated large-scale datasets can help the RS community develop more powerful deep-learning models with better performance and generalizability. 

\subsection{Analysis of Data Modalities}
To analyze the image sources used in the RS community, we summarize and visualize the data modalities used for different RS tasks. The relationships between data modalities and RS tasks are shown in Fig. \ref{HSB}. Although there is a wide range of image sources, optical data (RGB) is still the most frequently used modality for the majority of RS tasks. In Fig. \ref{TDomain}, we display the relationships between tasks and research domains.

\subsection{Analysis of Spatial Resolutions}
For RS images, spatial resolution has a high correlation to image content. In Fig. \ref{RES1}, we show the relationships between data resolution and the number of annotated classes. In general, this figure clearly shows that the number of semantic classes for RS image classification tasks is higher than those for RS object detection and segmentation tasks. The reason is that object-level and pixel-level datasets require much more annotation efforts when the number of semantic classes increases.

Another interesting finding is that most datasets have a resolution of smaller than 1m or larger than 10m. The datasets with resolution in the range between 1 to 10m are obviously scarce. The reason for this phenomenon is that many EO applications require either high-resolution ($\leq$1m) imagery or global coverage (Sentinel 1\&2, $\geq$10m). However, the EO data with resolution 1$\sim$10m also has great potential in a range of applications. More research attention should be devoted to filling this gap.
\begin{figure*}
	\centering
	\includegraphics[width=0.98\textwidth]{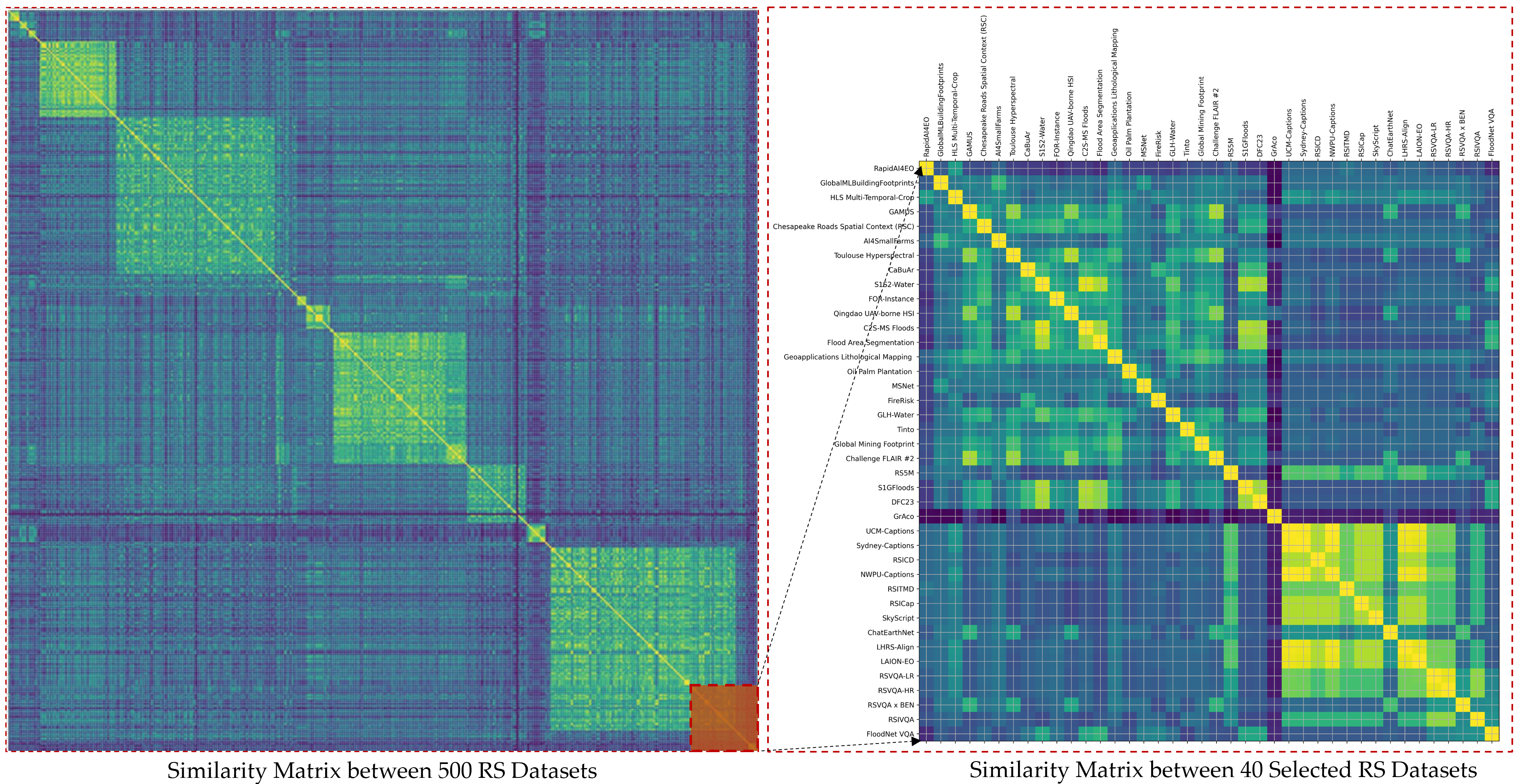}
	\caption{Visualization of the correlation between different datasets. A lighter color means a higher correlation. }
	\label{SIM}
\end{figure*}

\subsection{The correlation between different datasets}
\label{corr}
To provide a more global view of these datasets, for the first time, we propose to analyze the correlation between different datasets based on the attribute information provided in this study.

We treat the attribute information of each dataset as a data sample and measure the similarity between different datasets. Formally, let $n$ represent the number of samples, and $s$ represent the size of each sample in dataset $D$. We denote $v$ as the volume of $D$. Then the scale of $D$ can be quantitatively measured using $n,s,$ and $v$. Furthermore, we quantify the annotation level of $D$ and represent it using $m$. Specifically, we assign 1 to $m$ for image-level annotation, 2 for object-level, 3 for pixel-level, 4 for instance-level, 5 for panoptic-level, and 0 for no-label. Similarly, we also quantify the task of $D$ to $t$. According to the task type, $t$ can be 1 for RS image classification, 2 for object detection, 3 for semantic segmentation, 4 for change detection, and 0 for other tasks. Then we use $c$ to represent the number of annotated classes in dataset $D$, and $r$ to denote the max resolution of samples in $D$. With these definitions, $n,s,v,m,c,r,t$ are numerical values representing the attributes of $D$. 

Since the research domain of a dataset is provided by a word or phrase, it is non-trivial to measure the distance between them. For example, the research domain ``Ship'' should be closer to ``Sea'' than ``Tree'' or ``Aircraft.'' The research domain ``Tree'' should be more similar to ``Forest'' not ``Building.'' To this end, we propose to use the pre-trained word embedding \cite{mikolov2013efficient} models to compute the real-valued vector feature $d$ for each research domain. Here we denote $d$ as the textual features of the domain.

Following these pre-processing pipelines, we can quantify the attributes of dataset $D$ into two feature vectors: $F=[n,s,v,m,c,r,t]$ and the word embeddings for the research domain $d$. For two datasets $D_1$ and $D_2$, we use $F_1$, $F_2$, and $d_1$, $d_2$ to represent the features of these two datasets. Then, we can compute the similarity using the following formula:
\begin{equation}
\begin{aligned}
        \cos (\theta_1 ) &=   \dfrac {F_1 \cdot F_2} {\left\| F_1\right\| _{2}\left\| F_2\right\| _{2}}, \\
        \cos (\theta_2 ) &=   \dfrac {d_1 \cdot d_2} {\left\| d_1\right\| _{2}\left\| d_2\right\| _{2}}, \\
        sim(D_1,D_2) &= \cos (\theta_1 ) + \cos (\theta_2 ).
\end{aligned}
\end{equation}

In Fig. \ref{SIM}, the correlation matrix between over 500 RS datasets is visualized. The lighter the color, the higher the similarity. To our knowledge, this is the first work that analyzes the correlation between existing RS datasets. The correlation matrix reflects the distances between different pairs of the RS dataset. The distance information could be valuable for the RS community. Some possible ways to leverage the correlation between RS datasets for future research are outlined below.
\begin{enumerate}
    \item \textbf{Dataset recommendation}. Similar and related datasets can be recommended based on the relationships between datasets and the given dataset. This will help researchers find desired datasets for their research tasks.
    \item \textbf{Domain adaptation}. Domain adaption aims to improve the performance of a model on a target domain using the knowledge learned in the source domain. With the correlation map, researchers can easily find the proper source and target datasets for developing novel domain adaptation algorithms.
    \item \textbf{Dataset assembling}. The distance between datasets can also be used to assemble multiple small but similar datasets into a larger one (metadataset) for pre-training and evaluating large-scale RS foundation models.
    \item \textbf{Multi-task model training}. Similarly, using the distance between datasets, we can combine datasets with similar spatial resolution, data modalities, or research domains, but different tasks into a unified dataset for training multi-task foundation models.
\end{enumerate}

\begin{figure*}[htbp]
	\centering
	\includegraphics[width=0.98\textwidth]{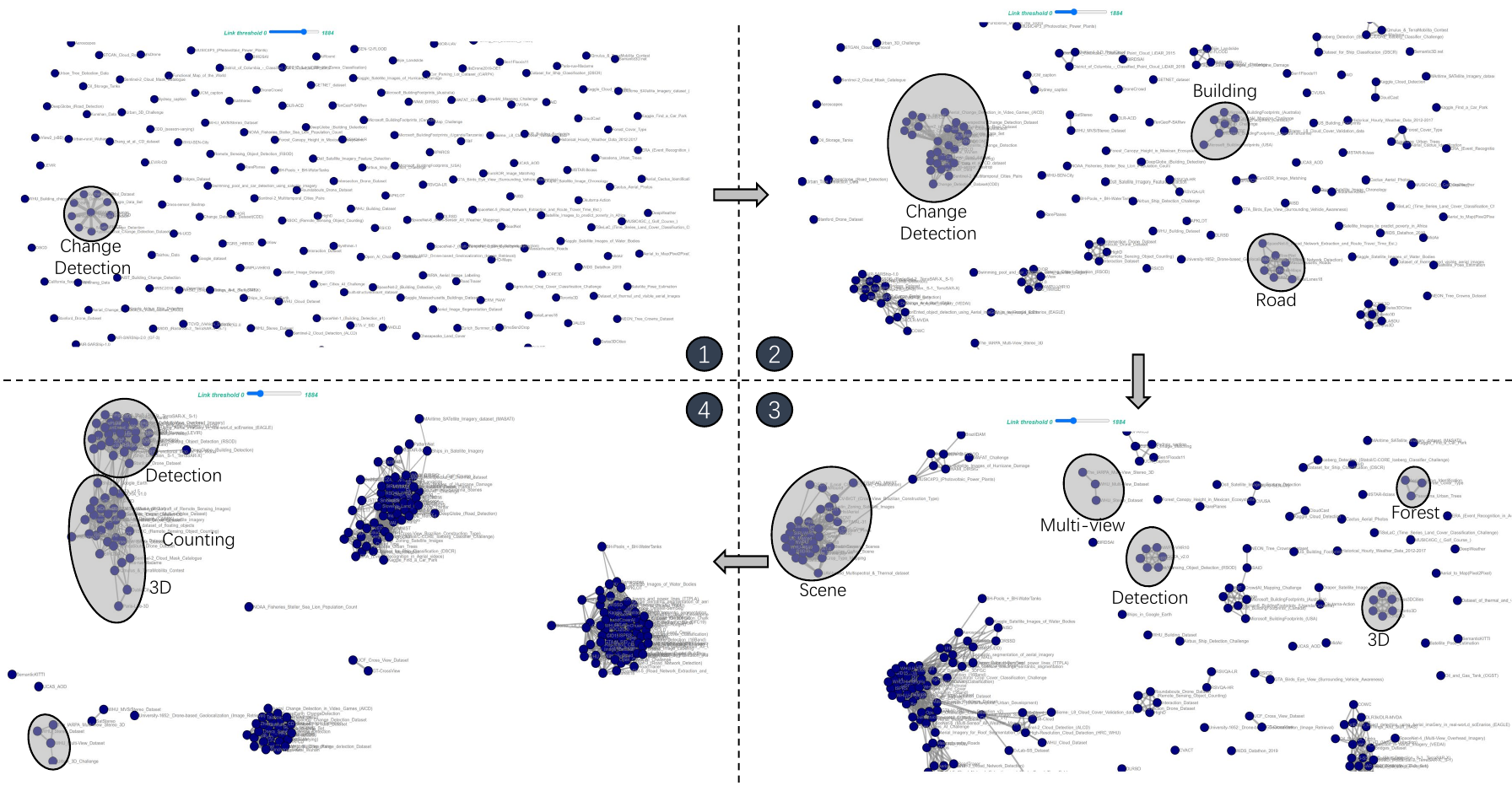}
	\caption{With the correlation matrix, we can visualize RS datasets using network graphs. The node represents the RS dataset, and the link between nodes denotes the similarity between nodes. Datasets gradually cluster together when the connecting threshold decreases.}
	\label{SIMF}
\end{figure*}

Furthermore, with the correlation matrix, we can visualize RS datasets using an interactive network graph. The node represents the RS dataset, and the link between nodes denotes the similarity between them. In Fig. \ref{SIMF}, we can see that different datasets gradually cluster together when the connecting threshold decreases.
\section{Dataset Ranking and Benchmark Building}
\label{RANK}
Researchers in the RS community have been publishing more and more datasets to benefit the development of new methods. However, algorithms can easily saturate their performance on these datasets \cite{dimitrovski2022aitlas}. Deep learning models can achieve almost perfect performance on small-scale or domain-specific datasets. However, small-scale datasets are more likely to have bias and cannot reflect the performance of methods in real-world complex scenarios \cite{long2021creating}. Methods developed on small datasets or specific domains are difficult to generalize to other scenarios. Considering these disadvantages, it is urgent to employ new benchmarks with large-scale, general research domains, and high-quality annotations for a fair and consistent evaluation of RS methods. Although the attributes of a large number of datasets are provided, it is still not intuitive to compare the quality of different datasets. Thus, for the first time, in this study, we propose to rank these datasets based on their attributes.

\subsection{Dataset Ranking Metrics}
Regarding the desirable properties of benchmark datasets, Long et al. \cite{long2021creating} propose the DiRS formula, so named for its focus on the diversity, richness, and scalability of datasets. These properties are good references for designing metrics to measure and rank the RS dataset. However, it is non-trivial to quantitatively measure the diversity and richness of existing datasets. In order to approximate the DiRS metric, we consider both data diversity and annotation diversity in this study.

To measure the data diversity, we first examine the research domain of one dataset. 
Some datasets constructed with specific domains will have limitations on the diversity of data sources. Hence, we first filter them and only keep datasets designed for general purposes, like LULC or general scene understanding. Next, we choose to measure the dataset scale using the number, size of samples, and the volume of the dataset, i.e., attribute variables $n,s,v$. Furthermore, we take the modality diversity, that is, the number of data modalities $k$, into consideration. Since images with higher spatial resolution can provide richer visual content, we also factor in the resolution $r$ as a part of the metric.

Considering the annotation richness, we use the number of annotation classes $c$ and the quantified annotation level $m$ to measure the richness of the labels. To start with, given 401 RS datasets, we first filter out datasets designed for specific domains. After this step, 114 datasets remain as candidates. Then, we use the aforementioned attributes to quantitatively measure the diversity and richness of each dataset. Since there exist significantly high values in the $n,s,v$ of different datasets, we use log normalization to standardize them into the range 0 to 1. Next, we normalize each of the dataset attributes in $r,m,c,k$ into the range 0 to 1. Finally, we add them together to form the final score for the given dataset. 

Based on the measurement defined above, we can compute the scores and rank these RS datasets. Based on the rankings, we select datasets for three different tasks. Specifically, for the RS image classification task, the top five ranked datasets are: 1). fMoW \cite{fmow2018}, 2) BigEarthNet \cite{sumbul2019bigearthnet}, 3) Million AID \cite{long2021creating}, 4) So2Sat LCZ42 \cite{zhu2019so2sat}, and 5) RSD46-WHU \cite{xiao2017high}. For the RS object detection task, the top five datasets are 1) fMoW \cite{fmow2018},  2) DIOR \cite{li2020object}, 3) xView \cite{xview2018}, 4) DOTA v2.0 \cite{9560031}, and 5) TGRS HRRSD \cite{zhang2019hierarchical}. Finally, for the RS semantic segmentation task, the top five ranked datasets are 1) SEASONET \cite{kossmann2022seasonet}, 2) OpenSentinelMap \cite{johnson2022opensentinelmap}, 3) SEN12MS \cite{schmitt2019sen12ms}, 4) GeoNRW \cite{baier2021synthesizing}, and 5) Five-Billion-Pixels \cite{2022FBP}. A complete list of the charts is displayed on \url{https://earthnets.github.io}, where the radar charts are used to compare the attributes of some top-ranked datasets.

\subsection{Dataset Selection for Benchmarking}
We aim to bridge the gap between CV and RS communities. In CV research, image classification, object detection, and semantic segmentation stand out as extensively developed tasks, with numerous state-of-the-art models proposed. Consequently, in this work, we select these three tasks for model benchmarking.
Regarding datasets, we aim to select several datasets designed with general purpose, large diversity, and high richness for developing and evaluating deep learning methods. Although many large-scale RS datasets meet these standards, it is unacceptable and not environment-friendly to benchmark all of them for the evaluation of RS algorithms. Thus, in this study, we choose to select two datasets for each task, including one with high resolution and one with low resolution for larger geographical coverage.

Following this constraint, the following datasets are selected. 1) \textbf{fMoW} with high resolution ($\sim$1m) data and \textbf{BigEarthNet} with low resolution (\textgreater10m) imagery are selected for image classification. 2) \textbf{DIOR} with high-resolution ($\sim$1m) data and \textbf{fMoW} with large objects are selected for the RS object detection task. 3) \textbf{GeoNRW} with high-resolution ($\sim$1m) images and \textbf{SEASONET} with low-resolution (\textgreater10m) images are selected for RS semantic segmentation. In total, there are five datasets selected to build a unified benchmark for three different tasks.

\begin{figure}
	\centering
	\includegraphics[width=0.498\textwidth]{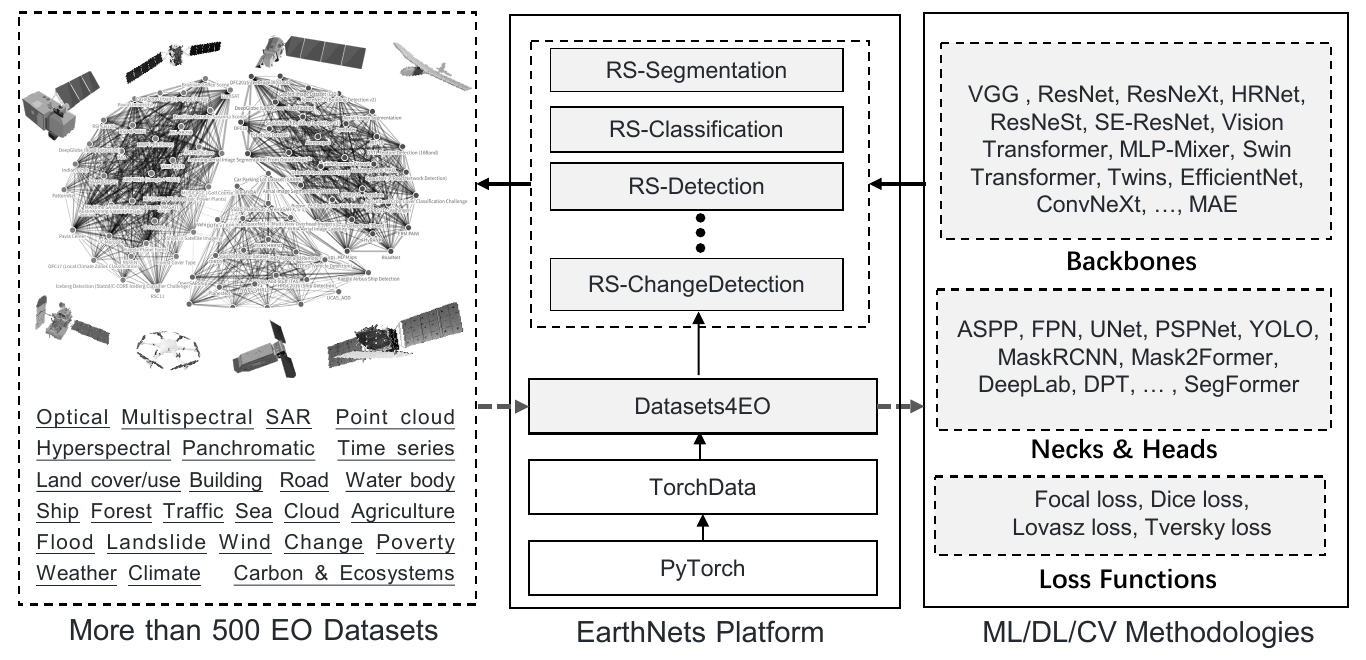}
	\caption{The architecture design of the proposed EarthNets platform. EarthNets is based on PyTorch \cite{paszke2019pytorch} and TorchData. It contains the Dataset4EO for a standard and easy-to-use dataset-loading library and some high-level libraries for different EO tasks.}
	\label{EarthNets}
\end{figure}

\section{The EarthNets Open Platform}
\label{ETN}
Large-scale, high-quality datasets are important for a faithful evaluation of RS algorithms, while other factors like training tricks, hyper-parameters, optimizers, and initialization methods are also critical for a fair and reliable comparison of different methods. Thus, an open platform is crucial for the fair evaluation, reproducibility, and efficient development of novel methods. However, there is still no unified deep-learning platform for different RS tasks. Torchgeo \cite{stewart2021torchgeo} mainly focuses on the data loading part. AiTLAS\cite{dimitrovski2022aitlas} mainly contains the codebase for the RS classification task. In contrast, we aim to build a new unified platform for the RS community that not only deals with dataset loading but also includes libraries for different RS tasks.

Fig. \ref{EarthNets} illustrates the overall architecture design of the proposed EarthNets platform. The platform is based on PyTorch \cite{paszke2019pytorch} and TorchData. The library Dataset4EO is designed as a standard and easy-to-use dataset-loading library. Note that Dataset4EO can be used alone or together with our high-level libraries, like RS-Classification, RS-Detection, etc.

For the design of the EarthNets platform, we consider two main factors. The first is the decoupling between dataset loading and high-level EO tasks. As shown in this study, there are over 500 RS datasets with different file formats, data modalities, research domains, and download links. Building a standard and scalable dataset-loading library can largely accelerate research for the whole RS community. Furthermore, researchers from other machine-learning communities can also benefit from the standard dataset-loading library. The second considered factor is pushing the RS data to a larger machine-learning community. There are several novel deep learning models published in the CV and machine learning community, including different backbones, models, and loss functions. The EarthNets platform is designed to easily apply these models to RS datasets and to fill in the gap between the RS and CV communities.

\section{Experiments}
\label{EXP}
In this section, we benchmark state-of-the-art deep learning models from the CV community on five selected RS datasets. We also compare them with the methods specifically designed for the RS datasets. The implementation details can be found in the supplementary materials.

\begin{table}[]
	\centering
	\caption{Image Classification Results on the fMoW \cite{fmow2018} Dataset. Top-1 accuracy, precision, recall, and F1 score are reported. The best results are in bold.}
	\scalebox{1.0}{
\begin{tabular}{c|ccccc}
\hline \hline
                          & \multicolumn{5}{c}{Image Classification}                                                      \\ \cline{2-6} 
\multirow{-2}{*}{Methods} & Pre-trained & Top-1          & P              & R              & F1                           \\ \hline 
ViT-Small                & Random      & 54.1           & 53.45          & 51.8           & 52.03                        \\
MLP-Mixer                & Random      & 43.11          & 40.33          & 41.09          & {40.18} \\ \hline
ResNet-50               & ImageNet    & 58.25          & 58.73          & 57             & 57.28                        \\
EfficientNet-b4         & ImageNet    & 58.8           & 58.7           & 57.01          & 57.33                        \\
ConvNext-Small             & ImageNet    & 62.05          & 63.81          & 60.34          & 61.19                        \\
Swin-Tiny              & ImageNet    & \textbf{66.42} & \textbf{66.33} & \textbf{65.29} & \textbf{65.5}                \\ \hline \hline
\end{tabular}}
\label{T-fmow}
\end{table}

\textbf{Metrics}: For multi-label image classification datasets, we report the following metrics: precision (P), recall (R), F1 score, and mean average precision (mAP). For precision, recall, and F1, we set the threshold value to 0.5 for all models. For object detection, mAP is used as the measurement for performance evaluation. Three metrics are used to evaluate the semantic segmentation task: overall (micro-averaged) Accuracy (aAcc), mean (macro-averaged) Accuracy(mAcc), and mean Intersection over Union (mIoU).

\begin{table}[]
	\centering
	\caption{Multi-label Image Classification Results on the BigEarthNet\cite{sumbul2019bigearthnet} Dataset. mAP, micro precision, micro recall and F1 score are reported. The best results are in bold.}
	\scalebox{1.0}{
\begin{tabular}{c|ccccc}
\hline \hline
                          & \multicolumn{5}{c}{Image Classification}                                                      \\ \cline{2-6} 
\multirow{-2}{*}{Methods} & Pre-trained & mAP            & P              & R              & F1                           \\ \hline
ResNet-18*\cite{manas2021seasonal}            & Random      & 79.80          & -              & -              & -                            \\
ResNet-18*\cite{manas2021seasonal}               & ImageNet    & 85.90          & -              & -              & -                            \\
ResNet-18*\cite{manas2021seasonal}              & MoCo-v2     & 85.23          & -              & -              & -                            \\
ResNet-50*\cite{manas2021seasonal}               & ImageNet    & 86.74          & -              & -              & -                            \\ \hline
ResNet-50              & ImageNet    & 85.74          & 76.87          & 75.89          & {76.38} \\
EfficientNet-b4      & ImageNet    & 84.48          & 73.84          & 77.19          & 75.48                        \\
ConvNext-Small          & ImageNet    & 85.59          & 73.64          & 79.91          & 76.65                        \\
Swin-Tiny          & ImageNet    & \textbf{87.19} & \textbf{78.01} & \textbf{80.22} & \textbf{79.1}                \\
MLP Mixer          & ImageNet    & 82.76          & 73             & 74.36          & 73.67                        \\ \hline \hline
\end{tabular}}
\label{T-BigEarthNet}
\end{table}
\subsection{Benchmarking Results and Comparisons}
In this section, we benchmark the five selected datasets using the proposed EarthNets platform. To avoid excessive computation costs, we choose to evaluate some representative state-of-the-art (SOTA) methods from the CV community on large-scale RS datasets.

\begin{table}[]
	\centering
	\caption{Object Detection Results on the DIOR\cite{li2020object} Dataset. The mAP performance is reported. The best results are in bold.}
	\scalebox{1.0}{
	\begin{tabular}{c|cccc}
	\hline \hline
\multirow{2}{*}{Method} & \multicolumn{4}{c}{Object Detection}                                                             \\ \cline{2-5}
                        & Backbone  & \multicolumn{1}{l}{Optimizer} & \multicolumn{1}{l}{Epochs} & \multicolumn{1}{l}{mAP} \\ \hline
RetinaNet* \cite{li2020object}              & ResNet-50  & -                             & -                          & 65.7                    \\
RetinaNet* \cite{li2020object}             & ResNet-101 & -                             & -                          & 66.1                    \\
PANet*  \cite{li2020object}                & ResNet-50  & -                             & -                          & 63.8                    \\
PANet*  \cite{li2020object}               & ResNet-101 & -                             & -                          & 66.1                    \\
Mask-RCNN* \cite{li2020object}             & ResNet-50  & -                             & -                          & 63.5                    \\
Mask-RCNN*  \cite{li2020object}            & ResNet-101 & -                             & -                          & 65.2                    \\
YoloV3* \cite{li2020object}             & DarkNet53 & -                             & -                          & 57.1                    \\ \hline
YoloV3                  & DarkNet53 & SGD                           & 120                        & 64.0                    \\
YoloV3                  & Swin-Tiny & AdamW                         & 120                        & 64.6                    \\
YoloV3                  & ConvNext-Small  & AdamW                         & 120                        & 67.6                    \\
Mask-RCNN               & ResNet-50  & SGD                           & 120                        & 68.5                \\
Mask-RCNN               & Swin-Tiny  & AdamW                         & 120                        & 70.5                \\
Mask-RCNN               & ConvNext-Small  & AdamW                         & 120                        & \textbf{72.4}                   \\ \hline \hline 
\end{tabular}}
\label{T-DIOR}
\end{table}

\begin{table}[]
	\centering
	\small	
	\caption{Semantic Segmentation Results on the SEASONET\cite{kossmann2022seasonet} Dataset. We report the aAcc, mAcc, and mIoU metrics. The best results are in bold.}
	\scalebox{0.72}{
	\begin{tabular}{c|cccccc}
	\hline \hline
\multirow{2}{*}{Method} & \multicolumn{6}{c}{Semantic Segmentation}               \\ \cline{2-7}
                        & Backbone    & Optimizer & Iter. & aAcc  & mAcc  & mIoU  \\ \hline
DeeplabV3* \cite{kossmann2022seasonet}             & DenseNet121 & --        & --    & --    & --    & 47.53 \\
DeeplabV3,PT* \cite{kossmann2022seasonet}        & DenseNet121 & --        & --    & --    & --    & 48.69 \\ \hline
DeeplabV3               & ResNet-50    & SGD       & 80k   & 82.87 & 58.49 & 47.5  \\
DeeplabV3               & ResNet-50    & SGD       & 160k  & 83.52 & 62.65 & 50.79 \\
DeeplabV3               & ConvNext-Small    & AdamW     & 120k   & 81.36 & 56.31 & 46.39 \\
DeeplabV3               & Swin-Tiny    & AdamW     & 120k   & 82.75 & 61.5  & 50.81 \\
Upernet                 & ResNet-50    & SGD       & 120k   & 83.2 &  60.36  & 49.59 \\
SegFormer               & MiT    & AdamW     & 120k   & \textbf{83.75} & \textbf{64.25}  & \textbf{53.87} \\
\hline \hline
\end{tabular}}
\label{T-SEASONET}
\end{table}

\begin{table}[]
	\centering
	\small	
	\caption{Semantic Segmentation Results on the GeoNRW\cite{baier2021synthesizing} Dataset. We report the aAcc, mAcc, and mIoU metrics. The best results are in bold.}
	\scalebox{0.71}{
	\begin{tabular}{c|cccccc}
	\hline \hline
                         & \multicolumn{6}{c}{Semantic Segmentation}                                                                \\ \cline{2-7}
\multirow{-2}{*}{Method} & Backbone      & Optimizer                  & \#Epochs & aAcc           & mAcc           & mIoU           \\  \hline
MultiTask* \cite{lu2022multi}                 & Transformer    & AdamW                      & 100k    & 76.75          & 71.89          & 57.3           \\
Lu et al.* \cite{lu2022multi}         & Transformer   & AdamW                      & 100k    & 76.53          & 70.12          & 56.2           \\ \hline
FCN                      & UNet          & SGD                        & 40k       & 78.8           & 66.86          & 55.6           \\
PSPNet                   & ResNet-50           & SGD                        & 40k       & 81.56          & 74.92          & 62.73          \\
Deeplabv3+               & ResNet-50           & SGD                        & 40k       & 81.91          & 75.26          & 63.01          \\
Deeplabv3+               & ConvNext-Tiny & AdamW                      & 40k       & 80.89          & 73.61          & 61.63          \\
Deeplabv3+               & Swin-Tiny        & SGD & 40k       & 78.09          & 70.01          & 56.78          \\
Deeplabv3+               & Swin-Tiny        & AdamW                      & 40k       & 81.11          & 74.28          & 62.18         \\
Upernet                  & ResNet-50           & SGD                        & 40k       & 81.87          & 75.7           & 63.1           \\
Upernet                  & ConvNext-Tiny & AdamW                      & 40k       & 82.08          & 74.9           & 63.48          \\
Upernet                  & Vit-Small     & AdamW                      & 40k       & 78.65          & 71.13          & 59.43          \\
Upernet                  & Swin-Tiny     & AdamW                      & 40k       & 82.31          & \textbf{75.68} & \textbf{64.48} \\
SegFormer                & MiT           & AdamW                      & 40k       & \textbf{82.55} & 75.63          & 64.38          \\ \hline \hline
\end{tabular}}
\label{T-GEONRW}
\end{table}

\textbf{Comparisons on the fMoW Dataset.} 
fMoW \cite{fmow2018} is a large-scale dataset built to recognize the functional purpose of buildings and land use. It contains 1 million images from over 200 countries, annotated with 63 different classes. In this study, we use the fMoW-rgb version of the dataset for model evaluation. Table \ref{T-fmow} reports the benchmarking results. In general, we can see that using the ImageNet pre-trained weights can greatly improve the performance. When we compare the CNN-based methods with the Transformer-based method, we find that Swin-Tiny \cite{liu2021swin} clearly outperforms other CNN-based methods in all four metrics. Among the CNN-based methods, ConvNext \cite{liu2022convnet} is the best-performing one. 

\textbf{Comparisons on the BigEarthNet Dataset.} 
BigEarthNet is a large-scale multi-label Sentinel-2 benchmark dataset annotated with the CORINE Land Cover classes. There are two versions of the labels, one with 43 categories and another with 19 categories. In this study, we adopt the new class nomenclature (19 categories) introduced in \cite{sumbul2020bigearthnet}. Regarding the methods, we evaluate four CNN-based architectures (ResNet-18, ResNet-50, EfficientNet-b4, ConvNext). For the Transformer-based method, we evaluate the Swin-Tiny, which is usually overlooked in existing benchmarking results. Furthermore, an MLP-based method, the MLP-Mixer \cite{tolstikhin2021mlp} is also compared. Additionally, we also compare the results reported by existing work \cite{manas2021seasonal} on the BigEarthNet dataset. 

Table \ref{T-BigEarthNet} reports the benchmarking results. In general, the results indicate that 
Swin-Tiny performs best on this multi-label classification dataset. However, there is no significant advantage compared with other CNN-based methods. Another conclusion we can make is that ResNet-50 is a strong baseline method. From the results, it can be seen that ResNet-50, pre-trained on ImageNet or using self-supervised MoCo-V2 \cite{chen2020improved}, can perform better than MLP-Mixer, EfficientNet-b4 on this dataset. The performance of ConvNext is competitive to ResNet-50, but lower than the transformer-based method Swin-Tiny. Note that $^*$ indicates that the results of the method are reported in existing work. Generally speaking, the results benchmarked using the EarthNets platform are higher than or comparable to existing reported results. 

\textbf{Comparisons on the DIOR Dataset.} 
Table \ref{T-DIOR} presents the benchmarking results on the DIOR dataset built for the object detection task. We chose two representative and widely-used object detection methods designed by the CV community. To be specific, YoloV3\cite{redmon2016you} and Mask-RCNN\cite{he2017mask} are evaluated on the DIOR dataset. YoloV3 is designed for lightweight and real-time object detection. Mask-RCNN is an extension of Faster-RCNN \cite{ren2015faster} with ROI alignment and a third segmentation branch. The experimental results reveal that Mask-RCNN performs better than YoloV3 on this dataset.  Concerning different backbones, the results clearly show that Swin-Tiny and ConvNext can outperform other compared methods. The Mask-RCNN method with ConvNext backbone achieves an mAP of 72.4\%, which is 6.3 percentage points higher than the best results reported in \cite{li2020object}. Notably, we observe that our benchmarked results can greatly outperform the same method reported in existing work. This comparison reveals that the choice of the optimizer, hyper-parameters, or other training tricks can greatly affect the final results, even when the same method is used.

\textbf{Comparisons on the SEASONET Dataset.} 
SEASONET is a large-scale multi-label LULC scene understanding dataset. It includes 1,759,830 images from Sentinel-2 tiles, and can be used for scene classification, segmentation, and retrieval tasks.
In this study, we evaluate segmentation performance on this dataset. On this dataset, we evaluate the widely-used semantic segmentation method DeeplabV3 \cite{chen2017rethinking} with three different backbones: ResNet-50, ConvNext, and Swin-Tiny. Upernet and SegFormer \cite{xie2021segformer} with mixed-Transformer encoders (MiT) are also compared. Table \ref{T-SEASONET} reports the benchmarking results. It can be seen that ResNet-50 and Swin-Tiny obtain comparable results and clearly surpass other backbones. SegFormer with MiT encoder clearly outperforms other models. We also find that the results obtained using EarthNets significantly outperform performance reported in existing work \cite{kossmann2022seasonet}.

\textbf{Comparisons on the GeoNRW Dataset.} 
The benchmarking results on the GeoNRW dataset are displayed in Table \ref{T-GEONRW}. Five segmentation methods, FCN \cite{longfcn}, DeeplabV3+ \cite{chen2018encoder}, PSPNet \cite{zhao2017pyramid}, Upernet \cite{xiao2018unified} and SegFormer \cite{xie2021segformer} with mixed-Transformer encoders (MiT), are evaluated on this dataset. We observe that Transformer-based models like SegFormer and Swin Transformer perform better than other methods. However, the performance of ViT-Small is worse than ResNet-50. In comparison to the reported results in existing work, we can find that using the EarthNets platform can obtain clearly better performance.

\section{Conclusion}
\label{CONC}
In this study, we present a comprehensive review and build a taxonomy for more than 500 publicly published datasets in the remote sensing community. Based on the attribute information of these datasets, we systemically analyze them concerning four aspects: volumes, resolution distributions, research domains, and the correlation between datasets. Next, a new benchmark including five selected large-scale datasets is built for model evaluation. A deep learning platform termed EarthNets is released to support a consistent evaluation of deep learning methods on remote sensing data. We further use the EarthNets platform to benchmark state-of-the-art methods on the new benchmark. The performance comparisons are insightful for future research.

\clearpage

\section*{Appendix}
{In this supplementary material, we present the datasets constructed for Earth observation that are not included in the image classification, object detection, semantic segmentation, and change detection tasks. Next, the implementation details of 
the proposed EarthNets platform and the benchmarking experiments are introduced. Finally, more visualization results and detailed benchmarking results are provided.}

\section{Other Earth Observation Tasks}
Image classification, object detection, semantic segmentation, and time-series modeling (change detection) are four fundamental tasks in both the computer vision and remote sensing research community. Although many real-world applications can be framed as one or more of these four tasks, there are still some important Earth observation tasks that do not fall into these categories. To provide a comprehensive review of these datasets, in this study, we arrange them into 30 different research domains, as presented in Table \ref{RSOTHER}. A detailed list of the datasets can be retrieved via the website page\footnote{\url{https://earthnets.retool.com/embedded/public/676aa812-0dca-4e3b-a596-b043d852571d}}.
\begin{table*}[]
\centering
\caption{Detailed Information of some other RS Datasets. These datasets are grouped into 30 different research domains in alphabetical order. The download links for all these datasets can be found at \url{https://earthnets.github.io}.}
\small
\scalebox{0.68}{
\begin{tabular}{c|ccccccccc}
\hline \hline
Domain                        & Name                                  & Year & \#Samples & Sample Size                & \#Classes & Modailty             & Resolution                 & Vol.(GB) \\  \hline
{Action Event} 
                        & Okutama-Action \cite{barekatain2017okutama}                      & 2017 & 77000     & 3840×2160     & 12        & RGB                        & UAV@10$\sim$45m  & 25.9      \\  \hline
\multirow{2}{*}{Agriculture}           & Paddy Rice Maps South Korea \cite{hyun_woo_jo_2022_5845896}         & 2022 & 12942     & 256           & /         & Sentinel-1                 & 10m              & 0.198     \\
                                       & Paddy Rice Labeling South Korea \cite{chanwoo_kim_2022_5846018}     & 2022 & /         & /             & /         & Sentinel-2                 & /                & 0.0016    \\ \hline
Air Quality                            & Air Quality e-Reporting \cite{EEA_Air}             & 2021 & /         & /             & /         & /                          & /                & /         \\  \hline
Anomaly Objects                        & Aerial Anomaly Detection \cite{jin2022anomaly}            & 2022 & /         & /             & 2         & RGB                        & UAV              & /        \\  \hline
Building                               & Urban 3D Challenge \cite{Urban3D2017}                  & 2017 & 157,000   & /             & 2         & RGB                        & 0.5m             & /      \\  \hline
Chronology                             & Draper Satellite Image Chronology \cite{Kaggle_Chronology}   & 2016 & 1,000     & 3100x2329     & /         & RGB                        & /                & 36     \\   \hline
\multirow{2}{*}{Cloud}                 & STGAN Cloud Removal \cite{DVN/BSETKZ_2019}                  & 2019 & 217190    & 256           & 2         & Sentinel-2,RGBIR           & 10m              & 1.5    \\
                                       & SEN12MS-CR \cite{ebel2020multisensor}                          & 2020 & 122218    & 256           & /         & Sentinel-1,Sentinel-2      & 10$\sim$60m      & 272    \\   \hline
\multirow{3}{*}{Counting}              & DLR-ACD \cite{bahmanyar2019mrcnet}                             & 2019 & 33        & 4458          & 2         & RGB                        & 0.045$\sim$0.15m & /        \\
                                       & DroneCrowd \cite{9573394}                          & 2020 & 3360      & 1920x1080     & 2         & RGB                        & UAV              & 1        \\
                                       & RSOC \cite{gao2020counting}                                & 2020 & 3057      & 2500          & 4         & RGB                        & /                & 0.082      \\    \hline
\multirow{2}{*}{Data Fusion}           & SEN1-2 \cite{schmitt2018sen1}                              & 2018 & 282384    & 256           & /         & Sentinel-1,Sentinel-2      & 10m              & 42.68   \\
                                       & QXS-SAROPT \cite{huang2021qxs}                          & 2021 & 40000     & 256           & /         & SAR,RGB                    & 1m               & 2.7   \\   \hline
Dehazing                               & SateHaze1k \cite{huang2020single}                          & 2017 & 1,200     & 512           & /         & RGB                        & 3m               & 1.2     \\   \hline
Forest                                 & Forest Canopy Height \cite{juan_miguel_requena_mullor_2019_3468645}                & 2018 & 1105      & /             & /         & LiDAR                      & 1m               & 62.9   \\   \hline
\multirow{6}{*}{Geo-localization}      & CVUSA \cite{workman2015wide}                               & 2017 & 44,416    & 750x750       & 0         & RGB                        & 0.3m             & 15.7   \\
                                       & University-1652 \cite{zheng2020university}                     & 2020 & 146,580   & /             & /         & RGB                        & /                & /     \\
                                       & UCF Cross View Dataset \cite{tian2017cross}              & 2017 & 35404     & 1200x1200     & 0         & RGB                        & /                & 59.5    \\
                                       & CVACT \cite{Liu_2019_CVPR}                               & 2019 & 128334    & 1200x1200     & 0         & RGB                        & 0.5m             & 152.8    \\
                                       & University-1652 \cite{zheng2020university}                     & 2020 & 50220     & 1024x1024     & 0         & RGB                        & /                & 58.4    \\
                                       & VIGOR \cite{zhu2021vigor}                               & 2021 & 90618     & 640x640       & 0         & RGB                        & 0.114m           & 94.2   \\     \hline
Geophysical                            & TenGeoP-SARwv  \cite{wang2019labelled}                      & 2019 & 37000     & /             & 10        & SAR                        & 5m               & 31.7    \\    \hline
Image Matching                         & EuroSDR Image Matching \cite{cavegn2014benchmarking}              & 2014 & /         & /             & /         & RGB                        & 6$\sim$13m       & /       \\   \hline
Image Registration                     & Dataset of thermal and visible \cite{thermal}      & 2019 & 110       & 336, 4000     & 0         & RGB,Thermal                & /                & 0.31     \\   \hline
\multirow{2}{*}{Image Translation}     & Aerial to Map(Pixel2Pixel) \cite{pix2pix2017}          & 2017 & 2,194     & 600           & /         & RGB                        & /                & 0.239   \\
                                       & WHU-SEN-City \cite{whu_sen_city}                        & 2019 & 18542     & 256           & /         & Sentinel-1,Sentinel-2      & 10m              & 4.3     \\   \hline
\multirow{6}{*}{Multiview 3D}          & The IARPA Multi-View Stereo 3D \cite{facciolo2017automatic}      & 2017 & /         & /             & /         & MS,LiDAR                   & 0.3m             & 75     \\
                                       & SatStereo \cite{SatStereo}                            & 2019 & 144       & /             & /         & Panchromatic               & 0.5m             & 127       \\
                                       & WHU MVS/Stereo Dataset \cite{liu2020novel}              & 2020 & 1776      & 5376          & /         & RGB                        & 0.1m             & 95.7    \\
                            & WHU TCL SatMVS \cite{Gao_2021_ICCV}                       & 2021 & 300       & 5,120         & /         & Panchromatic               & 2.1$\sim$2.5m    & 29        \\
                                       & WHU Multi-View Dataset \cite{Gao_2021_ICCV}              & 2020 & 28400     & 768           & /         & RGB                        & 0.1m             & 12.3     \\
                                       & WHU Stereo Dataset \cite{Gao_2021_ICCV}                  & 2020 & 21868     & 768           & /         & RGB                        & 0.1m             & 8.4  \\   \hline
\multirow{3}{*}{Plant/Tree}            & Cactus Aerial Photos \cite{lopez2019columnar}                & 2018 & 24,000    & 32            & 2         & RGB                        & /                & 0.053    \\
                                       & ReforesTree \cite{reiersen2022reforestree}                         & 2022 & 105       & 4000          & /         & RGB                        & 0.02m            & 7.5      \\
                                       & Urban Tree Detection Data \cite{ventura2022individual}           & 2020 & 60        & 256           & 2         & RGB-NIR                    & 0.6m             & 0.114    \\   \hline
Population                             & So2Sat Population \cite{doda2022so2sat}                   & 2022 & /         & /             & /         & DEM,Sentinel-2             & 10m              & /        \\   \hline
Pose                                   & Satellite Pose Estimation \cite{park2022speed+}           & 2019 & 15303     & 1920x1200     & /         & Grayscale                  & /                & 4.6      \\   \hline
Poverty                                & Poverty in Africa \cite{kaggle_poverty}                   & 2020 & 32823     & 256           & /         & RGB                        & /                & 5.58    \\   \hline
Soil Parameter                         & Hyperview Challenge \cite{Hyperview_Challenge}                  & 2022 & 2,886     & 11$\sim$250   & /         & Hyperspectral              & 2m               & 1.85     \\    \hline
\multirow{4}{*}{Super Resolution}      & ALSAT-2B \cite{djerida2021new}                            & 2021 & 5518      & 256           & /         & RGB-NIR                    & 2.5m/10m         & 0.044    \\
                                       & The WorldStrat \cite{julien_cornebise_2022_6810792}                      & 2022 & 3449      & 1054          & 8         & MS,Sentinel-2,RGB          & 1.5$\sim$60m     & 109.5  \\
                                       & SEN2VENmuS \cite{julien_michel_2022_6514159}                           & 2022 & 132955    & 256           & /         & Sentinel-1,Sentinel-2      & 10m$\sim$20m     & 87      \\
                                       & Proba-V Super Resolution \cite{marcus_martens_2018_6327426}            & 2018 & 1160      & 384           & /         & RNIR                       & 100$\sim$300m    & 0.71     \\    \hline
Unmixing                               & DLR HySU \cite{cerra2021dlr}                            & 2021 & 1         & 86x123        & /         & Hyperspectral              & 0.3m$\sim$1m     & 0.005     \\   \hline
\multirow{18}{*}{Urban 3D Point Cloud} & CORE3D \cite{brown2018large}                              & 2018 & /         & /             & /         & Panchromatic,MS,PointCloud & /                & /      \\
                & benchmark\_ISPRS2021 \cite{wu2021new}                & 2021 & 20        & 1024          & /         & RGB                        & 0.08m            & 3.22  \\
                                       & SemanticKITTI \cite{behley2019iccv}                       & 2019 & 4549M     & 39.2 km2      & 25        & PointCloud                 & /                & 80.2    \\
                                       & Toronto3D \cite{tan2020toronto3d}                           & 2020 & 78.3M     & 1.0 km2       & 8         & PointCloud                 & /                & 1.1     \\
                                       & Swiss3DCities \cite{Swiss3DCities,can2021swiss3dcities}                        & 2020 & 226M      & 2.7 km2       & 5         & PointCloud                 & /                & /     \\
                                       & DALES \cite{varney2020dales}                               & 2020 & 505.3M    & 10.0 km2      & 8         & PointCloud                 & /                & /        \\
                                       & LASDU \cite{li2013heihe}                               & 2020 & 3.12M     & 1.02 km2      & 5         & PointCloud                 & Aircraft @ 1200m & /        &  \\
                                       & Campus3D \cite{li2020campus3d}                            & 2020 & 937.1M    & 1.58 km2      & 14        & PointCloud                 & UAV              & /       \\
                                       & SensatUrban \cite{hu2022sensaturban}                         & 2020 & 2847.1M   & 6 km2         & 13        & PointCloud                 & UAV              & 20.6    \\
                                       & Heisenheim3D \cite{kolle2021hessigheim}                        & 2021 & 125.7M    & 0.19 km2      & 11        & PointCloud                 & /                & 58.8   \\
                                       & SUM-Helsinki \cite{gao2021sum}                        & 2021 & 19M       & 4 km2         & 6         & Mesh                       & 0.075m           & 8.35   \\
                                       & Oakland 3-D PointCloud \cite{munoz2009contextual}              & 2009 & 1.6 M     & 1.5 km        & 5         & PointCloud                 & /                & 0.033  \\
                                       & District of Columbia LiDAR 2015 \cite{Columbia_point_cloud}      & 2015 & /         & /             & 8         & PointCloud                 & /                & 34.4  \\
                                       & Semantic3D.net \cite{hackel2017isprs}                       & 2017 & 4000 M    & /             & 8         & PointCloud                 & /                & 24.5   \\
                                       & District of Columbia LiDAR 2018 \cite{Columbia_point_cloud}     & 2018 & /         & /             & 11        & PointCloud                 & /                & 279.8   \\
                                       & Paris-Lille-3D \cite{roynard2017parislille3d}                      & 2017 & 143 M     & 1.94 km2      & 50        & PointCloud                 & /                & 19.9  \\
                                       & DublinCity \cite{zolanvari2019dublincity}                           & 2019 & 260M      & 2.0 km2       & 13        & PointCloud                 & /                & 166.8  \\
                                       & Paris-rue-Madame \cite{serna2014paris}                    & 2014 & 20 M      & 1 km          & 26        & PointCloud                 & /                & /     \\   \hline
Vehicles                               & AM3D-Real \cite{AM3D}                           & 2022 & 1,012     & 720x480       & 2         & RGB,PointCloud             & 0.25m            & /      \\   \hline
VQA,Change Detection                   & CDVQA \cite{yuan2022change}                               & 2022 & 2968      & 512           & 19        & RGB                        & 0.5-3m           & /      \\  \hline
Water Body                             & Forel-Ule Index global inland waters \cite{Wang2020} & 2021 & /         & /             & /         & MS                         & 500m             & 0.004     \\    \hline
\multirow{3}{*}{Weather}               & Historical Hourly Weather \cite{Historical_Hourly}           & 2017 & /         & /             & /         & Weather Attributes         & /                & 0.075     \\
                                       & DeepWeather \cite{grnquist2020deep}             & 2020 & 20  & /             & 9         & Weather Attributes         & /                & 2,965   \\
                                       & ClimateNet \cite{kashinath2021climatenet}             & 2020 & 459  & 16 (768,1152)             & 9         & NC variables    & /                & 28   \\
                                       & EarthNet2021 \cite{requena2021earthnet2021}                      & 2021 & 32000     & 128           & /         & Sentinel-2                 & 20m              & 614.4     \\  \hline
 \multirow{3}{*}{Climate}                   & Earth Surface Temperature \cite{EST}                      & 2017 & 16     & /           & /         & Climate variables                 & /              & 0.09      \\ 
                                        & Greenhouse Gas \cite{IGGE}                      & 2017 & /     & /           & /         & Gas Emission            & /              & 0.0001    \\ 
                                   & Hurricane Wind Speed\cite{maskey2020deepti}                 & 2020 & 114634    & 366           & /         & LWIR                       & /                & 2.24      \\ \hline  \hline
\end{tabular}
}
\label{RSOTHER}
\end{table*}

\textbf{Image enhancement} aims to convert an input image to an enhanced one that has richer information and better details. For instance, super resolution \cite{djerida2021new,julien_cornebise_2022_6810792,julien_michel_2022_6514159,marcus_martens_2018_6327426}, image dehazing \cite{huang2020single}, and cloud removal \cite{DVN/BSETKZ_2019,ebel2020multisensor,ji2020simultaneous} are research tasks that are useful for improving the image quality for many downstream applications. 

\textbf{Multi-modal learning} attempts to combine the strengths of different modalities of data to enhance the representations for different tasks. There are some remote sensing datasets designed for multi-modal data fusion \cite{schmitt2018sen1,huang2021qxs} and cross-view geo-localization \cite{workman2015wide,zheng2020university,tian2017cross} tasks.

\textbf{3D understanding} is crucial for many real-world applications like urban planning, disaster monitoring, flood management, and autonomous driving. Hence, many datasets have been designed for different tasks, including 3D reconstruction, point cloud analysis \cite{tan2020toronto3d,Swiss3DCities,can2021swiss3dcities,li2020campus3d}, and multi-view stereo \cite{SatStereo,liu2020novel,Gao_2021_ICCV}.

\textbf{Earth system modeling} seeks to study physical, chemical, and biological processes in order to understand the Earth planet with complex integration of environmental variables. To this end, datasets are designed for plant/tree \cite{lopez2019columnar,reiersen2022reforestree,ventura2022individual} analysis, forest monitoring \cite{juan_miguel_requena_mullor_2019_3468645}, soil parameter estimation \cite{Hyperview_Challenge}, geophysical \cite{wang2019labelled}, air quality monitoring \cite{EEA_Air}, and population estimation \cite{doda2022so2sat}. There are also Earth observation datasets constructed for climate variable estimation \cite{kashinath2021climatenet} and weather forecasting \cite{grnquist2020deep}. 

\begin{figure*}[htbp]
\centering
\includegraphics[scale=0.28]{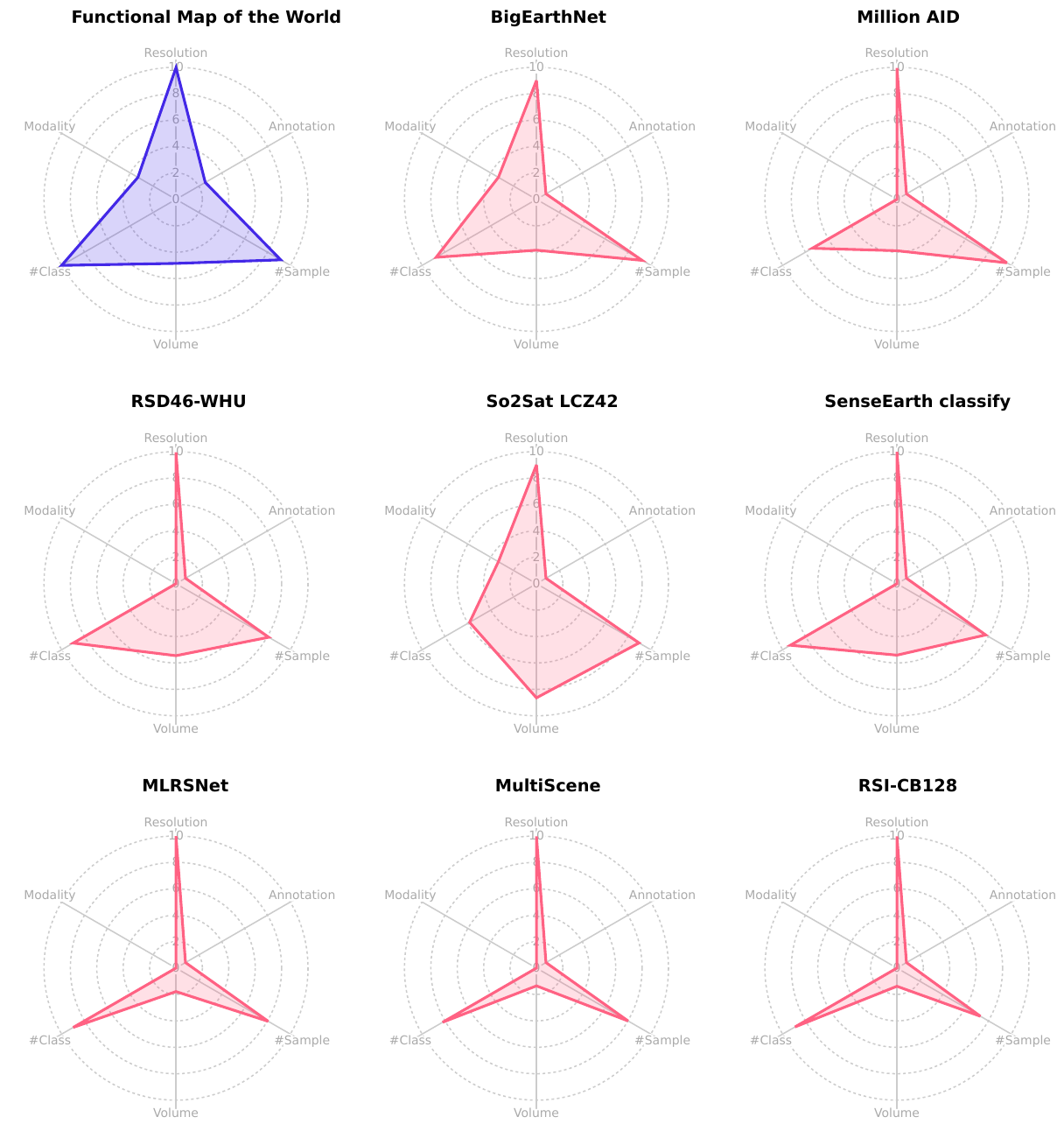}
\hspace{0.05in}
\includegraphics[scale=0.28]{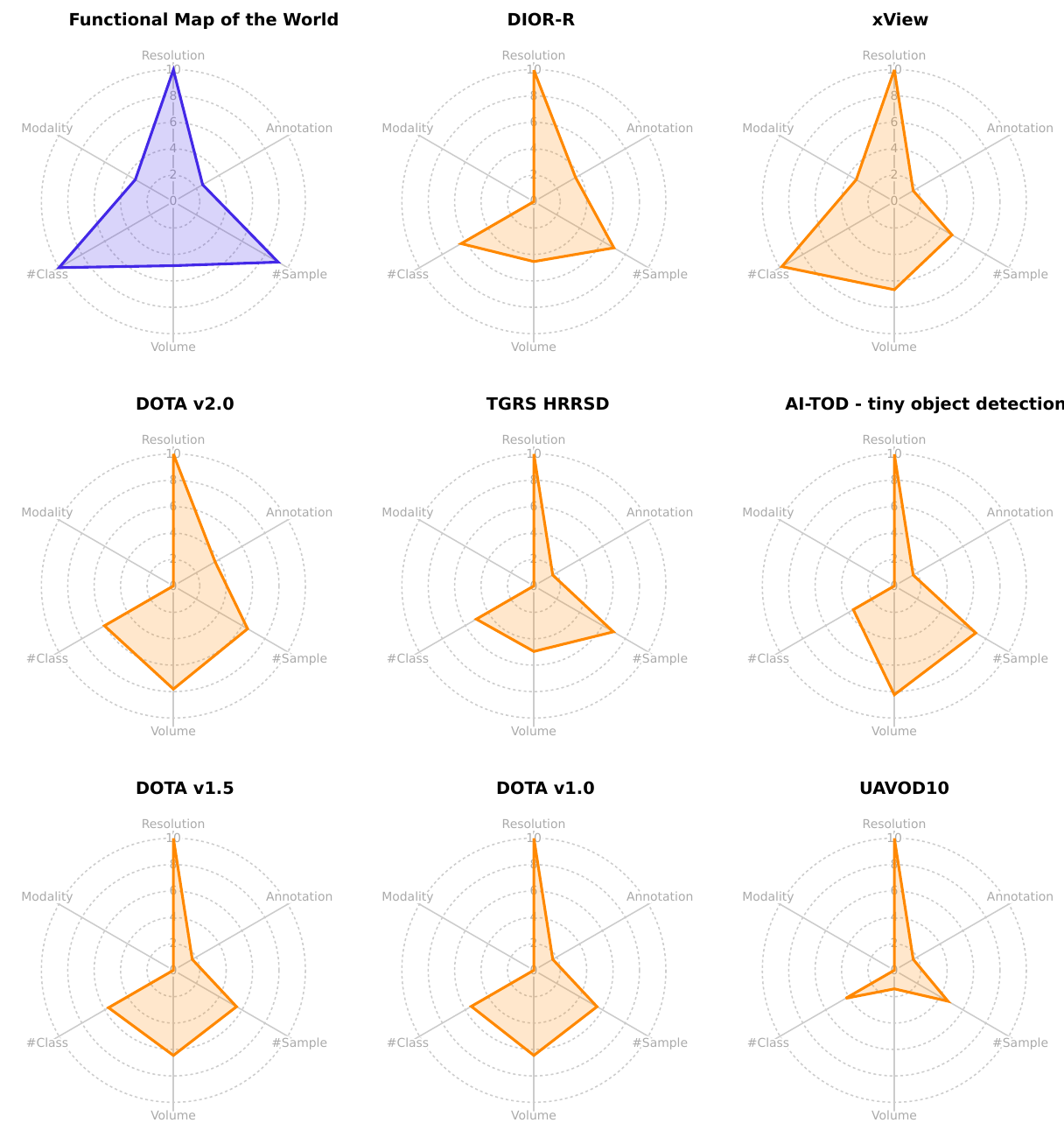}
\hspace{0.05in}
\includegraphics[scale=0.28]{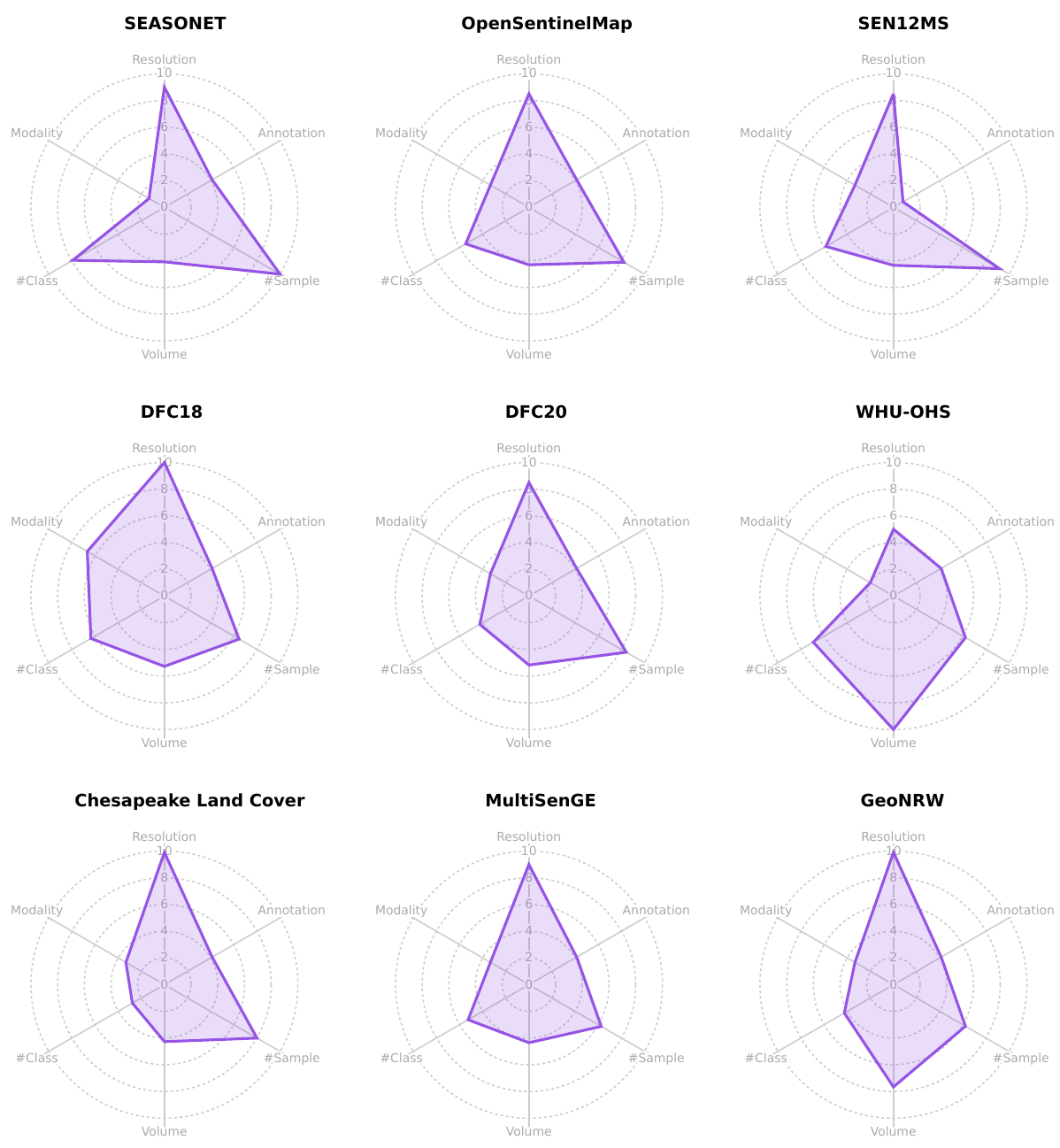}
\caption{Radar charts visualization of some top-ranked RS datasets for three different tasks. Six different attributes are compared and displayed in this figure. Note that all the attribute values are normalized to the range of 0 to 10.}
\label{radar}
\end{figure*}

\begin{figure*}
\centering
\includegraphics[scale=0.55]{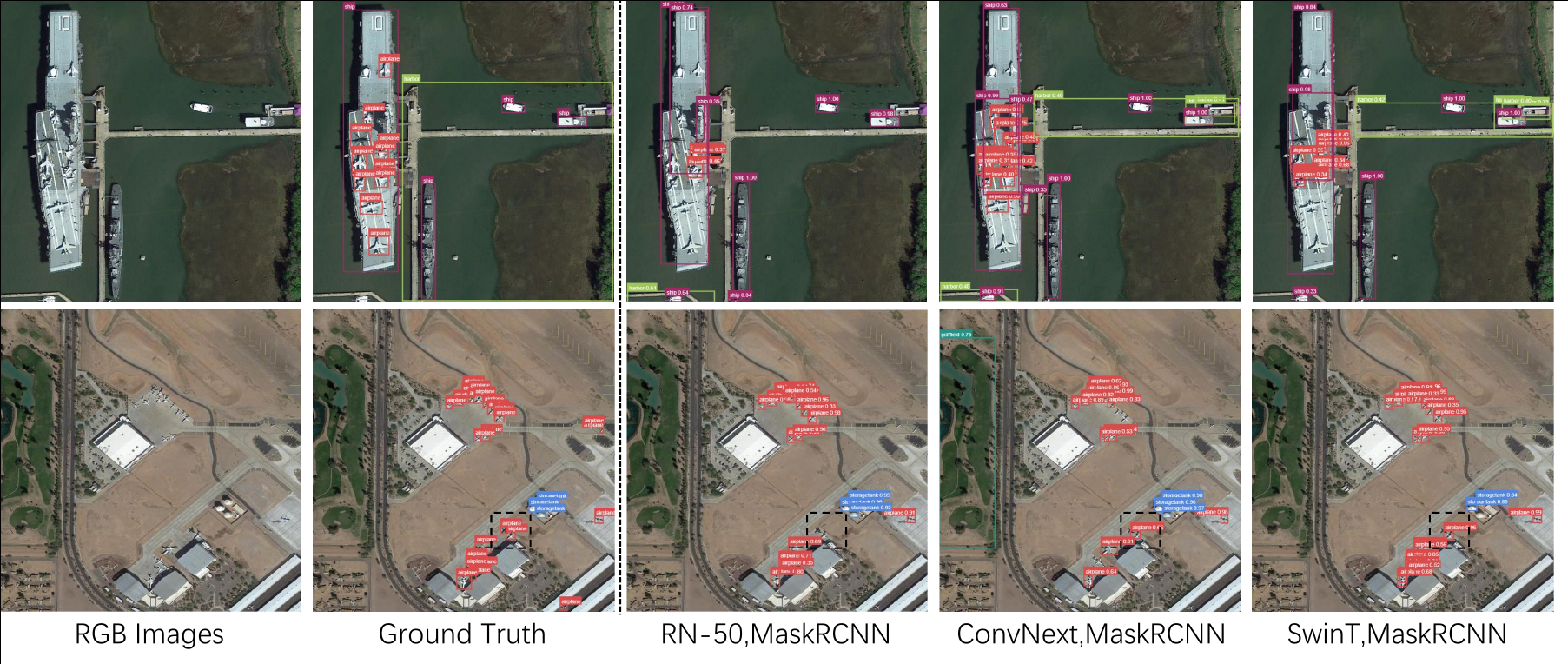}
\hspace{0.05in}
\includegraphics[scale=0.59]{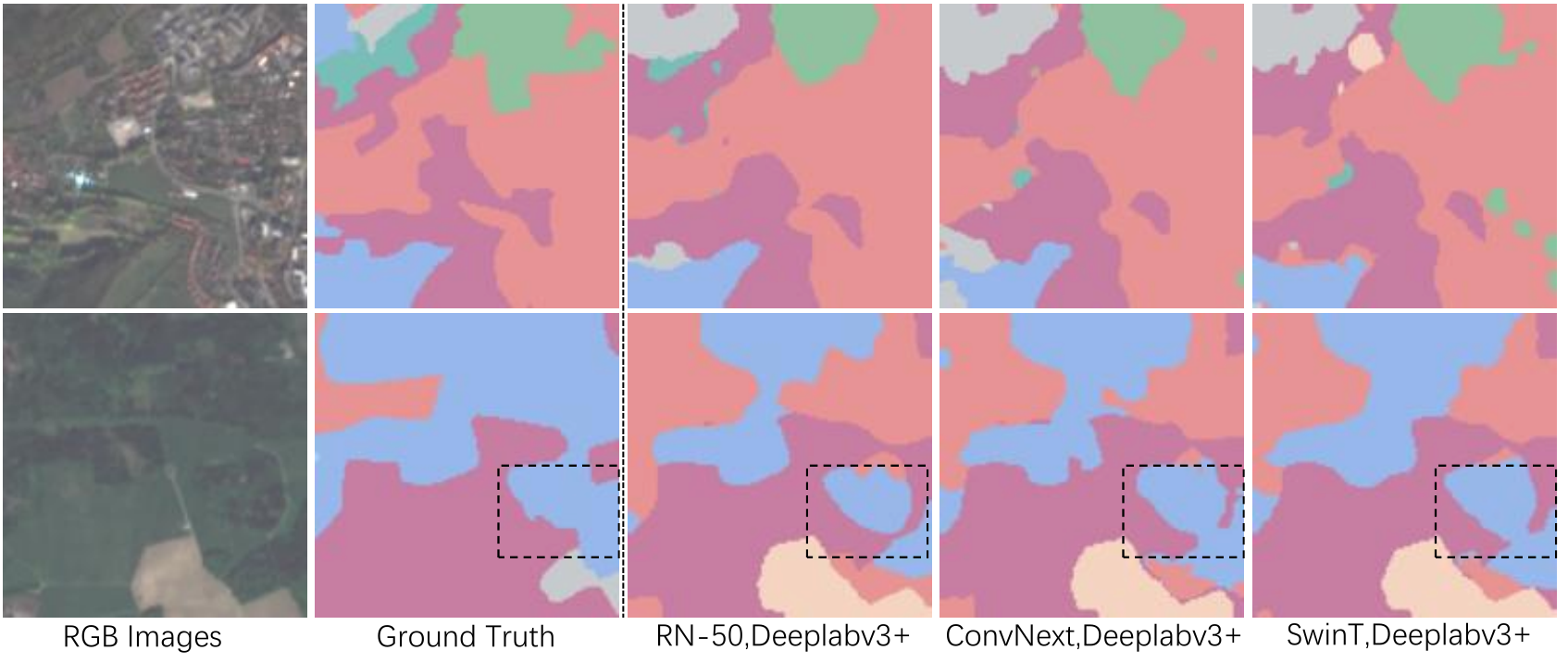}
\hspace{0.05in}
\includegraphics[scale=0.55]{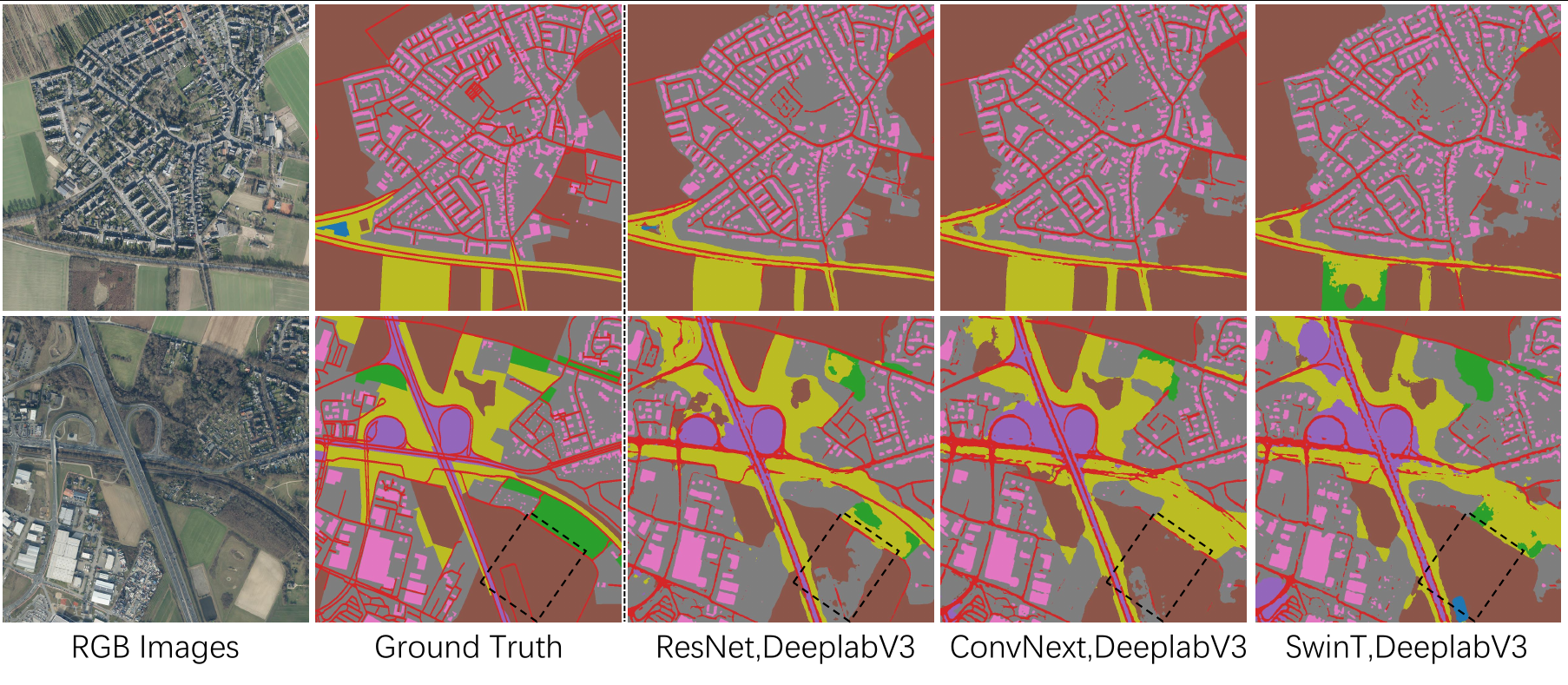}
\caption{Object detection and semantic segmentation results on three datasets. The first two rows present the object detection results on the DIOR dataset. The second two rows show the semantic segmentation results on the SEASONET dataset. The final two rows are the semantic segmentation results on the GeoNRW dataset.}
\label{example_show}
\end{figure*}

\section{Implementation of EarthNets Platform}
In this section, we will detail the implementation of the main libraries in the EarthNets platform \textbf{Dataset4EO}. At present, there are some difficulties in loading RS datasets, especially for researchers in other communities. 1) The datasets have different downloading links, folder structures, file formats, data modalities, research domains, and annotation levels. It would be helpful if it were possible to download, decompress, and split the dataset automatically. 2) Images in RS datasets usually contain multiple modalities and bands. To accelerate the training speed, it is better to move the data augmentation to GPU. 3) For datasets with very large volumes or time-series streaming data, support iterable-style data pipes would be useful to handle the dataset loading process.
\begin{figure*}
\centering
\includegraphics[scale=0.12]{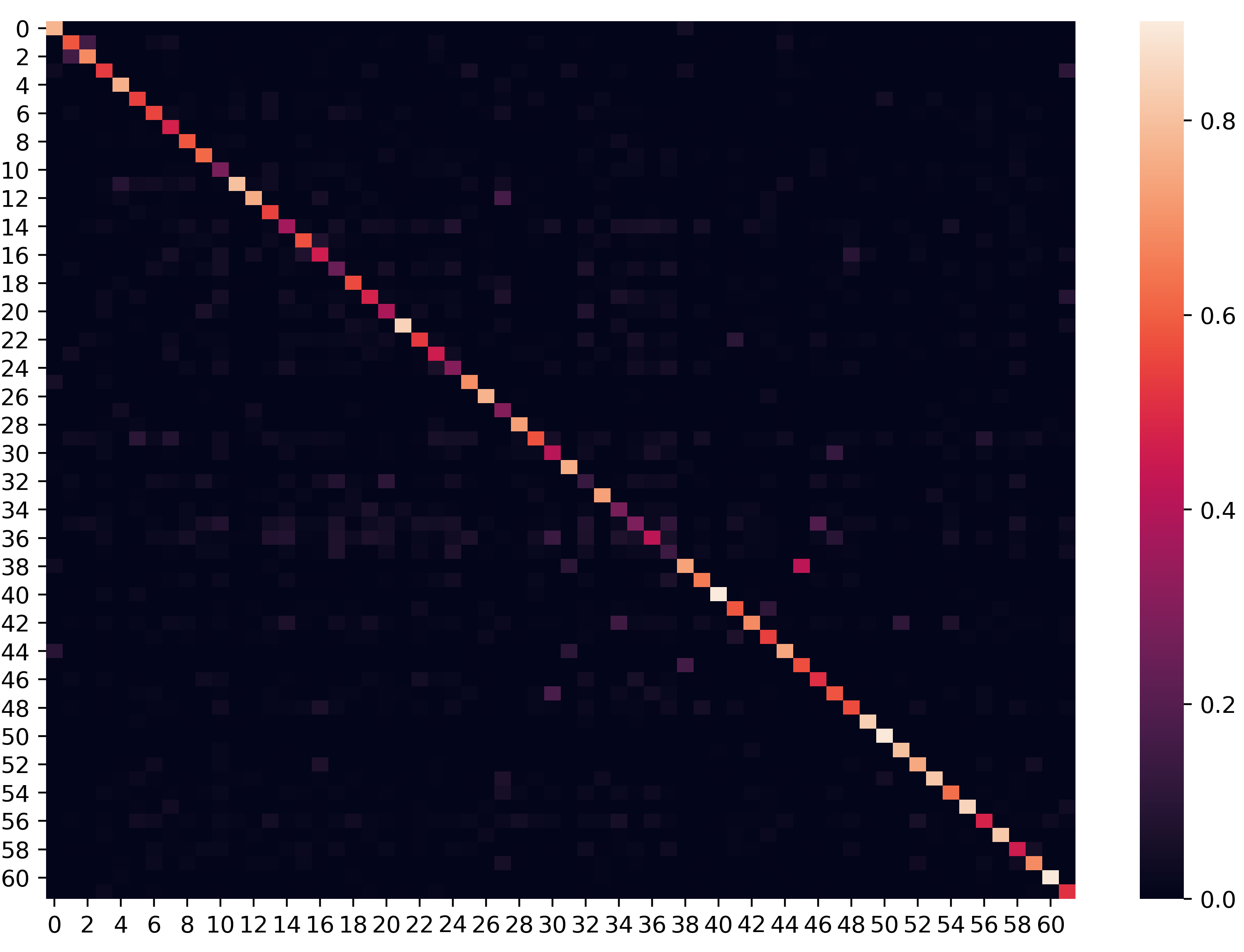}
\hspace{0.05in}
\includegraphics[scale=0.12]{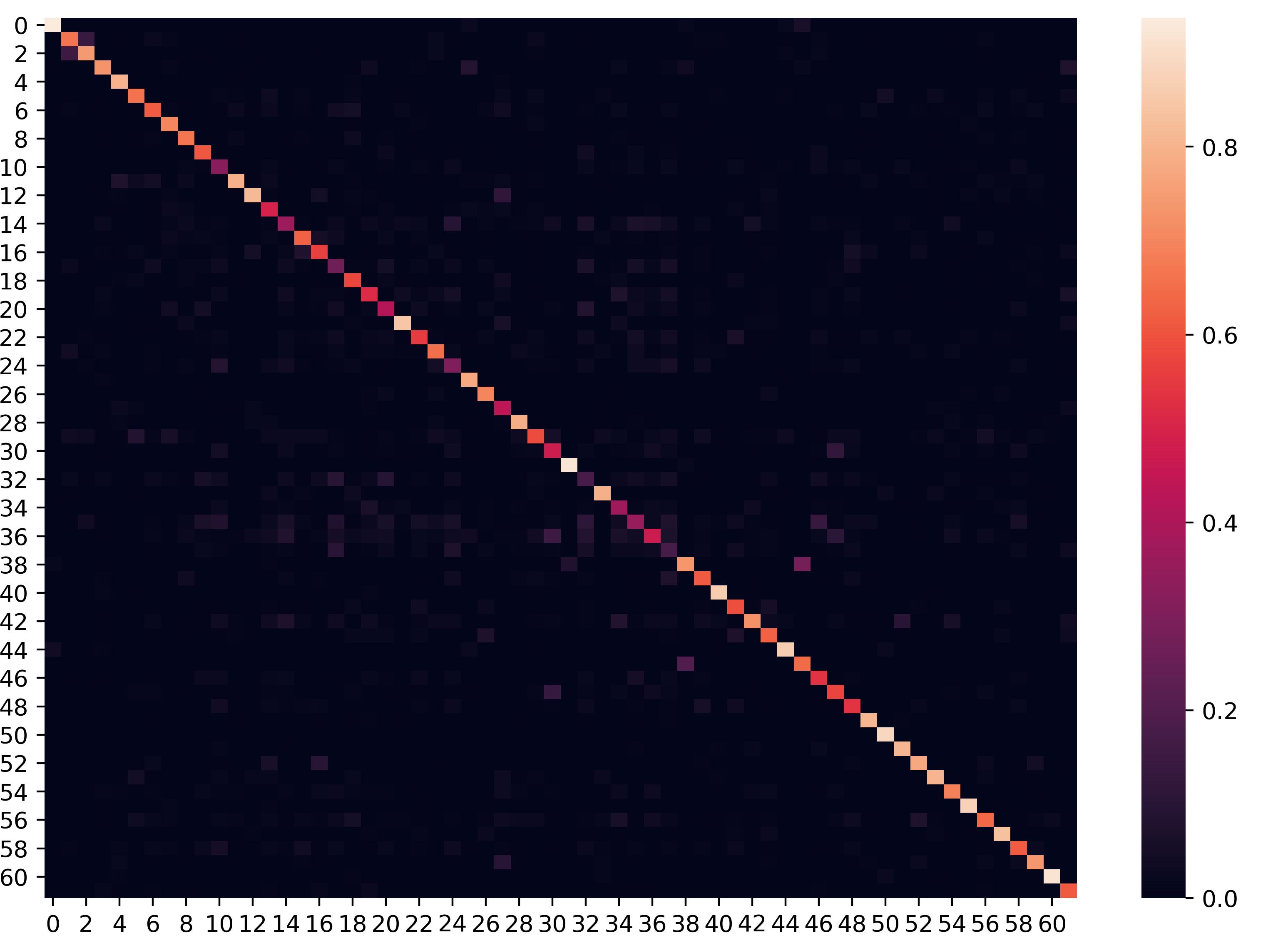}
\hspace{0.05in}
\includegraphics[scale=0.12]{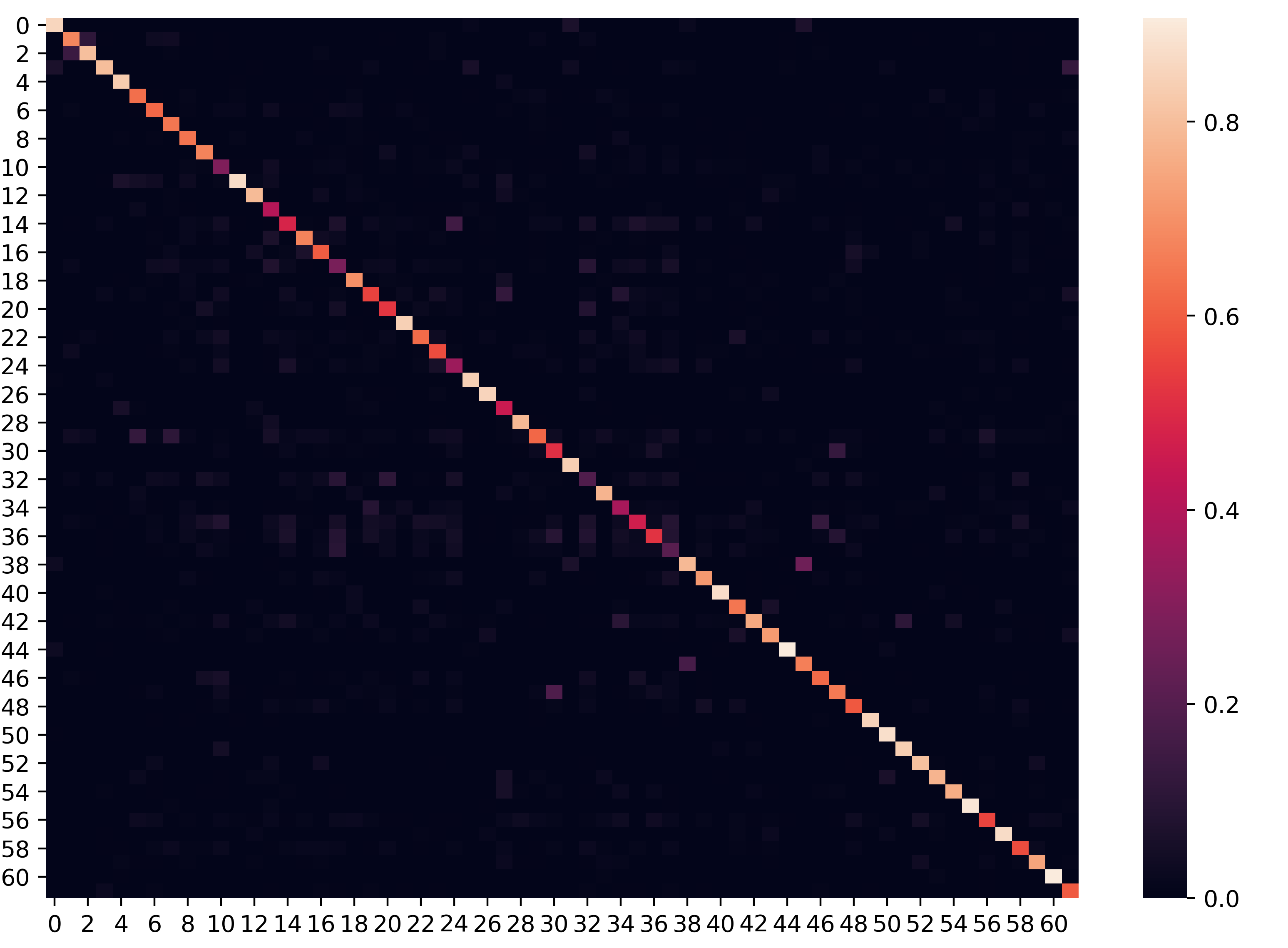}
\caption{Confusion matrices of different methods on the FMoW dataset \cite{sumbul2019bigearthnet}. From left to right are the confusion matrices for the ResNet-50, ConvNext, and Swin-Transformer models.}
\label{cfms}
\end{figure*}

\textbf{Remote Sensing Tasks} Based on the Dataset4EO library, we build the RS Classification, RS Detection, and RS Segmentation libraries. All these libraries share the same dataset loading module. To establish a deep connection with the CV community, we base these libraries for RS tasks on the libraries from OpenMMlab \cite{chen2019mmdetection}. EarthNets enables an easy adaptation of modern deep learning models from the CV community to the RS community. For example, backbone models like ResNet \cite{he2016deep}, EfficientNet \cite{tan2019efficientnet}, ConvNext\cite{liu2022convnet}, Vision Transformers \cite{dosovitskiy2020image}, MLP-Mixer \cite{tolstikhin2021mlp}, Swin Transformer \cite{liu2021swin}, and so forth can be used. Numerous  state-of-the-art architectures designed for CV tasks like RetinaNet \cite{lin2017focal}, UNet \cite{ronneberger2015u}, Deeplab \cite{chen2017rethinking}, YOLO \cite{redmon2016you}, Upernet \cite{xiao2018unified}, and SegFormer \cite{xie2021segformer} can be applied to RS data. By this means, EarthNets can serve as a bridge between the CV and RS communities.

\section{Implementation Details for Benchmarking Experiments}
\textbf{Optimizer}: For convolution-based models, SGD is used as the optimizer. The AdamW \cite{loshchilov2017decoupled} is used for optimizing the Transformer-based models. 
\textbf{Initialization}: By default, we use the ImageNet pre-trained weights for initializing the models. For some architectures with no ImageNet \cite{deng2009imagenet} pre-trained weights, we train them from scratch. Other hyper-parameters including batch size and learning rate are set differently for each dataset. More implementation details are provided in the public codes at \url{https://github.com/EarthNets}.

\section{Visualization Results of Benchmarking Experiments}
In Fig. \ref{radar}, we use radar charts to compare the attributes of some top-ranked datasets. We compare six different attributes of the datasets: the modality, the resolution, the annotation level, the number of classes, the number of samples, and the volume. Using radar charts makes it easy to compare and rank different datasets. A complete set of the charts is displayed on \url{https://earthnets.github.io}.

For object detection and semantic segmentation tasks, we provide the visualization results to directly compare the results of different network architectures on large-scale datasets in Fig. \ref{example_show}. The first two rows present the object detection results on the DIOR dataset. The second two rows show the semantic segmentation results on the SEASONET dataset. The final two rows are the semantic segmentation results on the GeoNRW dataset. From the visualization results we can further verify that Transformer-based networks work well on large-scale remote sensing datasets. This is consistent with the conclusions on the computer vision datasets. However, it is worth mentioning that in some local regions of the image, the CNN-based method, i.e, ResNet-50, works better than Swin-Transformer. Hence, an effective combination of these two architectures could yield better performance. In Table \ref{table_geonrw}, Table \ref{table_seasonet} and Table \ref{table_fmow}, we provide detailed experimental results for each semantic category. 

For the image classification task on the FMoW dataset, we also show the confusion matrices to compare the performance of different network architectures in Fig. \ref{cfms}. From left to right, the figure displays the confusion matrix for the ResNet-50, ConvNext, and Swin-Transformer.

\begin{table*}[]
\centering
\caption{Detailed Comparison Results on the GeoNRW \cite{baier2021synthesizing} Dataset}
\label{table_geonrw}
\scalebox{0.9}{
\begin{tabular}{c|cc|cc|cc}
\hline \hline
Methods                           & \multicolumn{2}{c|}{DeepLabv3+/RN50}             & \multicolumn{2}{c|}{DeepLabv3+/ConvNext-Tiny}    & \multicolumn{2}{c}{DeepLabv3+/Swin-Tiny} \\ \hline
Metrics                           & \textbf{IoU} & \multicolumn{1}{c|}{\textbf{Acc}} & \textbf{IoU} & \multicolumn{1}{c|}{\textbf{Acc}} & \textbf{IoU}        & \textbf{Acc}       \\ \hline
forest                            & 72.49        & 84.57                             & 71.07        & 84.93                             & 72.22               & 83.97              \\
water                             & 90.14        & 92.91                             & 91.4         & 94.54                             & 91.69               & 95.07              \\
agricultural                      & 86.66        & 92.33                             & 86.2         & 92.56                             & 86.48               & 92.66              \\
residential,commercial,industrial & 69.67        & 85.5                              & 67.96        & 84.8                              & 68.38               & 85.2               \\
grassland,swamp,shrubbery         & 38.92        & 58.25                             & 35.72        & 53.96                             & 36.98               & 56.23              \\
railway,trainstation              & 59.03        & 73.76                             & 56.66        & 77.29                             & 58.89               & 78.58              \\
highway,squares                   & 43.97        & 51.96                             & 39.89        & 46.17                             & 41.31               & 48.79              \\
airport,shipyard                  & 55.37        & 79.18                             & 59.35        & 73.98                             & 58.77               & 75.34              \\
roads                             & 43.05        & 53.14                             & 39.32        & 48.3                              & 38.86               & 47.95              \\
buildings                         & 70.75        & 81.04                             & 68.76        & 79.54                             & 68.24               & 78.97              \\ \hline \hline
\end{tabular}}
\end{table*}

\begin{table*}[]
\centering
\caption{Detailed Comparison Results on the SEASONET \cite{kossmann2022seasonet} Dataset}
\label{table_seasonet}
\scalebox{0.8}{
\begin{tabular}{c|cc|cc|cc}
\hline
Methods                                    & \multicolumn{2}{c|}{DeepLabv3+/RN50}             & \multicolumn{2}{c|}{DeepLabv3+/ConvNext-Tiny}    & \multicolumn{2}{c}{DeepLabv3+/Swin-Tiny} \\ \hline
Metrics                                    & \textbf{IoU} & \multicolumn{1}{c|}{\textbf{Acc}} & \textbf{IoU} & \multicolumn{1}{c|}{\textbf{Acc}} & \textbf{IoU}        & \textbf{Acc}       \\ \hline
Continuous Urban Fabric                    & 36.38        & 44.72                             & 33.28        & 39.83                             & 36.02               & 44.87              \\
Discontinuous Urban Fabric                 & 74.01        & 87.85                             & 71.22        & 86.83                             & 73                  & 87.06              \\
Industrial or Commercial Units             & 53.94        & 67.27                             & 47.52        & 60.65                             & 52.11               & 65.89              \\
Road and Rail Networks and Associated Land & 42.94        & 57.27                             & 33.73        & 44.09                             & 39.06               & 53.54              \\
Port Areas                                 & 33.28        & 46.01                             & 25.28        & 29.79                             & 30.77               & 38.1               \\
Airports                                   & 56.54        & 74.48                             & 44.64        & 55.86                             & 56.74               & 68.67              \\
Mineral Extraction Sites                   & 59.36        & 76.3                              & 52.06        & 69.84                             & 57.53               & 75.17              \\
Dump Sites                                 & 25.4         & 32.51                             & 8.65         & 9.79                              & 22.39               & 28.24              \\
Construction Sites                         & 8.52         & 9.7                               & 2.42         & 2.53                              & 7.12                & 7.84               \\
Green Urban Areas                          & 30.58        & 43.06                             & 23.23        & 29.66                             & 27.62               & 37.05              \\
Sport and Leisure Facilities               & 39.7         & 48.86                             & 29.98        & 36.82                             & 36.18               & 44.6               \\
Non-irrigated Arable Land                  & 83.67        & 91.7                              & 80.99        & 90.23                             & 82.65               & 91.87              \\
Vineyards                                  & 75.34        & 88.8                              & 66.46        & 82.52                             & 73.73               & 85.66              \\
Fruit Trees and Berry Planatations         & 42.95        & 53.5                              & 33.82        & 43.59                             & 40.44               & 51                 \\
Pastures                                   & 64.1         & 78.73                             & 59.46        & 75.49                             & 61.89               & 76.15              \\
Broad-leaved Forest                        & 65.09        & 80.71                             & 62           & 79.62                             & 63.72               & 80.45              \\
Coniferous Forest                          & 76.01        & 89.41                             & 74.02        & 88.6                              & 75.16               & 88.92              \\
Mixed Forest                               & 28.56        & 36.48                             & 24.2         & 30.31                             & 26.54               & 33.45              \\
Natural Grasslands                         & 34.23        & 42.65                             & 28.07        & 84.97                             & 32.66               & 40.26              \\
Moors and Healthland                       & 59.66        & 78.55                             & 54.33        & 73.16                             & 58.86               & 76.18              \\
Transitional Woodland / Shrub              & 25.39        & 31.29                             & 20.46        & 25.17                             & 23.03               & 28.03              \\
Beaches, Dunes, Sands                      & 38.39        & 67.53                             & 40.91        & 53.36                             & 43.94               & 59.31              \\
Bare Rock                                  & 0            & 0                                 & 29.26        & 36.52                             & 34.67               & 50.25              \\
Sparsely Vegetated Areas                   & 10.13        & 18.06                             & 7.24         & 8.71                              & 7.45                & 9.57               \\
Inland Marshes                             & 31.86        & 44.94                             & 23.77        & 31.19                             & 28.2                & 38.27              \\
Peat Bogs                                  & 56.68        & 72.8                              & 50.81        & 63.44                             & 54.5                & 69.69              \\
Salt Marshes                               & 70.42        & 86.18                             & 60.76        & 77.6                              & 66.55               & 80.6               \\
Intertidal Flats                           & 76.26        & 86.21                             & 77.72        & 89.63                             & 80.02               & 90.66              \\
Water Courses                              & 64.86        & 77.79                             & 54.93        & 68.02                             & 60.61               & 72.87              \\
Water Bodies                               & 77.94        & 86.09                             & 73.38        & 83.31                             & 78.25               & 87.76              \\
Coastal Lagoons                            & 79.77        & 90.37                             & 80.55        & 89.6                              & 83.7                & 92.3               \\
Estuaries                                  & 59.04        & 80.07                             & 60.1         & 69.98                             & 65.3                & 77.48              \\
Sea and Ocean                              & 95.16        & 97.5                              & 95.54        & 97.52                             & 96.19               & 97.75              \\ \hline \hline
\end{tabular}}
\end{table*}

\begin{table*}[]
\centering
\caption{Detailed Comparison Results on the FMoW \cite{fmow2018} Dataset}
\label{table_fmow}
\scalebox{0.9}{
\begin{tabular}{c|ccc|ccc|ccc}
\hline \hline
Methods                         & \multicolumn{3}{c|}{\textbf{RN50}}                        & \multicolumn{3}{c|}{\textbf{ConvNext-Tiny}}                                   & \multicolumn{3}{c}{\textbf{Swin-Tiny}}                                       \\ \hline
Metrics                         & \multicolumn{2}{c}{\textbf{Precision}} & \textbf{Recall} & \multicolumn{2}{c}{\textbf{Precision}} & \multicolumn{1}{c|}{\textbf{Recall}} & \multicolumn{2}{c}{\textbf{Precision}} & \multicolumn{1}{c}{\textbf{Recall}} \\ \hline
Airport                         & 77.18              & 90.29             & 83.22           & 93.72              & 86.89             & 90.18                               & 85.71              & 90.29             & 87.94                               \\
Airport\_hangar                 & 58.25              & 57.82             & 58.03           & 66.50              & 58.55             & 62.27                               & 67.71              & 66.81             & 67.26                               \\
Airport\_terminal               & 68.45              & 72.54             & 70.43           & 74.20              & 71.57             & 72.86                               & 80.34              & 75.93             & 78.07                               \\
Amusement\_park                 & 53.25              & 82.80             & 64.81           & 73.10              & 83.24             & 77.84                               & 79.85              & 85.35             & 82.51                               \\
Aquaculture                     & 76.71              & 81.28             & 78.93           & 80.08              & 85.53             & 82.72                               & 82.92              & 84.68             & 83.79                               \\
Archaeological\_site            & 54.40              & 43.49             & 48.34           & 66.11              & 51.82             & 58.10                               & 63.08              & 53.39             & 57.83                               \\
Barn                            & 55.02              & 57.89             & 56.42           & 61.58              & 60.66             & 61.12                               & 61.71              & 69.40             & 65.33                               \\
Border\_checkpoint              & 47.54              & 31.69             & 38.03           & 70.00              & 26.78             & 38.74                               & 64.55              & 38.80             & 48.46                               \\
Burial\_site                    & 58.39              & 60.23             & 59.30           & 66.73              & 63.06             & 64.84                               & 64.35              & 70.88             & 67.46                               \\
Car\_dealership                 & 61.86              & 66.49             & 64.09           & 60.92              & 76.61             & 67.87                               & 67.22              & 74.38             & 70.62                               \\
Construction\_site              & 27.70              & 16.78             & 20.90           & 32.13              & 15.47             & 20.88                               & 29.34              & 22.44             & 25.43                               \\
Crop\_field                     & 80.07              & 89.51             & 84.53           & 79.63              & 92.35             & 85.52                               & 86.96              & 91.98             & 89.40                               \\
Dam                             & 75.58              & 74.31             & 74.94           & 81.38              & 75.54             & 78.35                               & 79.02              & 83.49             & 81.19                               \\
Debris\_or\_rubble              & 54.84              & 19.92             & 29.23           & 49.61              & 25.00             & 33.25                               & 40.72              & 26.56             & 32.15                               \\
Educational\_institution        & 36.43              & 35.57             & 35.99           & 36.46              & 41.88             & 38.98                               & 48.22              & 46.46             & 47.32                               \\
Electric\_substation            & 57.33              & 59.22             & 58.26           & 62.71              & 65.08             & 63.87                               & 67.14              & 69.84             & 68.46                               \\
Factory\_or\_powerplant         & 46.00              & 32.80             & 38.29           & 56.73              & 34.58             & 42.97                               & 59.69              & 47.77             & 53.07                               \\
Fire\_station                   & 24.54              & 17.25             & 20.26           & 26.26              & 16.09             & 19.95                               & 27.73              & 24.84             & 26.21                               \\
Flooded\_road                   & 56.28              & 40.12             & 46.85           & 57.64              & 40.74             & 47.74                               & 69.83              & 50.00             & 58.27                               \\
Fountain                        & 47.86              & 33.22             & 39.22           & 51.65              & 39.27             & 44.62                               & 55.20              & 46.69             & 50.59                               \\
Gas\_station                    & 38.32              & 42.08             & 40.11           & 41.84              & 43.58             & 42.70                               & 52.45              & 56.99             & 54.63                               \\
Golf\_course                    & 84.54              & 72.74             & 78.20           & 84.35              & 72.58             & 78.02                               & 83.97              & 83.42             & 83.69                               \\
Ground\_transportation\_station & 53.03              & 59.32             & 56.00           & 55.28              & 65.34             & 59.89                               & 62.47              & 71.12             & 66.51                               \\
Helipad                         & 45.77              & 37.36             & 41.14           & 65.27              & 31.87             & 42.83                               & 56.98              & 46.15             & 51.00                               \\
Hospital                        & 30.14              & 30.69             & 30.41           & 30.28              & 36.58             & 33.13                               & 35.13              & 33.69             & 34.40                               \\
Impoverished\_settlement        & 69.20              & 83.17             & 75.55           & 77.46              & 90.87             & 83.63                               & 84.07              & 91.35             & 87.56                               \\
Interchange                     & 77.08              & 79.49             & 78.27           & 69.93              & 83.59             & 76.16                               & 85.34              & 81.84             & 83.55                               \\
Lake\_or\_pond                  & 30.14              & 16.67             & 21.46           & 43.56              & 33.33             & 37.77                               & 45.28              & 36.36             & 40.34                               \\
Lighthouse                      & 72.89              & 73.29             & 73.09           & 78.96              & 73.83             & 76.31                               & 78.92              & 81.77             & 80.32                               \\
Military\_facility              & 57.71              & 51.35             & 54.35           & 58.67              & 60.65             & 59.65                               & 61.80              & 59.62             & 60.69                               \\
Multi-unit\_residential         & 40.95              & 32.17             & 36.03           & 47.38              & 37.57             & 41.91                               & 51.17              & 45.39             & 48.11                               \\
Nuclear\_powerplant             & 75.86              & 75.86             & 75.86           & 92.00              & 79.31             & 85.19                               & 83.87              & 89.66             & 86.67                               \\
Office\_building                & 13.12              & 8.76              & 10.51           & 17.76              & 9.29              & 12.20                               & 18.84              & 9.72              & 12.83                               \\
Oil\_or\_gas\_facility          & 73.04              & 76.55             & 74.75           & 79.38              & 83.07             & 81.18                               & 77.78              & 81.52             & 79.61                               \\
Park                            & 27.81              & 20.08             & 23.32           & 36.67              & 26.69             & 30.89                               & 37.57              & 27.24             & 31.58                               \\
Parking\_lot\_or\_garage        & 28.77              & 35.33             & 31.71           & 35.28              & 41.92             & 38.31                               & 46.46              & 47.19             & 46.82                               \\
Place\_of\_worship              & 42.18              & 51.19             & 46.25           & 46.87              & 54.06             & 50.21                               & 51.96              & 61.25             & 56.22                               \\
Police\_station                 & 14.13              & 12.90             & 13.48           & 17.33              & 15.07             & 16.12                               & 20.59              & 18.77             & 19.64                               \\
Port                            & 73.44              & 79.72             & 76.45           & 73.53              & 88.97             & 80.52                               & 78.66              & 87.90             & 83.03                               \\
Prison                          & 65.23              & 46.52             & 54.31           & 61.42              & 54.32             & 57.65                               & 71.61              & 53.76             & 61.42                               \\
Race\_track                     & 90.18              & 81.83             & 85.80           & 85.81              & 87.65             & 86.72                               & 87.27              & 88.72             & 87.99                               \\
Railway\_bridge                 & 58.35              & 50.15             & 53.94           & 59.11              & 57.56             & 58.33                               & 64.69              & 63.89             & 64.29                               \\
Recreational\_facility          & 68.37              & 78.95             & 73.28           & 72.19              & 80.23             & 76.00                               & 75.11              & 84.55             & 79.55                               \\
Road\_bridge                    & 54.50              & 69.25             & 61.00           & 62.75              & 63.15             & 62.95                               & 72.13              & 74.33             & 73.22                               \\
Runway                          & 74.15              & 86.05             & 79.66           & 86.35              & 86.58             & 86.47                               & 90.49              & 92.63             & 91.55                               \\
Shipyard                        & 56.86              & 52.73             & 54.72           & 64.71              & 40.00             & 49.44                               & 66.28              & 51.82             & 58.16                               \\
Shopping\_mall                  & 50.89              & 49.30             & 50.08           & 53.69              & 57.28             & 55.43                               & 62.32              & 60.30             & 61.29                               \\
Single-unit\_residential        & 58.07              & 64.63             & 61.17           & 57.34              & 73.22             & 64.31                               & 65.39              & 69.23             & 67.25                               \\
Smokestack                      & 56.40              & 43.56             & 49.16           & 53.53              & 60.08             & 56.62                               & 58.82              & 64.63             & 61.59                               \\
Solar\_farm                     & 83.18              & 83.51             & 83.34           & 80.76              & 84.84             & 82.75                               & 84.70              & 86.84             & 85.75                               \\
Space\_facility                 & 89.13              & 93.18             & 91.11           & 88.64              & 88.64             & 88.64                               & 87.76              & 97.73             & 92.47                               \\
Stadium                         & 80.22              & 79.83             & 80.02           & 80.78              & 81.74             & 81.26                               & 83.81              & 87.11             & 85.43                               \\
Storage\_tank                   & 75.00              & 79.95             & 77.40           & 77.59              & 79.95             & 78.75                               & 80.77              & 85.28             & 82.96                               \\
Surface\_mine                   & 81.86              & 75.00             & 78.28           & 80.50              & 78.85             & 79.67                               & 77.84              & 82.01             & 79.87                               \\
Swimming\_pool                  & 62.86              & 64.35             & 63.60           & 68.84              & 63.78             & 66.21                               & 76.08              & 74.90             & 75.48                               \\
Toll\_booth                     & 84.85              & 88.33             & 86.56           & 87.37              & 94.87             & 90.96                               & 89.46              & 95.77             & 92.51                               \\
Tower                           & 47.76              & 32.20             & 38.47           & 64.29              & 36.54             & 46.60                               & 55.32              & 43.72             & 48.84                               \\
Tunnel\_opening                 & 82.05              & 82.05             & 82.05           & 83.20              & 86.75             & 84.94                               & 86.95              & 89.00             & 87.96                               \\
Waste\_disposal                 & 45.78              & 29.44             & 35.84           & 61.81              & 31.41             & 41.66                               & 56.98              & 49.01             & 52.70                               \\
Water\_treatment\_facility      & 68.49              & 69.04             & 68.77           & 73.69              & 75.77             & 74.72                               & 74.12              & 79.41             & 76.67                               \\
Wind\_farm                      & 89.82              & 94.20             & 91.96           & 91.82              & 96.18             & 93.95                               & 90.69              & 96.31             & 93.42                               \\
Zoo                             & 51.40              & 37.86             & 43.60           & 61.18              & 42.80             & 50.36                               & 59.26              & 46.09             & 51.85 \\ \hline \hline
\end{tabular}}
\end{table*}

\bibliography{refs}
\bibliographystyle{unsrt}

\end{document}